%% file: Manuscripts.tex
\documentclass[a4paper]{article}
\usepackage{adjustbox}
\usepackage[top=1.in, bottom=1.in, left=1.in, right=1.in]{geometry}
\usepackage{nopageno}
\usepackage{tabls}
\usepackage{cites}
\usepackage{epsf}
\usepackage{appendix}
\usepackage{ragged2e}
\usepackage{mathtools}
\usepackage{mdframed}
\usepackage{caption}
\usepackage{physics}
\usepackage{enumitem}
\usepackage{mathtools}
\usepackage{braket}
\usepackage{booktabs}
\usepackage[flushleft]{threeparttable}
\usepackage{algorithm}
\usepackage{algpseudocode}
\usepackage{nomencl} 
\makenomenclature
\usepackage{tikz}
\usepackage{graphicx}
\usepackage{float}
\floatstyle{plaintop}
\restylefloat{table}

\floatstyle{plain}
\restylefloat{figure}
\usepackage{amssymb}
\usepackage{pifont}
\usepackage{amsmath} 

\definecolor{light-gray}{gray}{0.95}
\newcommand{\code}[1]{\colorbox{light-gray}{\small\color{blue}\textbf{\texttt{#1}}}}

\def\bmheadfont{\reset@font\fontfamily{\rmdefault}\fontsize{10bp}{12bp}\bfseries\selectfont\raggedright\boldmath}%
\newcommand\bmhead{\@startsection{subparagraph}{5}{\z@}%
                                 {6pt \@plus1ex \@minus .2ex}%
                                 {-1em}%
                                 {\bmheadfont}}

\usepackage{url}
\providecommand{\keywords}[1]
{
  \small	
  \textbf{\textit{Keywords---}} #1
}

\title{Assessment of Reinforcement Learning Algorithms \\ for Nuclear Power Plant Fuel Optimization}
\author{
  \textbf{Paul Seurin$^1$, Koroush Shirvan$^1$}\footnote{Corresponding author. \textit{E-mail address}: kshirvan@mit.edu (K. Shirvan)} \\
  \\
  $^1$Massachusetts Institute of Technology \\
77 Massachusetts Avenue, Cambridge MA, 02139
\\ 
  \url{paseurin@mit.edu}, \url{kshirvan@mit.edu}
}
\date{} 

\begin{document}

\maketitle


\justify 
\begin{abstract}
The nuclear fuel loading pattern optimization problem belongs to the class
of large-scale combinatorial optimization. It is also characterized by multiple objectives and constraints, which makes it impossible to solve explicitly. Stochastic optimization methodologies including Genetic Algorithms and Simulated Annealing are used by different nuclear utilities and vendors but hand-designed solutions continue to be the prevalent method in the industry. To improve the state-of-the-art, Deep Reinforcement Learning (RL), in particular, Proximal Policy Optimization is leveraged. This work presents a first-of-a-kind approach to utilize deep RL to solve the loading pattern problem and could be leveraged for any engineering design optimization. This paper is also to our knowledge the first to propose a study of the behavior of several hyper-parameters that influence the RL algorithm. The algorithm is highly dependent on multiple factors such as the shape of the objective function derived for the core design that behaves as a fudge factor that affects the stability of the learning. But also an \textit{exploration/exploitation} trade-off that manifests through different parameters such as the number of loading patterns seen by the agents per episode, 
the number of samples collected before a policy update \code{nsteps}, and an entropy factor \code{ent\_coef} that increases the randomness of the policy during training. We found that RL must be applied similarly to a Gaussian Process in which the acquisition function is replaced by a parametrized policy. Then, once an initial set of hyper-parameters is found, reducing \code{nsteps} and \code{ent\_coef} until no more learning is observed will result in the highest sample efficiency robustly and stably. This resulted in an economic benefit of 535,000 - 642,000 \$/year/plant. Future work must extend this research to multi-objective settings and comparing them to state-of-the-art implementation of stochastic optimization methods.
\end{abstract} \hspace{10pt}

\begin{mdframed}
\begin{table}[H]   
\nomenclature{TRPO}{Trust Region Policy Optimization}
\nomenclature{ACKTR}{Actor Critic using Kronecker-Factored Trust Region}
\nomenclature{PG}{Policy Gradient}
\nomenclature{A2C}{Advantage Actor Critic}
\nomenclature{PPO}{Proximal Policy Optimization}
\nomenclature{RL}{Reinforcement Learning}
\nomenclature{FIM}{Fisher Information Matrix}
\nomenclature{\code{VF\_coef}}{Value Function Coefficient}
\nomenclature{\code{ent\_coef}}{Entropy Coefficient}
\nomenclature{\code{n\_steps}}{Number of steps per agent before parameter update}
\nomenclature{\code{LR}}{Learning Rate}
\nomenclature{\code{Clip} or $\epsilon$}{Clipping range}
\nomenclature{\code{noptepochs}}{Number of epoch to run the gradient ascent}
\nomenclature{NF}{Numframes: Number of pattern generated per episode}
\nomenclature{LP}{Loading Pattern}
\nomenclature{$N_{FA}$}{Number of fuel assemblies in the core}
\nomenclature{FAs}{Fuel Assemblies}
\nomenclature{BP}{Burnable Poison}
\nomenclature{CRD}{Control Rods}
\nomenclature{NPP}{Nuclear Power Plant}
\nomenclature{CO}{Combinatorial Optimization}
\nomenclature{SO}{Stochastic Optimization}
\nomenclature{NT}{Nemenyi Test}
\nomenclature{FT}{Friedman Test}
\nomenclature{SE}{Sample Efficiency}
\nomenclature{HT}{Hypothesis Testing}
\printnomenclature
\end{table}
\end{mdframed}

\section{Introduction} 
\label{sec:intro}
The commercial Nuclear Power Plants (NPP)s in the USA, comprised of Pressurized Water Reactors (PWR)s and Boiling Water Reactors (BWR)s, represent the largest single source of carbon-free energy in the country, generating 20\% of the electricity use year-round. Nevertheless, existing power plants are threatened to shut down early as they fail to be economical against alternative ones such as gas power plants or heavily subsidized solar and wind. Optimizing the fuel cycle cost via the optimization of the core Loading Pattern (LP) is one way of combating this lack of competitiveness. The loading pattern optimization problem aims to optimize the fuel cycle cost which accounts for 20-30\% of the total cost of electricity generation for the existing fleet \cite{nei2020nuclear} while meeting operational, regulatory, and safety goals. For the remainder of the paper, the notation LCOE (Levelized Cost of Electricity) refers to the fuel cycle cost. To solve this problem, a reactor designer must choose a set of Fuel Assemblies (FAs) to refuel its reactor. A reactor is fueled with a certain number of solid FAs of two types to choose from: either fresh or burned.  The burned FAs (about two-thirds of them) have already been in the core for a certain period (here about 18 or 36 months), while the fresh FAs have not. There is also a discrete choice of fresh FAs (which corresponds to a discrete set of input parameters such as the average enrichment of the fuel). This problem, therefore, belongs to the class of Combinatorial Optimization (CO), which is a sub-field of applied mathematics combining techniques to solve problems over discrete structures. Without a suitable objective function, combinatorial problems require evaluations of all possibilities to ensure a global optimum is found. 

For the loading pattern optimization, the size of the state space and the time to evaluate one candidate solution makes it prohibitive to utilize brute force enumeration. Furthermore, in this application, a closed-form objective function and analytical derivatives are not accessible. As such, Stochastic Optimization (SO) with heuristic methods has prevailed.  Simulated Annealing (SA) and Genetic Algorithm (GA) are utilized to drive the search in \cite{kropaczek2011copernicus,park2009multiobjective} and \cite{parks1996multiobjective} respectively. Ant Colony System (ACS), Particle Swarm Optimization (PSO), Cross-Entropy (CE), Population-based Incremental Learning (PBIL), and Artificial Bee Colonies (ABC) were also successfully leveraged \cite{moura2019optimization,meneses2015acrossentropy,lima2008anuclear}, while hybrid methods employing Quantum Evolutionary Algorithm (QEA) and Max-Min Ant System (MMAS) are utilized coupled with Tabu Search (TS) in \cite{wu2016quantum} and \cite{lin2014themaxmin}, respectively. This list is however non-exhaustive.

Other classes of methods leverage Deep Learning from which Deep Reinforcement Learning (RL) also originates. Deep learning was experimented in \cite{erdogan2003apwr,li2022comparative} for PWR or \cite{ortiz2004using} for BWR, in which the authors utilized pre-trained neural networks to reduce the computational burden of evaluating a pattern candidate in their algorithms. The deep models can take as input the composition of what is contained inside a nuclear reactor, specifically the FAs. For instance, two important parameters namely the $k_{eff}$ (the effective neutron multiplication factor, which is a measure of the change in the fission neutron population in a multiplying system with leakages) and the power fraction are predicted by a single layer neural network in \cite{erdogan2003apwr} from the fuel assembly distribution in the core, which was trained with 2000 randomly generated LPs. Then a GA leverages this surrogate to evaluate candidates during the optimization of the Almaraz Power Plant in Spain. This resulted in the reduction of the computing time by a factor of 60. Nonetheless, there are several limitations in these papers: the Figures of Merits (FOMs) utilized for training are the only ones possible for optimization, which limits the surrogate application \cite{erdogan2003apwr}. Furthermore, it is known that after using a surrogate, the best pattern found must be hand-optimized because of the surrogate modeling discrepancy \cite{park2009multiobjective}. Very recently in \cite{li2022comparative}, two neural networks were pre-trained on almost 140,000 data points with almost 60,000 validating data points to predict $k_{eff}$ and another important parameter the power peak factor (ppf), respectively. The neural network-based optimization (accompanied with various heuristics such as GA or PSO) only spent 0.7\% of the computing time compared with using a reactor physic code. In spite of checking the accuracy of the trained model on new data points generated during the optimization to ensure the soundness of the model developed (which remained low), it took about 98 and 120 hours to pre-train the networks. Nevertheless, many more parameters must be considered for a real-world loading pattern optimization. This large training time and low accuracy on test data are evident impediments to seamless scaling to more FOMs. 

Instead of replacing the core simulator, screening procedures could be utilized to improve the search process efficiency. In \cite{yamamoto2003application} a dynamically trained Neural Network was partially used as a screening procedure by evaluating a subset of the FOMs to screen out poor quality patterns accounting for errors in the surrogate, while a GA drives the search. In \cite{gocalvez2006sensitivity}, a neural network and two GAs are coupled to solve 
LP design and the Burnable Poison (BP) assignment for the TMI-1 core. BPs are utilized in FAs to control the excess of power during the early operational cycle of a nuclear reactor before refueling. The Beginning of Cycle (BOC) $k_{\infty}$ (the infinite neutron multiplication factor, which is a measure of the change in the fission neutron population in an infinite multiplying system (e.g., reflected assembly) from one neutron generation to the subsequent generation) for each assembly is used to predict core parameters for the LP design problems, while BP design data are also used for the BP design problem. Two networks are trained for each problem, one being utilized as a fine filter while training only on solutions with higher fitness. This substantially increases the number of good candidates' pattern evaluations but is limited by the target FOMs trained to be represented by the surrogate, and the network has to be somehow dynamically retrained with the generated candidate to avoid failed extrapolation and poor accuracy. 

On the other hand, Deep RL has been applied to solving CO Problems somewhat differently. For instance, in \cite{bello2016neural,khalil2017learning,li2021deep}, RL learning agents are optimizing by taking sequential decisions to build a candidate solution while receiving information (also known as reward signals) on the quality of solutions that it produces. The decisions could be which city to travel to next, and the reward be the length of a partial tour in case of the Traveling Salesman Problem (TSP). The trained network effectively learns the anatomy of a class of problems (e.g., all possible TSP with N-city distribution in $([0,1]^2)^N$ with 20 to 100 cities \cite{bello2016neural}), and can solve the ones with similar underlying structures in a matter of seconds \cite{li2021deep}. The network does not replace the computer program as utilized by the works mentioned in representing the pattern but infers how to construct the most optimal one. The neural networks output a set of action values rather than FOMs, which translates which actions to take to maximize the reward or objective function, which itself embeds the FOMs. Moreover, the network is learning a screening procedure online, while other methods require the careful preparation of datasets in advance. Previous works related to core optimization utilizing RL include \cite{lima2008anuclear}, in which Q-learning was leveraged to inform a “global updating rule” in an Ant Colony System (ACS), and in \cite{nissan1997upgrading} to automate revision rule-sets translated into a procedural language that a recurrent neural network can produce. However, the closest application of deep RL to core optimization can be found in \cite{radaideh2021physics} where the authors made the parallel between optimizing a nuclear fuel assembly and the classical application of RL to games, in which fuel designer tactics are implemented as rules of a game RL agents are trying to solve. With the help of expert rules, the authors found that RL outperforms classical heuristics methods such as GA and SA when the problem becomes large and challenging, which is the case for core optimization as well.

In this work, agents are trained to build a LP for PWR optimization. The agents will sequentially construct different LPs by choosing the arrangement of FAs in the core, as well as their compositions. The reward will be based on an objective function that the core designer aims to maximize: the objective function will be a scalar value which contains the LCOE and key safety and operational constraints. SIMULATE3 \cite{rempe1989simulate}, a reactor physic code with licensing grade accuracy, will be utilized to obtained the figure of merits of interests. This paper studies the set of hyper-parameters that enable the agents to uncover high-quality LPs from the ground up in a reasonable amount of time. The main novelties of this work are listed as follows:
\begin{itemize}
    \item 
    Propose a novel and innovative Deep RL-compliant framework to optimize large-scale constrained problems and employ it to the core loading pattern optimization, becoming the first application of deep RL to this problem.
    \item
    Provide a parameter sweep to tune hyper-parameters for RL, an intuition behind the behavior in response to a change in these parameters, and understand the role of \textit{exploration/exploitation} trade-off in the defined problem.
    \item 
    Utilize the Levelized Cost of Electricity (LCOE) to rigorously assess the true economic performance of the different algorithm configurations. 
\end{itemize}

The framework developed in this work is an extension of a work presented at the PHYSOR conference held in Pittsburgh in 2022 \cite{seurin2022pwr}, and is currently being utilized by Constellation (formerly known as Exelon) Corporation, 
a leading U.S. nuclear electric power generation company. There, a simple hyper-parameter study applied to our problem was conducted. Nevertheless, here we prolong this study by deriving a problem-agnostic approach to obtain a sample efficient method to apply RL, which also extends to any engineering design problem. We however only deal with single-objective optimization. A multi-objective version of this work is also presented at the International Conference on Mathematics and Computational Methods Applied to Nuclear Science and Engineering (M\&C 2023) \cite{seurin2023pareto}. 

In section \ref{sec:reinforcement} the theory of RL with the motivation behind the development of Proximal Policy Optimization (PPO) (section \ref{sec:solvingrlproblem}) as well as its major hyper-parameters (section \ref{sec:ppo_hyperparameters}) are given. Related works are presented in section \ref{sec:relatedwork}. In section \ref{sec:rlalgorithmicassessement}, the formulation of the loading pattern optimization in an RL-compliant setting and the description of the problem which will be used to illustrate the efficacy of this novel method are presented. The results are given in section \ref{sec:results} with a summary of the findings in this work in section \ref{sec:shortsummary}, and an application of the method derived is applied to classical benchmarks in section \ref{sec:applicationtobenchmark}. Lastly, section \ref{sec:conclusion} will conclude this work and present guidance for future studies.

\section{Reinforcement Learning} 
\label{sec:reinforcement}
\subsection{Definition}
\label{sec:defin}
Neuro-Dynamic Programming or Deep Reinforcement Learning (RL) \cite{bertsekas1996neuro} is a class of Machine Learning algorithm that emerged from exact dynamic programming methods for high dimensional problems that suffer from daunting computational requirements. The RL framework is characterized by a trial and error scheme between the generation of experiences by one or several agents taking decisions (or actions), and the improvements of these decisions over time, with the help of a reward signal produced by an emulator for games or an objective function in case of optimization. Figure \ref{fig:rlscheme} summarizes the procedure:
\begin{figure}[H]
    \centering
    \includegraphics[scale = 0.8]{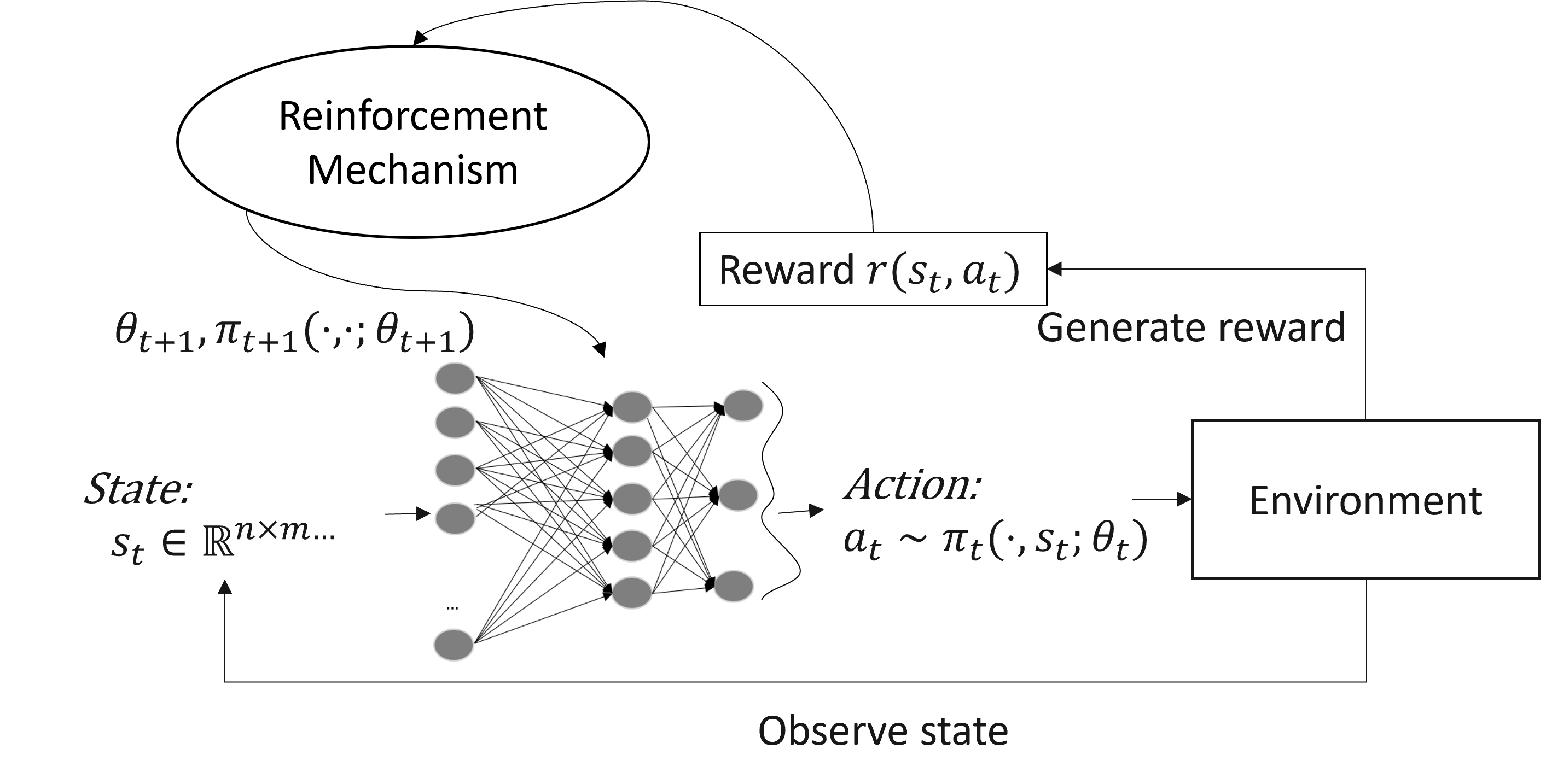}
    \caption{Classical RL iteration procedure. The policy is computed through a neural network in the figure. Each agent is following the same policy. The reinforcement mechanism is the algorithm (here PPO) that updates the parameter $\theta$ based on the agents' experiences.}
    \label{fig:rlscheme}
\end{figure}
One or several agents in parallel are trained to take a sequence of actions $a \in \mathcal{A}$ from observable states $s \in \mathcal{S}$ following a policy: $a \sim \pi(\cdot,s;\theta_k)$, where $\theta_k$ is a learnable parameter to improve this sequence. A policy $\pi = (d_1,...)$ is a sequence of decision rules $d_i : \mathcal{S} -> \mathcal{A}$, which represents the probability of taking action $a \in \mathcal{A}$ being in state $s \in \mathcal{S}$. It specifies which action to take when in state $s$. Each action $a \in \mathcal{A}$ taken after observing a state $s \in \mathcal{S}$ generates a reward $r(s,a) \in \mathbb{R}$ or $r(s,a) \in \mathbb{R}^N$ in case of multi-objective RL. The number of actions taken, also named the length of an episode or the horizon, is denoted $\tau$. Over the course of an episode, a history $\mathcal{H} = (s_0,a_0,...,s_{\tau})$ is generated, on which the sum of rewards $\sum_{t = 0}^{\tau} r_t(a_t,s_t)$ must be maximized. The quality of the sequence is then defined through a value- or state-action value function $V^{\pi_{\theta}}(s;\theta) = \mathbb{E}_{\mathcal{H}}(\Sigma_{t = 0}^{\tau} r(s_t,\pi_{\theta}(s_t)))$ or $Q^{\pi_{\theta}}(s,a;\theta) = \mathbb{E}(\sum_{t \ge 0} r_(s_t,a_t)|s_0 = s,a_0 = a,a_t = \pi_{\theta}(s_t))$ (where $V^{\pi_{\theta}}(s_0;\theta) =  Q^{\pi_{\theta}}(s_0,\pi_{\theta}(s_0);\theta)$) respectively. The value function $V^{\pi_{\theta}}$(s) is the expected sum of rewards following a certain policy $\pi_{\theta}$ being in state $s$, while the action value function $Q^{\pi_{\theta}}(s,a)$ is the expected sum of rewards taking action $a$ being in this state going forward.  If the optimum policy $\pi^{\star}$ is found, following it would result in the highest value function. The classical formulation of the RL problem starting from an initial state $s_0$ then becomes:
\begin{align*}
    & \text{$\max_{\theta}$} \quad \mathbb{E}_{\mathcal{H}}(\Sigma_{t = 0}^{\tau} \gamma^t r(s_t,a_t)) =   Q^{\pi_{\theta}}(s_0,\pi_{\theta}(s_0);\theta) = V^{\pi_{\theta}}(s_0;\theta)\\
    & \text{subject to} \quad a_t = \pi(s_t;\theta)
\end{align*}

where $\gamma \in [0,1]$ is a hyper-parameter added to favor short-term reward if closer to 0, which translates the appeal that early returns are more important than long-term ones (e.g, early return on investment are more attractive to investors). Additionally, it helps make the problem mathematically sound (e.g., prevents the cumulative reward to explode). Deep RL will refer to RL methods that utilize deep learning architectures to represent the value function and the policy. We will use the RL notation to denote both RL and deep RL interchangeably. Therefore $\theta$ could represent the weights in a neural network. The power of the RL mechanism lies in the possibility to exceed expert knowledge by training a policy based on a reward signal \cite{bengio2018machine}. Nevertheless, if the policy is poorly parameterized, or if the experiences are 
not representative of the search space, the policy will perform inadequately \cite{bengio2018machine}. The latter is called the \textit{exploration/exploitation} trade-off. A sufficient amount of \textit{exploration} is necessary for the agents to get a proper representation of the search space, such that they can properly optimize their decisions during the \textit{exploitation} phase. 
\subsection{Theory behind solving an RL problem}
\label{sec:solvingrlproblem}
There are two types of algorithms to solve this optimization: \textit{value iteration} and \textit{policy iteration}. The former class of methods benefited from many improvements \cite{hessel2018raindbow} and is widely used, but they were inferior to policy-based algorithms in particular Proximal Policy Optimization (PPO) \cite{schulman2017proximal} in our application. \textit{Policy iteration} is based on optimizing the policy directly by a parameterization $\pi(\cdot,\cdot |\theta)$. The problem to solve becomes 
\[\max_{\pi(\cdot | \theta) \in \Pi} V(\pi(\cdot|\theta))\]
where $\Pi$ is the manifold of smoothly parametrized policies.

Policy-based algorithms have the advantage of simplicity by having fewer hyper-parameters to tune and also scale well for parallel computing. There are three types: tabular policy iteration, derivative-free optimization (e.g., covariance matrix adaptation (CMA), Cross Entropy Method (CEM)...), and policy gradients \cite{schulman2017trust}. Tabular policy-iteration-based approach cannot be leveraged due to the size of our search space. Derivative-free optimization approaches suffer from low sample complexity meaning that it requires more samples to collect (i.e, objective function evaluation) to obtain satisfying result and they do not behave well for large problems \cite{schulman2017trust}. This was one of the motivation for introducing advanced sample efficient (defined in \cite{botvinick2019reinforcement} as "The amount of data required for a learning system to attain any chosen target level of performance") gradient-based methods which already demonstrated superiority on supervised learning tasks in many instances \cite{schulman2017trust}.
In the original policy gradient method, the parameter $\theta$ is updated by policy gradient ascent with the relation:
\begin{equation}
    \theta_{t+1} = \theta_t + \eta_t \nabla_{\theta}V(\pi(\theta))
    \label{eq:policygradientupdates}
\end{equation}
where $\nabla_{\theta}V(\pi(\theta)) = \mathbb{E}_{\mathcal{H}}( \Sigma_{t = 1}^{\tau}R(\tau) \nabla_{\theta} \log \pi_{\theta}(a_t|s_t)))$. This formulation translates the fact that directions with high expected returns $R(\tau) = \sum_{t = 0}^{\tau - 1} \gamma^{'}r_t$ or $R(\tau) = \sum_{t^{'} = t}^{\tau - 1} \gamma^{t^{'}-t}r_{t^{'}}$ will be pushed with greater magnitude. The estimate of the gradient can be obtained by Monte-Carlo sampling over multiple trajectories, which led to classes of methods called REINFORCE \cite{williams1992simple}. This method suffers however from high variance, high sensitivity to step sizes, sample inefficiencies, and unstable learning \cite{wu2017scalable}.  One alternative to reduce the variance is to replace $R(\tau)$ with the advantage function $A_{\pi_{\theta}}(s,a;\theta) = Q^{\pi_{\theta}}(s,a;\theta) - V^{\pi_{\theta}}(s;\theta)$, which represents the additional expected return by taking action $a$ compared to the average expected return by taking random actions. 

Nevertheless, these methods have lots of shortcomings. Therefore much research on alternative gradient techniques and improved loss functions were conducted. This yields a set of new policy gradient paradigms with Conservative Policy Iteration \cite{kakade2002approximately} and Natural policy gradient \cite{kakade2001anatural}, which led to Trust Region Policy Optimization \cite{schulman2017trust} and finally Proximal Policy Optimization (PPO) \cite{schulman2017proximal}. 

PPO was developed as an improvement of TRPO. PPO is a trade-off between simplicity of implementation, tuning, and Sample Efficiency (SE). It is a major concern for RL, as the amount of data in real world problems is usually scarce per unit of computing time. This is the case in this work because of the slow time of the physics simulation software. Therefore the optimization techniques developed are primarily aimed at increasing this SE \cite{wu2017scalable}.  For this purpose, PPO relies on a clipping term and a learning rate to scale up to massive parallelism, which TRPO was struggling to do \cite{schulman2017proximal}. The PPO version utilized in this work is the first-order constraint-free clipping loss method, where the clipping plays the same role as the KL divergence in TRPO \cite{schulman2017trust}, ensuring that the policies between updates are close enough to each other and avoid catastrophic losses of performance. For the latter, if $r_t(\theta) = \frac{\pi_{\theta}}{\pi_{\theta_{old}}}$ is the \textit{probability ratio}, the new surrogate objective (presented as a loss) to  optimize is a clipped objective:
\begin{equation}
L^{clip}(\theta) = \mathbb{E}(\min(r_t(\theta)A(\theta),\text{clip}(1 - \epsilon,1 +\epsilon, r_t(\theta))A(\theta)))
\label{eq:clipped_obj}
\end{equation}
where $\epsilon$ is an hyper parameter, which guarantee to penalize large policy updates. Indeed, it prevents the probability ratio to lies outside $[1-\epsilon,1+\epsilon]$. Moreover, as the advantage estimation $A_{\pi_{\theta}}(s,a;\theta)$ was used, another value loss $L^{VF}(\theta) = (V^{\pi_{\theta}}(s_t;\theta) - V_{target}(s_t))^2$ was introduced to ensure a proper approximation of the value function (which is used in the Generalized Advantage Estimation (GAE) to estimate $A_{\pi_{\theta}}(s,a;\theta)$ \cite{schulman2017proximal}). Lastly, an entropy term $H(\theta)$, which corresponds to the entropy of the policy distribution, was introduced to enhance exploration. The final loss becomes:
\begin{equation}
    L^{final}(\theta) = \mathbb{E}(L^{clip}(\theta) - c_1L^{VF}(\theta) + c_2 H(\theta))
    \label{eq:ppoloss}
\end{equation}

where $c_1$ and $c_2$ are hyper-parameters (given by \code{vf\_coef} and \code{ent\_coef} in stable-baselines \cite{stable-baselines}, respectively).

\subsection{Proximal Policy Optimization hyper-parameters}
\label{sec:ppo_hyperparameters}
To understand the role of the hyper-parameters studied in appendices \ref{sec:ppohyperparameterstarts} and \ref{appendix:appendixalternate} and summarized in section \ref{sec:shortsummary}, we first need to understand how PPO works. PPO \cite{schulman2017proximal} is based on policy improvement by pessimistic lower bound maximization. It is built on top of TRPO and relies on a clipping term and a learning rate, while TRPO relies on a complex second-order method to ensure policy improvement, as discussed in section \ref{sec:solvingrlproblem}. A pseudo code for the classical PPO training is detailed in algorithm \ref{alg:ppopseudocode}.

A set of hyper-parameters is first defined and the environment (e.g., reward, state, and action spaces), as well as the learning parameters, are initialized (e.g., weights of a neural network). The algorithm runs as long as a termination criterion has not been reached (e.g., number of generations). PPO is executed in parallel with \code{ncores}. Before a parameter update, each agent (corresponding to each core) generates \code{n\_steps} trajectories by taking actions and observing states (lines 7 to 13 in algorithm \ref{alg:ppopseudocode}). The notation \code{NF} stems from Numframes and corresponds to the length of an episode, which will be explained in section \ref{sec:mdp}. Then the different inputs to the PPO surrogate objective must be evaluated. The cumulative reward $R(\tau)$ is evaluated (line 14) to calculate the value function loss, and the variance-reduced advantage-function estimators $\hat{A}$ is estimated for the clipped portion of the objective (line 15). Then the networks' parameters of the policy ($\theta$) are updated thanks to a Monte-Carlo estimate of the surrogate loss (lines 16 to 18). Then, a counter k is incremented and if a pre-defined condition is reached (e.g, k equal to 40,000 iterations) \code{done} is set to \code{True} and the algorithm terminates (lines 19 to 23). Then the last policy is returned (line 24). 

\begin{algorithm}[H]
  \small
    \caption{Pseudocode for Proximal Policy Optimization (PPO)}
    \label{alg:ppopseudocode}
    \begin{algorithmic}[1]
     \State \textbullet Set algorithm's hyper-parameters (\code{n\_steps},\code{ent\_coef},\code{VF\_coef},\code{noptepochs},$\epsilon$,$\gamma$,\code{LR},$\lambda$, \code{nminibatches},\code{max\_norm}, \code{condition})
     \State \textbullet Initialize the environment (i.e., input variables, reward, action, and observation space)
     \State \textbullet Initialize the neural network parameters
     \State \textbullet Set \code{done} = \code{False}
     \State \textbullet Set $k = 0$ 
     \While{not \code{done}}
             \For{Agent $j = 1$,..., \code{ncores}}
                 \For{p = 1,..., \code{n\_steps}}
                    \State \textbullet An action is sampled from the current policy $a_p \in \mathcal{A} \sim \pi_{\theta_k}$
                    \State \textbullet Action $a_p$ is  evaluated by the environment and a reward $r_p$ is returned
                    \State \textbullet A new state $s_p \in \mathcal{S}$ is observed and the state is set to $s_p$
                \EndFor
            \EndFor
        \State \textbullet The cumulative reward $R_k$ is evaluated with $\tau$ equal to \code{ncores} $\times$ \code{NF} and $\gamma$
        \State \textbullet The advantage function $\hat{A}_k$is estimated via GAE with $\lambda$ and $\gamma$
        \State \textbullet The neural network parameters ($\theta_k$) are updated by stochastic gradient ascent (for \code{noptepochs} steps, with learning rate \code{LR}, maximum value for the gradient \code{max\_norm}, with \code{nminibatches} batches) such that:
        \[
    \theta_{k+1} \in arg\max_{\theta} \frac{1}{\text{\code{ncores}} \times \code{n\_steps}}\sum_{i = 1}^{\text{\code{ncores}}}\sum_{\tau = 0}^{\code{n\_steps} - 1}( \min(\frac{\pi_{\theta}(a_{\tau,i}|s_{\tau,i}))}{\pi_{\theta_{k}}(a_{\tau,i}|s_{\tau,i})}A^{\pi_{\theta_k}}(a_{\tau,i},s_{\tau,i}),g^{*}(\epsilon,A^{\pi_{\theta_k}}(a_{\tau,i},s_{\tau,i}))\] 
    \[- \code{VF\_{coef}} (V^{\pi_{\theta}}(s_{\tau,i};\theta) - R_i(\tau))^2) + \text{\code{ent\_coef}} \times H(\theta) \]
  \State \textbullet $k = k + 1$ 
  \If{\code{condition}}
  \State \textbullet Set \code{done} = \code{True}
  \EndIf
    \EndWhile
    \State \textbullet  Return $\pi_{\theta_{N_{gen}}}$
\end{algorithmic}
\end{algorithm}
where $g^{*}(\epsilon,A) = \{ \begin{array}{l}
    (1 + \epsilon)A \quad \text{if} \quad A \ge 0\\
    (1 - \epsilon)A \quad \text{if} \quad A < 0 
  \end{array} $.

The most influential hyper-parameters will be studied in appendix \ref{sec:ppohyperparameterstarts}. The first hyper-parameter is the shape of the reward. The reward drives the search towards an influential area of the search space. It is going to be studied in appendix \ref{sec:weights_study}. Another hyper-parameter is the length of the episode $\tau$ or \code{NF} in this work. Its role in the \textit{exploration/exploitation} trade-off will be unraveled in appendix \ref{sec:nf_study}. A higher value will render the learning more difficult as more loading design must be optimized, but it might allows the agents to explore more how the search space behaves. Another fundamental parameter is \code{n\_steps}, which corresponds to the number of actions each agent will perform before one optimization step. Thus the value obtained by multiplying \code{n\_steps} with the number of agents is the number of steps before $\theta$ is updated. It controls, therefore, the estimation of the gradient of the PPO loss. This parameter has a large influence on the quality of the best pattern found as it plays a major role in the \textit{exploration/exploitation} trade-off of RL by controlling the estimation of the gradient of the PPO's loss. The last hyper-parameter is \code{ent\_coef}. The latter plays a role in the \textit{exploration/exploitation} trade-off as it influences the contribution of the entropy term in equation \ref{eq:ppoloss}. A lower value will increase the exploitation, which will be studied in appendix \ref{sec:entcoef_study}.

We have seen in this section the main motivation behind the development of PPO as well as the major hyper-parameters that characterize it. The question then becomes how can we apply it to solve our problem? The first step is to analyze how RL was applied to similar problems in the realm of CO, which is the topic of section \ref{sec:relatedwork}.

\section{Related works and classical applications of RL to CO problems}
\label{sec:relatedwork}

Deep RL has been applied to solve optimization for more than two decades \cite{bertsekas1996neuro}. Nevertheless, the application to large-scale problems specifically large-scale CO \cite{mazyavkina2021reinforcement} is more recent, and we will focus on the latter literature for the review, as loading pattern optimization falls into this class. CO problems are by essence extraordinarily  hard to solve, in that they are NP-hard problems. For instance, solving the TSP with Exact methods requires dynamic programming. The global solution is obtained in $O(n^22^n)$ (called Held-Karp algorithm \cite{bertsimas2008introduction}), much lower than $n!$ obtained by enumerating all possible permutations. It is still very high, and preemptive from applicability for real-world large-scale problems. Deep learning and in particular deep RL are frameworks that do not work with analytical representation of an objective function but with objects (e.g., function approximator) that are constructed statistically with a large number of data and then mathematically optimized. The following characteristics are generally necessary to solve the CO problems:
\begin{enumerate}
    \item 
    An objective function and a reward associated (e.g., sum of partial costs or differences between two solutions)
    \item
    A neural architecture (e.g., Feedforward, Graph, pointer networks...)
    \item
    A decision or MDP formulation (either a partial solution or a complete solution is available at each step)
    \item
    An algorithm used to solve the problem (e.g., A3C, DQN...)
\end{enumerate}

A large literature exists \cite{mazyavkina2021reinforcement,li2021deep}, in which authors are solving different canonical problems (e.g., Traveling Salesman (TSP) or Knapsack Problem (KP)). In \cite{khalil2017learning}, the authors are using a Graph embedding called structure2vec \cite{dai2016discriminative}, Deep Q-learning, and fitted Q-iteration to solve typical CO problems: (MVC, MAXCUT, and TSP). Expressing these problems as a Graph, they are learning which node (e.g., a city) is going to be selected for each new action (decision) and the reward is the sum of the partial tour formed at each step. In \cite{bello2016neural}, they are leveraging a policy gradient with critic to train a recurrent policy network (in particular a pointer network \cite{vinyals2015pointer}) that predicts the optimal tour in the TSP with the reward being the negative total tour length  and the item distribution for the Knapsack. Similarly, in \cite{nazari2018deep}, they are solving the TSP vehicle routing by policy gradient (REINFORCE \cite{williams1992simple} and A3C \cite{mnih2016asynchronous} for the stochastic version of VRP), with a simplified version of the pointer network. A mixture of Monte-Carlo tree search and deep RL is proposed in \cite{xing2020solve} to solve the TSP, where the graph defined by the cities is represented by the tree, which policy is updated depending on a value function trained by RL. There, the reward is equal to $r(s_{t+1}) = c(s_t) - c(s_{t+1})$, the difference between the previous solution and the new one.  Based on the same idea, \cite{emami2018learning} utilized policy gradient for solving problems involving permutation but does not yet generalize well, while \cite{kool2018attention} is utilizing it with graph embedding instead of a pointer network to learn good heuristics for the TSP and VRP on different instance sizes. In these works, the weights are trained to generate solution end-to-end of unseen instances of problems at inference rather than restarting the training every time a new instance is studied \cite{mazyavkina2021reinforcement}. However, the end-to-end inference strategy in \cite{li2021deep} is not implementable when the objective function is time-consuming to evaluate \cite{solozabal2020constrained}, which is the case for the problem at hand \cite{seurin2022pwr}.

Another different approach includes \cite{delarue2020reinforcement}, which combined value-based iteration with Mixed-Integer-Programming, which exploits the combinatorial structure of the problem at hand, to define a deterministic policy that is evaluated by solving the optimal Bellman equation applied to the Capacitated Vehicle Routing Problem (CVRP). They compete with state-of-the-art RL approaches but are only solving one problem at a time, which contrasts to the previous works cited where multiple instances are solved at once \cite{mazyavkina2021reinforcement}. Their idea is also limited to the existence of a functional form of the problem, which cannot be of use in this work \cite{seurin2022pwr}.

However, the closest application of deep RL to core optimization can be found in \cite{radaideh2021physics} where the authors made the parallel between optimizing a nuclear fuel assembly and the classical application of RL to games, in which fuel designer tactics are implemented as rules of a game RL agents are trying to solve. With the help of expert rules, the authors found that RL outperforms classical heuristics methods such as GA and SA when the problem becomes large and challenging, which is the case for core optimization as well. In \cite{radaideh2021large}, the authors coupled RL and Evolutionary Strategy (ES) to solve the multi-zone BWR bundle optimization. RL was pivotal for narrowing down the search space in high-quality regions to inform ES with \textit{candidate} solutions, where ES then diversified the search, improving the number of feasible candidate designs from 14 (RL standalone) to 82. In these works, however, no study of the significance of the RL hyper-parameters was conducted and it is unclear why RL behaves better than SO-based approaches. They would therefore benefit greatly from an understanding of the behavior of the latter to be able to apply RL optimally for large-scale constrained engineering problems.

Following up on these promising works, we derived a framework in which agents are trained to build a LP for PWR optimization. The agents will sequentially construct different LPs by choosing the arrangement of FAs in the core, as well as their compositions. The reward will be based on an objective function that the core designer aims to maximize: the objective function will contain the LCOE and key safety and operational constraints. This approach differs from heuristic-based methods, in which actions would be taken randomly, while here the actions depend on the understanding of the objective space from the agents. Such added physical insights will likely lead to better solutions.

The next section \ref{sec:rlalgorithmicassessement} will derive an RL-compliant formulation of the loading pattern optimization problem, define proper action and state spaces, present the experimental setup or the core design studied, as well as the reward for the problem at hand,.
\section{RL algorithmic's assessment on the core optimization problem}
\label{sec:rlalgorithmicassessement}
\subsection{Markov Decision Process (MDP) for the core loading pattern optimization}
\label{sec:mdp}
The loading pattern optimization problem can be solved by casting the problem with an appropriate formulation. One such formulation is the Markov Decision Process (MDP). A (discounted) MDP is a tuple $\mathcal{M} = (\mathcal{A},\mathcal{S},\gamma,p,r)$, where $\mathcal{A}$ is the action space, $\mathcal{S}$ is the state space, $\gamma$ is a discount factor, $p$ is a transition probability with Markov Property:  $p(s_{t+1}|s,a) = p(s_{t+1} |s = s_t,a = a_t)$ (each state depends only on the previous state). An MDP generates a history $\mathcal{H} = (s_0,a_0,...,s_{\tau},a_{\tau})$, where $\tau$ is the episode length. A policy $\pi = (d_1,...)$ is a sequence of decision rules $d_i : \mathcal{S} -> \mathcal{A}$. Solving the MDP corresponds to finding a policy $\pi^{\star}$ that solves $\max_{\pi} V^{\pi}(s_0)$ as described in section \ref{sec:solvingrlproblem}. In the literature for solving canonical CO problems such as the Traveling Salesman Problem (TSP), the Maximum Cut Problem (Max-Cut), the Bin Packing Problem (BPP), the Minimum Vertex Cover (MVC), or even the Maximum Independent Set (MIS), the recipe for casting the problem in an MDP are similar \cite{mazyavkina2021reinforcement}. The state space $\mathcal{S}$ represents a full or partial solution, an action $a$ adds a node to the partial solution or perturbs a sub-optimal one, the reward is directly related to the cost function characterizing each problem (e.g., the cost of a partial tour or the difference between two tours in case of the TSP), and the horizon is the number of nodes to construct a solution or arbitrary stopping criteria if complete solutions are perturbed at each step. The mapping between state and action is enabled by the use of an encoder such as a recurrent neural network (RNN), a Graph Neural Network (GNN), or a Multi-Layer Perceptron (MLP), with sometimes the help of attention mechanisms. See section \ref{sec:relatedwork} for practical examples.

For our purposes, an action is constructing one LP. The action space, therefore, depends on the number of FAs available, and the complexity of the problem (i.e., decision variables including shuffling, rotation, and pure assignment). The state space is the space represented by all possible LPs achievable from our choice of fresh fuels and existing burned assemblies. A state must correspond to a unique core \cite{erdogan2003apwr}. Consequently, the state of a core will be augmented by a feature embedding of the physics characteristic of each core generated (e.g., $F_{\Delta H}$, BP, and enrichment of each fuel assembly...): $s \in \mathcal{S} = \{s^{i} : i = 1,2,...,N_{FA}\}$, where $s^{i} = \{s^i_1,...,s^i_d\}$, where d is the embedding size of the fuel assembly $s^i$  as will be described in section \ref{sec:description_core}. The mapping between the state and the action is enabled by one-layer feature extractor and two hidden layers with 64 hidden units each fully connected, which is the default fully-connected neural network in stable-baselines.

The $\gamma$ is fixed at 0.99 in this study. The transition probability is either deterministic or stochastic. In the deterministic case, a choice of FAs will always yield the same pattern. For the stochastic case, a choice of action can be complemented by random assignments if the action is prohibited (e.g., a fuel assembly has already been picked). The reward is given by the computation of the objective function derived in section \ref{sec:objandconstr}, which is a testament to the quality of the LP. The episode terminates when a certain number of LPs Numframes (\code{NF}) has been generated. 

\subsection{Description of the Problem within the RL framework}
\label{sec:description}
In this section, the nuclear reactor core studied to apply the method, the objective, and the constraints involved will be described.
\subsubsection{Description of the reactor core, action, and state space}
\label{sec:description_core}
\begin{figure}[H]
    \centering
    \includegraphics[scale = 0.5]{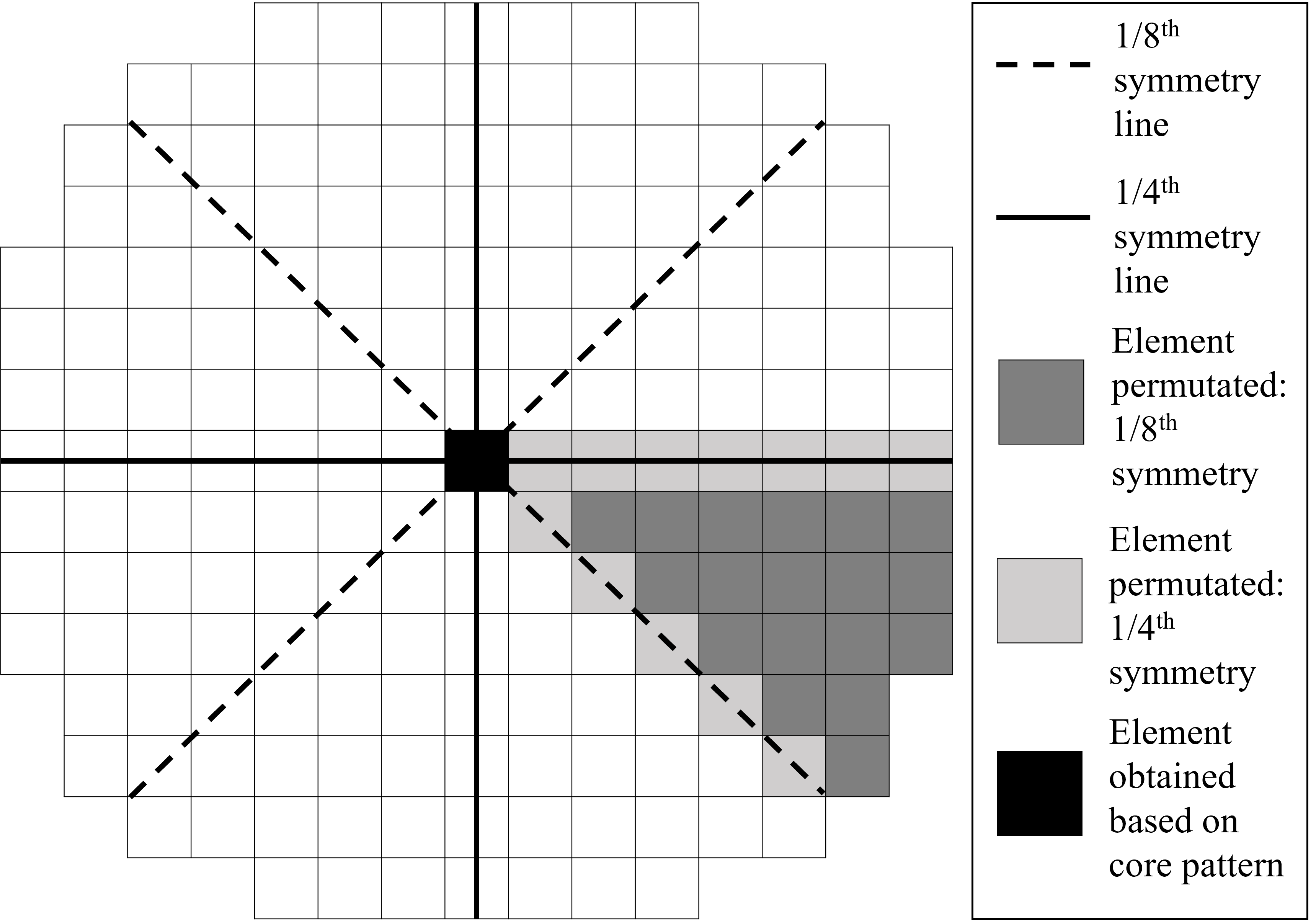}
    \caption{PWR core arrangement and symmetries. Elements within the eighth symmetry (dark gray) are permuted together, while elements on the symmetry lines (light gray) are permuted together. The center element is chosen to have the lowest enrichment and highest BP concentration.}
    \label{fig:pwrcore}
\end{figure}

To demonstrate the performance of the RL algorithms, a loading pattern optimization study is performed for the equilibrium cycle of a 4-loop Westinghouse-type reactor with 193 assemblies, 89 of which are fresh. This means that 89 will remain in the reactor (but potentially at a different positions) for the next operational cycle after refueling, while 16 will remain for three cycles in total. The schematic is shown in Figure \ref{fig:pwrcore}. The equilibrium cycle is obtained by applying the same LP recursively until there is the convergence of the physics parameters. Therefore, we know exactly how much energy is extracted from each FA (also called exposure or burnup) and how much they cost, which is why we can express the LCOE in \$/MWh \cite{kropaczek2011copernicus}. To reduce the search space, mirror one-eighth symmetry boundary conditions are applied. This symmetry is widely utilized for scoping analysis in literature (as in \cite{lima2008anuclear,wu2016quantum,lin2014themaxmin,erdogan2003apwr,ortiz2004using,yamamoto2003application}). 
If the agents never receive high-quality rewards or if it remains flat, they will never learn. To help them receive good rewards more rapidly, certain rules to the loading pattern optimization were applied. Throughout the literature, there are important rules/heuristics enforced. Twice-burned FAs or assemblies with the highest exposure are placed at the core periphery systematically \cite{kerkar2007exploitation,delcampo2004development,castillo2004bwr} which reduces the load on the reactor vessel (called a low leakage pattern) for increasing its lifetime \cite{kerkar2007exploitation,lin2014themaxmin}, and improves the cycle length (equal to the theoretical length of an operational cycle) \cite{wu2016quantum,lin2014themaxmin}, lowering the fuel cycle cost \cite{kropaczek2011copernicus}. It also can lower the BOC excess of reactivity, hence the Moderator Temperature Coefficient (MTC) \cite{wu2016quantum} at the same time. It is however a trade-off with the peaking factor flattening, which will be harder to accomplish. The MTC and peaking factors (e.g., $F_{\Delta h}$) must remain below a threshold to ensure the theoretical safety of the reactor. Some optimization techniques are even solely utilizing heuristics to create a LP \cite{nissan1997upgrading}. Imposing heuristics can have its limitation and for this work, few of them will be enforced, while most will be learned by the RL agents. One typical rule includes putting twice-burned fuel at the periphery and preventing fresh ones from being located there. To even out the power, a ring of fresh FAs between the in-board and the periphery is placed. On top of that, to limit the risk of fuel failure during Reactivity Initiated Accident (RIA), no square of fresh FAs are allowed in the core, which is a typical rule often enforced by the Nuclear Regulatory Commission (NRC). Because the FAs are not homogeneously burned, they are sometimes rotated for the next cycle. However, since the impact of assembly rotations is relatively small \cite{lin2014themaxmin}, it is not included in the current study. Similarly, the compositions of the twice-burned FAs (but not locations) are fixed. The original number of possible patterns in eighth symmetry was $\sim 10^{52}$.  Such rules, based on core design best practices, reduce the number of possibilities to $\sim 10^{29}$.

Each action represents a choice of a template pattern and all the FAs that are going to be placed inside it. The action space for the fresh fuels is constant and depends on the available fresh fuels provided by the user. Here, the fresh FAs (14 choices for eighth symmetry) are chosen from 24 different types that are characterized by their enrichment and BP distribution. In this work IFBA and WABA are utilized as BP. The uranium 235 enrichment varies between 4.00 to 4.95\%, the IFBA and WABA loading are discrete between 128x or 156x, and 12 or 24 guide tubes, respectively. WABAs are placed in water rods in which control mechanisms could be inserted in case of an emergency shutdown of the reactor. Therefore, locations under Control Rods (CRD) cannot hold WABA to not obstruct their insertions. Therefore a randomly sampled fresh fuel without WABA is selected if the FA chosen by the agent for this specific location is forbidden. It could be possible to learn to not take forbidden actions via learning rules as in a game prior to running SIMULATE3 as done in \cite{radaideh2021rule} with CASMO4e. Nevertheless, we found out that it did not penalize the learning to simply replace forbidden by randomly sampling valid ones. The action space must then decrease when choosing once- and twice-burned FAs. To improve the search efficiency we need to prevent random actions to be taken \cite{nijimbere2021tabu}. For this purpose, we adopted the \code{MultiDiscrete} option of stable-baselines to define the action space. There, for each once- and twice- burned location, the action space decreases by 1 for each new action, which prevent forbidden action to be taken.

\begin{figure}[H]
    \centering
    \includegraphics[scale = 0.9]{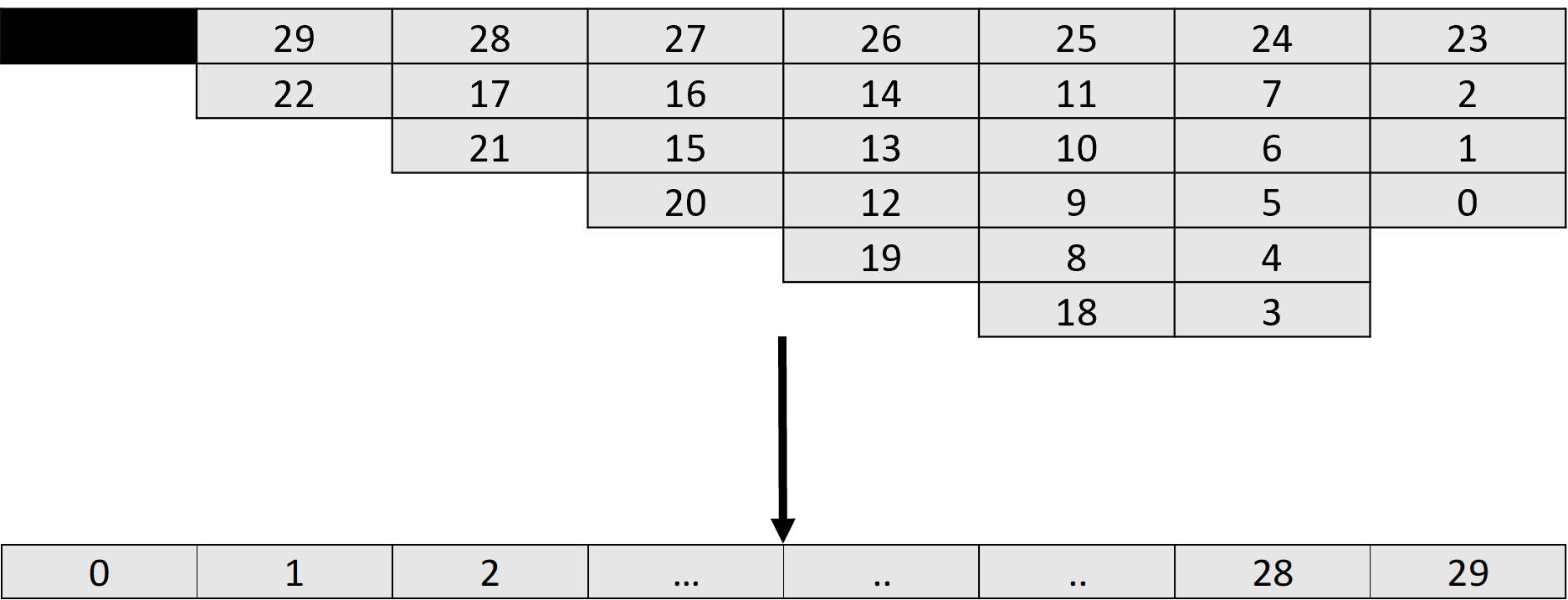}
    \caption{Example of transformation of a 2D core to a 1D array. The FAs in the eighth symmetry appear first in the array, while the element in the quarter symmetry lines appear second starting from the diagonal, due to the particular treatment needed for these locations. The center assembly is blurred as it does not belong to the action space.}
    \label{fig:2dto1d}
\end{figure}

The new action obtained by accounting for the forbidden actions must then be decoded before being evaluated. The decoding is straightforward. Each entry is a location in the 2D core as shown in figure \ref{fig:2dto1d}.

Each entry of the action's vector is an integer, which corresponds to a fuel assembly. The input is spanned sequentially, while the core locations are treated depending on their types and locations as provided in figure \ref{fig:2dto1d}. First fresh fuel locations are filled and then once- and twice-burned locations are treated. For fresh fuel, once-, and twice-burned location j, integer $k^j$ corresponds to the kth element of the list of possible fresh fuel, available once-, and twice-burned FAs, respectively. The list of possible fresh fuel is constant, while the list of available burned assemblies decreases.

Then the state must also be defined. Initially, no pattern were generated and the starting state is an array of zeros. As the algorithm progresses, the starting state at the beginning of each \textit{episode} will be the embedding of the best pattern ever found. The reason is to mimic the way a core designer would iterate on the best pattern found as it progresses toward finding a loading pattern by hand. The parameters chosen for embedding are the relative position of the FAs, their enrichment, their BP's content (IFBA or WABA), the Radial Power Fraction (RPF), the $F_{\Delta h}$, and the pin peak exposure at the End Of Cycle (EOC) for each individual fuel assembly. For an equilibrium cycle, the first 4 fully defined a fuel pattern and the last 3 were added to help the agent. However, no thorough analysis of the influence of embedding has been conducted. The core is then evaluated following the objective function defined in section \ref{sec:objandconstr}.

\subsubsection{Reward construction with the Objectives and Constraints}
\label{sec:objandconstr}
The classical formulation of an optimization problem can be proposed as follows without loss of generality:
\begin{align*}
    &\min_{x} f(x) = [f_1(x),f_2(x),....,f_F(x)]^T \\
    &\text{s.t.} \quad g_i(x) \le 0 \quad \forall i\in \{1,...,N\} \\
     & h_j(x) = 0 \quad \forall j\in \{1,...,M\}\\
     & x = [x_1,x_2,...,x_K]^T \in \mathcal{S} \subset \mathbb{R}^K
\end{align*}
\label{eq:fObj}
\noindent
Here, a single objective optimization problem will be solved, hence $F$ is equal to 1. The first portion of the objective function is the LCOE. The LCOE is computed following the proposed methodology in \cite{saccheri2004atight}, which results from the sum of the costs over different levelized periods or fuel cycles. The levelized period is characterized by the effective-full-power-days (EFPD) or cycle length under which the reactor operates before the temporary shutdown for refueling. The intrinsic characteristics of the fuel assembly loaded after the aforementioned period such as its average fuel enrichment and WABA content are needed to compute the cost for each stage of the fuel cycle: uranium extraction, transport, conversion, enrichment, fabrication, and waste disposal. Another factor is the amount of energy extracted during the lifetime of a fuel assembly or burnup expressed in GWd/tHm. The formula is given in equation \ref{eq:lcoedef}:
\begin{equation}
    LCOE = \frac{EFPY(T_{lev})}{\eta \times K_f(T_{lev}) \times 24} \times \frac{r}{1 - \exp^{-rT_{lev}}} \sum_{i = 1}^{N_{FA}}\alpha_i\frac{1}{Bu_i}\sum_{i_k}C_{ik}
    \label{eq:lcoedef}
\end{equation}
where $K_f(T_{lev}) = K_{av} \times (1 - \frac{T_{FO} + T_{MO}}{T_{lev}})$ is the capacity factor that depends on the fuel lifetime $T_{lev}$ = $\frac{T_{cy}\times n_{batchs}}{365.25}$ (where $T_{cy}$ is the cycle length and $n_{batchs} = 3$ is the number of batches in the core), $K_{av}$ the availability factor, $T_{FO}$ the refueling outage, and $T_{MO}$ the maintenance outage. $EFPY(T_{lev}) = K_f \times T_{lev}$ is the effective full power year,   $\alpha_i$ is the proportion of assembly of type i in the core, and $\sum_{i_k}$ is the sum over the (levelized) component of the cost $C_{ik}$ over the lifetime of the fuel assembly (including the cost of WABA). $C_{ik} = c_{ik} \times \exp^{- r T_{ik}}$, where $c_{ik}$ is the kth overnight cost in $\$/kgU$ of a fuel assembly, r is the discount rate, and $T_{ik}$ the corresponding levelized period. The latter is negative if the stage started before the beginning of the cycle (e.g., fabrication). The cost of WABA is accounted considering that one unit cost 1500 \$, hence the average contribution of each fuel assembly in $\$/kgU$ is $C_{i,waba} = Wab \times 1500 \times \exp^{- r T_{fabric}}\frac{1}{Mass}$, where $Wab$ is the total number of WABA units and $Mass$ is the mass of the core.  Lastly, $Bu_i$ is the burnup of fuel assembly i. 

Nevertheless, the regulatory and operational constraints must also be satisfied but ensuring feasibility is already a hard problem. This study considers the following set of 7 essential constraints for operation and safety. A description of their derivation and significance with regards to safety can be found in the website of the United States Nuclear Regulatory Commission (NRC) (e.g., for peaking factors \cite{nrcXXXX0523}): 
\begin{enumerate}
    \item
	$c_1$: Ensuring a lower limit for the cycle length to respect the refueling outage schedule, where $c_1 \geq 500$ EFPD is chosen as a constraint.
	\item
	$c_2$: An upper limit on the peak pin peaking factor $F_q$, which is a safety parameter measuring the ratio between the maximum local power produced to the average power generated by the fuel rods in the core. It is monitored to limit the fuel centerline temperature and Pellet-Cladding concerns. The constraint is $c_2 \leq 1.85$.
	\item
	$c_3$: An upper limit of the rod-integrated power $F_{\Delta h}$, which is a safety parameter measuring the maximum total power generated by a fuel rod. It is monitored in particular to respect the margin to departure from nucleate boiling ratio. The constraint is $c_3 \leq 1.45$.
	\item
	$c_4$: To help
maintain the core critical, diluted Boron in the coolant is used in PWRs but it must be limited due to safety reasons. In particular, we consider an upper limit of the boron concentration $C_b$ to control the axial offset due to boron deposition and prevent positive temperature coefficient of reactivity. The constraint is set to $c_4 \leq 1200$ ppm.
	\item
	$c_5$: An upper limit of the pin peak discharge burnup $Bu_{max}$ for fuel performance concerns (e.g., hydrogen buildup). The constraint is $c_5 \leq 62$ GWd/tHm (\textbf{G}iga \textbf{W}att \textbf{d}ay per \textbf{t}on of \textbf{H}eavy \textbf{M}etal).
	\item
	$c_6$ and $c_7$: Limits of the number of different enrichment ($c_6$) and BP ($c_7$) patterns in fresh fuel loading batches, respectively. The constraints are $2 \le c_6 \le 3$, $1 \le c_7  \le 3$. The upper bounds correspond to potential constraints imposed by the fuel vendors. The lower bounds are assumed based on expert judgment to enable effective reduction of power peaking in the core.
\end{enumerate}
 
 These constraints are called \textit{instantaneous} \cite{liu2021policy} because they must be satisfied after one action. They are also dubbed \textit{implicit} \cite{liu2021policy} because they are obtained with a 3D nodal nuclear depletion simulation code SIMULATE3 \cite{rempe1989simulate}.
In literature,  three main categories of objective function are utilized namely penalty function-, step-wise or piece-wise-, and ratio-based methods \cite{li2022areview}. Here a penalty-based method was found to work better. The LCOE and the constraints are included in the objective function that needs to be maximized:
\begin{equation}
    \max_{x} f(x) = - LCOE (x) - \sum_{i \in C} \gamma_i \Phi(x_i) + 1 \times \delta_{\forall i, x_i \le c_i}
    \label{eq:NucCoreObj}
\end{equation}
where $C = \{ (c_{i})_{i \in \{1,..,7\}} \}$ is the set of constraints and $(\gamma_{i})_{i \in \{1,..,7\}}$ is a set of weights attributed to each constraint. If $ (x_{i})_{i \in \{1,..,7\}}$ are the values reached for the core ($x$) for the corresponding constraints $(c_{i})_{i \in \{1,..,7\}}$, $\Phi(x_i) = \delta_{x_i \le c_i} (\frac{x_i - c_i}{c_i})^2$, where $\delta$ is the Kronecker delta function. This formulation was deemed appropriate, as it is a single objective constraint problem, and the penalization disappears when the constraints are satisfied. 

The equality constraints are also \textit{instantaneous} but called \textit{explicit} \cite{liu2021policy} because they can be dealt with maskable actions, random replacement, or by the construction of the MDP. We write them in terms of the number of FAs and their locations in the core following the methodology of \cite{kropaczek1991incore}. Let's define $\chi_{L,T,B,E}$ such that:
\[\chi_{L,T,\mathcal{B},\mathcal{E}} = \{\begin{array}{l}
     1 \quad \text{if a fuel assembly located at position L is of burned-type T, with BP } \mathcal{B}  \\
      \quad \text{   and enrichment } \mathcal{E}  \\
     \\
    0 \quad  \text{otherwise} 
\end{array}\]
where $T$, $B$, $E$ are the set of possible burned-type (i.e., fresh, once, or twice), BP, and enrichment, respectively. One and only one assembly must be placed at each location:
\begin{equation}
\forall l \in \{1,...,N_{FA}\}, \quad \sum_{\tau = 1}^3 \sum_{b \in \mathcal{B}} \sum_{e \in \mathcal{E}} \chi_{l,\tau,b,e} = 1
\end{equation}
Then, fresh fuels cannot be placed at the periphery, which can be written in terms of:
\begin{equation}\forall Fresh, \quad \sum_{l} \sum_{b \in \mathcal{B}} \sum_{e 
\in \mathcal{E}} \chi_{l,Fresh,b,e} = 0 \quad l \quad \text{is a periphery}\end{equation}
Additionally, BPs cannot be placed underneath control rods:
\begin{equation}\forall \beta \in B \setminus 1_W \quad \sum_{l} \sum_{\tau = 1}^3 \sum_{e 
\in \mathcal{E}} \chi_{L,T,\beta,e} = 0 \quad l \quad \text{is under a CRD}\end{equation}
where $B \setminus 1_W$ are the FAs with any BP and WABA. There is exactly the same number of fresh and once-burned, one fresh in the center, and four twice-burned per quadrant:
\begin{equation}\sum_{l = 1}^{N_{FA} - 1}  \sum_{b \in \mathcal{B}} \sum_{e 
\in \mathcal{E}} \chi_{l,Fresh,b,e} = \sum_{l = 1}^{N_{FA}}  \sum_{b \in \mathcal{B}} \sum_{e 
\in \mathcal{E}} \chi_{l,Once,b,e} = 88\end{equation}
\begin{equation}\sum_{l = 1}^{N_{FA}} \sum_{b \in \mathcal{B}} \sum_{e 
\in \mathcal{E}} \chi_{l,Twice,b,e} = 16\end{equation}
As discussed in section \ref{sec:description_core}, these constraints are solved by construction.

The RL-compliant formulation of the core optimization via an MDP has been defined in section \ref{sec:mdp}, the input variables were defined in section \ref{sec:description_core}, and objectives and constraints were defined in section \ref{sec:objandconstr}. It is now possible to solve the problem. 

\section{Results and Discussion}
\label{sec:results}

In section  \ref{sec:prerequisite} we provide the metrics and statistical tools required to conduct the study. In section \ref{sec:shortsummary} we summarized the results of the studies conducted in appendices \ref{sec:ppohyperparameterstarts} and \ref{appendix:appendixalternate}. Lastly,  in section \ref{sec:applicationtobenchmark}, we apply the method derived to benchmark functions and demonstrate the viability of the method beyond our loading pattern problem.

 \subsection{Pre-requisites for hyper-parameter tuning}
 \label{sec:prerequisite}
 As mentioned in section \ref{sec:solvingrlproblem} two major parameters coming from PPO derivation include the learning rate \code{LR} and \code{Clip} (the $\epsilon$ in algorithm \ref{alg:ppopseudocode}). Surprisingly, they were found to not influence the optimization's result drastically. \code{noptepochs}, and the depth of the neural architecture were also studied but had very little effect or deleterious effect when getting away from the default values. Therefore only \code{weights}, \code{Numframes}, \code{n\_steps}, and \code{ent\_coef} results will be thoroughly presented in appendix \ref{sec:ppohyperparameterstarts}. The results for the other hyper-parameters are given in appendix \ref{appendix:appendixalternate} but both are summarized in section \ref{sec:shortsummary}. 

To compare the performance of the hyper-parameters of the newly developed MDP two metrics are necessary: (1) the mean objective per episode, which measure how the RL agents are learning, and (2) the max objective obtained for a candidate, which is the goal of the core designer. Another interesting measure is given in \cite{francois2012comparison} the "Optimization Technique Performance (OTP)": as 
\begin{equation}
OTP = \frac{fitness}{\text{number of calls}}
\label{eq:otp}
\end{equation}
which is equivalent to the classical \textit{sample efficiency} (SE) in the RL framework. Nonetheless, in this work, the performance is not only measured in the classical RL sense (i.e., for a system to be able to reproduce a high cumulative reward after training) but also needs to account for the best pattern found. Three equivalent SE measures will therefore be given: One is the Improvement Ratio (IR). The IR is defined as the ratio between the number of samples needed to reach the maximum objective and the total number of samples for the run. A low IR signifies that the agents fail to keep improving the pattern while a high one means that they keep bettering. The other one will be the average mean reward obtained per episode in the latest generation (assuming all experiments generated a similar number of samples, this is equivalent to the SE in the RL sense). A high mean objective SE means that the agents are performing well. Moreover, since the experiments are run over a day and not until convergence, they will be sometimes combined with the $\sigma$SE, which is the standard deviation of the mean reward obtained per episode in the last generation. If the latter is low it means that the agents are close to convergence. We are aiming at seeing both high mean objective SE and $\sigma$SE.  To sum up, assessing the performance of hyper-parameters is a multi-measure problem. It needs:
\begin{enumerate}
    \item 
    A high Max objective function ever found that complies with the designer's interest.
    \item
    A high mean objective SE (at the time the algorithm stops) over multiple generations and a high $\sigma$SE (potential for future improvement) which comply with the RL interest.
    \item
    A high improvement ratio, which means that the algorithm has recently improved and is not behaving sub-optimally.
\end{enumerate}

Additionally, the policy is by essence stochastic hence comparing the results over one single optimization run is theoretically not sufficient. Commonly, a better approach would be based on running N times each algorithm on the same problem (e.g., N equal to 10 in \cite{ivanov2021assessment}, 15 in \cite{meneses2009particle}, 50 in \cite{meneses2015acrossentropy}, or 30 in \cite{francois2012comparison}) and evaluate the best, mean, and standard deviation of the objective over all experiments for each type of algorithms and draw conclusions from it. However, this does not provide the level of certainty at which these conclusions are valid (e.g., how high N should be). The latter issues can be addressed via non-parametric statistical tests, which is a class of popular methods adopted for instance in the world of swarm and evolutionary computations \cite{derrac2011apractical,schlunz2016acomparative}. 
 
 Hypothesis Testing (HT) \cite{casella2002statistical,derrac2011apractical} is a classical inference tool that lies at the basis of the parametric and non-parametric tests. In HT, two hypotheses $\mathcal{H}_0$, the null hypothesis, and $\mathcal{H}_1$, the alternative hypothesis are a partition of the hypothesis space. The goal for the experimenter is to decide whether to accept or reject $\mathcal{H}_0$ after observing a sample based on a \textit{hypothesis test}, which defined the sample region for which the null hypothesis is accepted or rejected. For our purpose, the $\mathcal{H}_0$ could be that the two algorithms are equivalent in performance for pairwise comparison \cite{derrac2011apractical}, and rejecting $\mathcal{H}_0$ would signify that for a particular or a set of problems, one algorithm outperforms the other. Furthermore, in the case of multiple algorithms, the error from drawing a false conclusion from multiple pairwise comparisons may accumulate. The Friedman Test (FT) \cite{derrac2011apractical} is the non-parametric version of the repeated measures ANOVA \cite{casella2002statistical,derrac2011apractical} test, which is used to compare a set of assumptions at the same time. The inputs to the tests for each instance of hyper-parameters will be the value of the max objective and max mean objective per episode for each experiment. This test however only confirms that at least two instances come from different distribution but do not provide conclusion on pairwise tests. Once an FT is successful, post-hoc tests can be performed \cite{derrac2011apractical} to draw the a conclusion between each assumption (e.g., which set of hyper-parameters are the best). Even though it has limited power \cite{derrac2011apractical}, we will use here the Nemenyi Test (NT) due to its simplicity and that it has been applied in a work related to loading pattern optimization in \cite{schlunz2016acomparative}. At most 10 experiments for each variation of the parameters are run (with the same seed for each variation of the parameters studied and 10 seeds in total). The p values will also be provided, which provide quantitative information on how certain are we to reject one or the other hypothesis compared to a level of significance $\alpha$. Roughly, the p-value is the probability of observing an instance at least as extreme as your results if the null hypothesis $\mathcal{H}_0$ was true. The smaller it is, the more likely we are to reject $\mathcal{H}_0$. For a mathematical definition see \cite{casella2002statistical}. The tests are implemented through the scipy.stats \cite{2020SciPy-NMeth} and scikit posthocs \cite{Terpilowski2019} packages for Friedman and Nemenyi, respectively, with $\alpha$ equal to 0.05 as suggested in \cite{schlunz2016acomparative}. 

A previous in-house study led to the initial PPO hyper-parameters: \code{nminibatches} is 4, \code{LR} is 0.00025, \code{max\_grad\_norm} is 0.5, $\epsilon$ is 0.2, \code{n\_steps} is 4, \code{vf\_coef} is 1.0, and \code{noptepochs} is set to 10. These were obtained by changing the hyper-parameters by hand until we had a significant number of feasible patterns within 24 hours of run. Then, all the weights in equation \ref{eq:NucCoreObj} were set to 25,000 except for constraints $c_6,c_7$, for which 1,000 is used. The vectorized state is passed through a feature extractor (one flattened layer) and the extracted features input a feed-forward two-layer perceptron neural network policy with 64 hidden weights per layer.  The time to reach a feasible solution is also of importance. For instance, emergency core re-designs can be limited to 1-3 days because about a million dollars per day is assumed to be lost during the shutdown of the NPP. Therefore, it is important to tune the hyper-parameters to 
 achieve a similar level of speed. As a result, we decided to limit the computing time allocated to our optimizations. For all experiments, PPO agents ran from scratch for 24 hours, which amounted to about 40,000-50,000 SIMULATE3 calls or objective function evaluation on the Massachusetts Green Energy High-Performance Computing Center, when using one node of 2x16  cores (i.e., 32 agents) Intel Xeon 2.1 GHz with 128 GB RAM. Section \ref{sec:shortsummary} summarizes our findings.

\subsection{Summary of the studies conducted}
\label{sec:shortsummary}
In this section, we present a summary of the findings made in this work (from appendices \ref{sec:ppohyperparameterstarts} and \ref{appendix:appendixalternate}). 

We first studied the shape of the objective function via the \code{weights} in appendix \ref{sec:weights_study}.  This parameter behaves as a fudge factor which could result in an improvement on the economic side but also the stability of the learning. Then, we studied the parameter \code{NF} in appendix \ref{sec:nf_study}. This parameter plays a role in the exploration/exploitation trade-off. A higher value will yield a lower mean objective. Due to the significance of the statistical tests, we suggest however, to always choose a value of 1, with the restarted state being the best ever found. This formulation can be thought of as a gradient ascent problem: At each episode, based on the understanding of the curvature of the objective space, the agents are trying to find the steepest ascent direction. We also studied \code{n\_steps}, which is a hyper-parameter that controls the exploration/exploitation by playing with the bias/variance trade-offs in the estimation of the gradient of the PPO loss. A very low value may result in instability of the agents and being stuck in local optima as few samples are exploited to evaluate the gradient. A higher value may result in slow learning and a poorer quality of the best pattern found because more samples will be needed before optimization of the policy network's parameters. Lastly, we studied \code{ent\_coef}, which controls the randomness of the policy during training. It also controls the exploration/exploitation trade-off. Lower values yield faster convergence but are prone to local optima. On the other hand, a high value will result in slower convergence. A sweet spot between the speed of convergence, the quality of the pattern found, and the stability of the learning of the agent resulted in our choices of hyper-parameters and are summarized in table \ref{tab:summaryofhyper}. 

The reader should note that the value of the hyper-parameter \code{n\_steps} which was found to be optimal is excessively different from the default value in stable-baselines (4 against 128). This also underlines the need to perform a thorough evaluation of the hyper-parameters' performance.
\begin{table}[H]
    \centering
        \caption{Summary of the hyper-parameters studied, the corresponding ranges, the default and final chosen value for each, and appendices in which these studies can be found.}
    \begin{tabular}{c|c|c|c|c}
    \hline
    Hyper-parameters & Range Studied & Default in stable-baselines & Best chosen & Corresponding appendix \\
    \hline
     \code{weights}    &  [1,000 - 100,000] & -$^{\star}$ & 25,000 & \ref{sec:weights_study}\\
     \code{NF}    &  [1-100] & -$^{\star}$ & 1 & \ref{sec:nf_study}\\
     \code{n\_steps} & [1-75] & 128 & 4 & \ref{sec:nsteps_study}\\
     \code{ent\_coef} & [0 - 0.01] & 0.01 & 0.001 & \ref{sec:entcoef_study}\\
     \hline
    \end{tabular}
    \label{tab:summaryofhyper}
    \begin{tablenotes}
      \item  \footnotesize$^{\star}$These two hyper-parameters (\code{weights} and \code{NF}) are specific to the core design optimization and not stable-baselines, hence there are no default values for them.
    \end{tablenotes}
\end{table}
We also recall the conclusions of the studies in appendix \ref{appendix:appendixalternate}:
\begin{enumerate}
    \item 
    The default value of \code{LR} equal to 0.00025 demonstrated the optimum trade-off between fast learning with a premature convergence for higher values of \code{LR} but no learning at all for lower values. See appendix \ref{appendix:learningra} for more details.
    \item
    The clipping parameter $\epsilon$ can be very sensitive as a change from 0.02 to 0.01 substantially worsened the performance of the agents as showcased in table \ref{tab:clipping_study}. The default value of 0.02 corresponds to a sweet spot between higher exploration but slow convergence for higher values but slow and unstable learning for lower values. See appendix \ref{appendix:clipstudy} for more details.
    \item
    For the hyper-parameter \code{max\_grad\_norm} the differences are not as visible as for \code{LR} or $\epsilon$. Other values than the default 0.5 could be competitive such as 1.0, 1.2, and 1.5. However, when coupled with the final \code{ent\_coef}, 0.5 demonstrated better SE measures than all of them. See appendix \ref{appendix:maxgradnormstudy} for more details.
    \item
    A value of 10 for \code{noptepochs} was kept. Higher values yielded a higher entropy of the policy network and therefore more random actions were taken. A lower value of 5 exacerbated similar performance in terms of the best pattern found and $\sigma$SE but a lower mean objective SE, while a value of 1 prevented the agents to ever learn anything. See appendix \ref{appendix:noptepochsstudy} for more details.
    \item
    The hyper-parameter \code{nminibatches} can be sensitive and low values (below 4) are preferred. In spite of observing a smooth learning, higher values yielded very low mean objective SE with very high $\sigma$SE. The default value of 4 exacerbated better SE measures compared to 2. See appendix \ref{appendix:nminibatchesstudy} for more details.
    \item
    Reducing the impact of the value function via reducing \code{VF\_coef} seemed to increase the exploitation. Lower value often resulted in lower $\sigma$SE and higher mean objective SE and the trade-off sits between 1.0 and 0.8. The study also revealed that improving the value function loss had more impact on the SE measures than improving the policy one. See appendix \ref{appendix:vfcoefstudy} for more details.
    \item
     Halving the breadth (i.e., the number of hidden nodes per layer) of the policy network systematically decreased the mean objective SE and increased the $\sigma$SE, while the best pattern found is very similar. On top of that, more layers (i.e., increasing the depth) resulted in slower learning but a better average behavior. The default value of 2 was kept thanks to a higher mean objective SE and $\sigma$SE. See appendix \ref{appendix:architecturestudy} for more details.
\end{enumerate}
Overall, we argue that the most sample efficient way to apply RL is as follows: Due to the significance of the statistical tests for \code{NF}, a value of 1 should always be considered, where a pattern is replaced by a problem-dependent input. Then, for a given amount of sample, the hyper-parameter \code{n\_steps} must be decreased to find the lowest value for which learning is observed. Thereupon, the entropy coefficient \code{ent\_coef} must be decreased until learning is not efficient anymore. The other hyper-parameters could then be perturbed for trading-off exploration and exploitation even more. 

We should note that the approach mentioned above did not leverage any particular structure of the loading pattern problem beyond the shape of the reward. The approach is problem agnostic in that it can be extended to any other engineering problem in particular with combinatorial structure, where physic solver or SO-based approaches must be utilized.  Typically, RL can seamlessly replace GA as the optimizer in \cite{yilmaz2022lot} for solving an instance of the Hybrid Flowshop Scheduling Problem (HFSP). An action would represent a chromosome structure, which results in a candidate. If the episode length \code{NF} is greater than 1 (which we argument against), \code{NF} HFSP solutions or chromosomes are generated per episode. The difference with this paper of course is that we provide one solution based on the template strategy (see section \ref{sec:description_core}). In \cite{yilmaz2022lot} however, they utilize GA as a decision-support tool to gain managerial insights about strategy implemented to solve a problem. Utilizing our approach instead could help improve their insights on lot streaming strategies and subplots methodologies for gaining even more performance on the HFSP. Similarly, \cite{seurin2023pareto} addresses the multi-objective version of our problem and the multi-objective RL can replace the NSGA-II in the Multi-Objective Disassembly Line Balancing Problem (MODLBP) \cite{yilmaz2022tactical} to gain managerial insights about tactical strategies and operational scenarios for the MODLBP \cite{yilmaz2022tactical}.

In section \ref{sec:applicationtobenchmark} we apply this approach to classical benchmarks and empirically demonstrate that the aforementioned approach extends to other types of problems.

\subsection{Application of the method to classical benchmarks}
\label{sec:applicationtobenchmark}

In this section, we apply our approach to four classical benchmark functions: one is the Sphere function and the others originate from the CEC17 test suite \cite{awad2017problem} namely, func4 (shifted and rotated rosenbrock), func5 (shifted and rotated rastrigin), and func6 (shifted and rotated expended schaffer). For the \code{NF} and \code{nsteps} tests we choose an input dimension of nx equal to 2, while 10 for \code{ent\_coef}. As we are aiming at solving CO problems, the input space is a subset of $\{-35,35\}^{nx}$ for the Sphere and $\{-100,100\}^{nx}$ for the functions of the CEC17 test suite. Hence, the number of solutions is $(nx)^{70}$ or $(nx)^{200}$. Moreover, the data points in the discretization of \code{NF}, \code{nsteps}, and \code{ent\_coef} are spaced enough to illustrate the behavior of these parameters and are not guided by any desiderata of fine-tuned optimization. The benchmark, as well as the algorithm, are available in NEORL's package \cite{radaideh2023NEORL}. The results are showcased in figure \ref{fig:fullbenchmarks}, where the optimal objective for each function is the number times 100 (e.g., for func2 it is $2 \times 100 = 200$):
\begin{figure}[H]
    \centering
    \includegraphics[scale = 0.6]{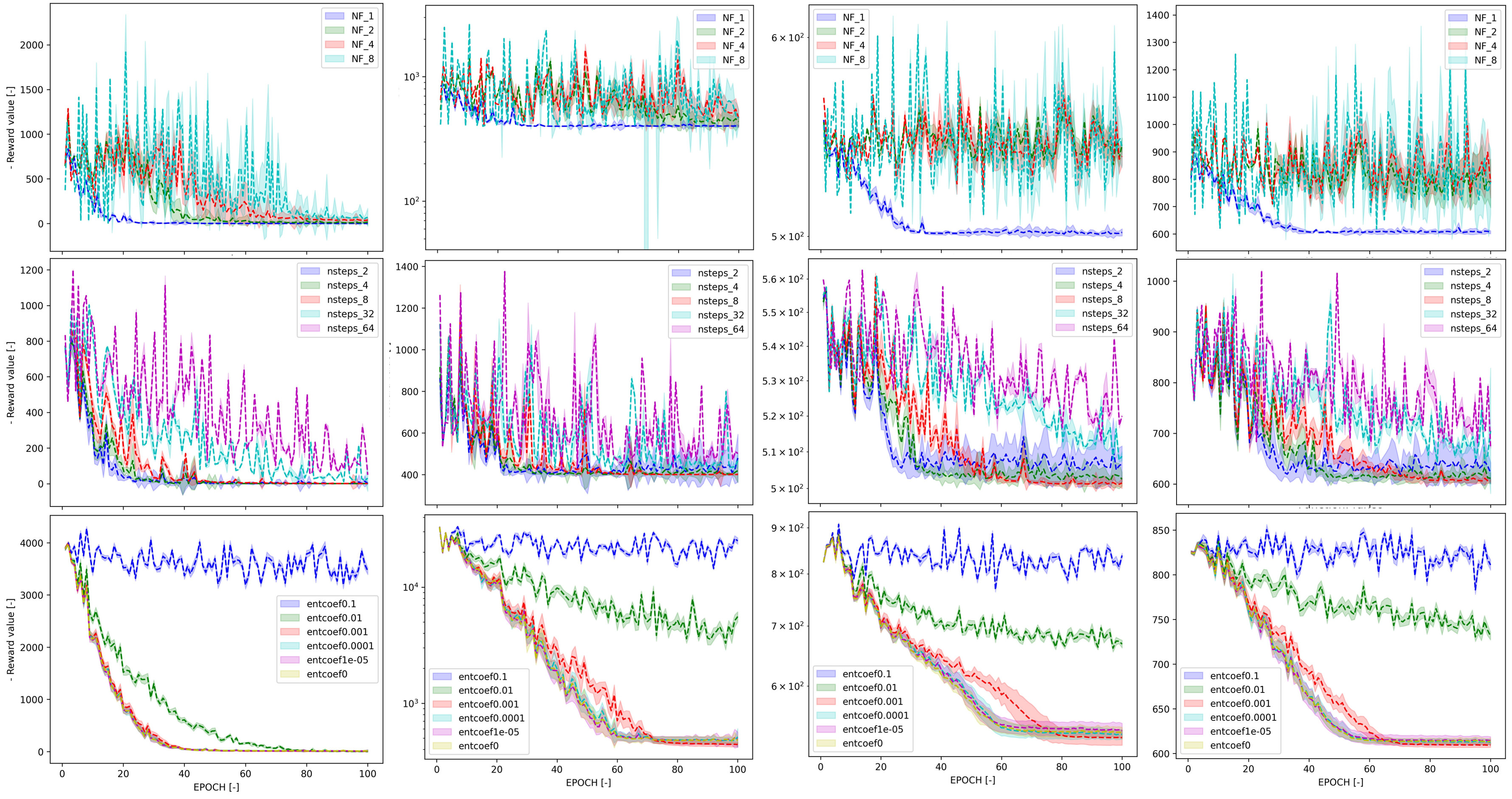}
    \caption{Application of our methodology with changes in (from top to bottom) \code{NF}, \code{nsteps}, and \code{ent\_coef} to four classical benchmarks (from left to right) Sphere, func4, func5, and func6. We use \code{ncores} equal 8. 20 experiments of 10,000 samples are run for each experiment for \code{NF} and \code{nsteps}, and 20,000 for \code{ent\_coef}. The other hyper-parameters are similar to that of the PWR loading pattern problem (see section \ref{sec:shortsummary}).}
    \label{fig:fullbenchmarks}
\end{figure}
As observed on the first row of figure \ref{fig:fullbenchmarks}, \code{NF} equal to 1 outperforms the other. In practice, for the higher \code{NF} to converge to the optimal values, we needed to increase the number of samples collected to at least 40,000, matching the observation for the PWR problem, which is not efficient. For the second row, we observe a trade-off between the different \code{n\_steps}, where lower values result in faster convergence, but the value of 8 (hence \code{ncores}$\times$ \code{nsteps} = 64, similar to case 2 of table \ref{tab:nsteps_stats}) results in more stable learning (e.g, for func6 cases 2 and 4 have high variance). Lastly, for the third row, very high \code{ent\_coef} leads to no learning (e.g., 0.1 and 0.01 (for all but the Sphere)), due to the high randomness of the policy as observed for the PWR problem, while decreasing its value results in faster convergence. Interestingly, a value of 0.001 results in the global optimum systematically as observed for the PWR problem. We, therefore, demonstrated the extent to which our approach to applying RL described in the last paragraphs of section \ref{sec:shortsummary} is more sample efficient beyond only the PWR loading pattern optimization problem.
\section{Concluding remarks}
\label{sec:conclusion}
Deep RL is a framework which gained popularity in recent years thanks to its ability to uncover solutions to problems which outclass expert designers \cite{bengio2018machine} with applications to ATARI games \cite{mnih2015human}, GO \cite{silver2017mastering}, or CO \cite{mazyavkina2021reinforcement} for instance. There, RL does not work with an analytical representation of an objective function but rather with objects (e.g., function approximators) that are constructed statistically with a large number of data and which are then mathematically optimized. In that respect, this work proposed the first-of-a-kind approach to deal with an RL-compliant PWR loading pattern optimization.
 
 In this work, a first-of-a-kind hyper-parameter study of the RL framework applied to solve the loading pattern optimization problem over 24 hours is presented.  The shape of the reward, specifically the \code{weights} assigned to the constraints, is pivotal in finding feasible patterns. It behaves as a fudge factor that could lead to substantial economical improvement but also affect the stability of the solutions found throughout different experiments, which is an essential factor for scaling the algorithm to other applications. On top of that, we found that the quality of the best LP found is highly dependent on the \textit{exploration/exploitation} trade-off that manifests through the \code{NF}, \code{n\_steps}, and \code{ent\_coef} coefficients. Low values for the latter result in high exploitation, which allows the agents to find high-quality patterns very rapidly, and is interesting for the core designer in case of emergency core re-design. We show that the optimum value for \code{NF} is 1, which means that RL is optimally leveraged when it is applied similar to a Gaussian Process where the acquisition function is learned via a policy, which then converges close to the optimum solution after a certain number of steps. However, the behavior for low values for \code{n\_steps} and \code{ent\_coef} is not always linear anymore, where a sweet spot must be found to trade off quality, speed of convergence, and stability. While \code{n\_steps} controls the quality of the estimation of the gradient of the loss, \code{ent\_coef} controls the randomness of the policy. They must both be decreased until no learning is observed or instability occurs to achieve the highest sample efficiency. Additionally, the small deviation of the SE metrics over the experiments for the set of hyper-parameters found demonstrated that RL-based approaches are robust and stable, which is what core designers are looking for to adopt a novel set of techniques in their work. We also briefly studied the other parameters but empirically showed that they were less important. 
 
 Most importantly, we argue that for different engineering problems, the values can differ but the behavior could be similar. We demonstrated this by applying this approach to four classical benchmarks with an integer input space. It makes the approach developed in this work suitable for any large-scale constrained problem in which expensive physic solver or SO-based approaches must be utilized, in particular in the realm of CO.

 On top of that, due to the nature of the end-user, it is fundamental to propose a multi-measure approach in the decision of determining the best set of hyper-parameters.  We show that classical metrics expected by the core designer such as the max objective is not sufficient anymore when dealing with the RL framework, for which the mean and standard deviation of the objective are testament of the potential of the RL agents to improve in longer runs. In this prospect, we proposed to look at the max objective and the best mean objective per episode, the improvement ratio, the mean objective SE, the $\sigma$SE, and the p-value of statistical tests namely FT and NT. Nonetheless, these metrics based on a single scalar value for the reward will not be sufficient when a multi-objective-based optimization is performed. In this prospect, the work in \cite{seurin2023pareto} extends the current study with multi-objective RL and attempts to propose appropriate supplementary metrics to tackle this limitation. 
 
 Lastly, we found that the economic gain from the original pattern (with slightly empirically optimized hyper-parameters) of appendix \ref{sec:weights_study} with LCOE at 5.590 to 5.529 of appendix \ref{sec:entcoef_study} is of the order of 535,000 - 642,000 \$/year/plant for a 1,000 and 1,200 MWe NPP, respectively. This illustrates the requirement to conduct a thorough hyper-parameter study. Nevertheless, these hyper-parameters only belong to the realm of the PPO framework and we have not compared the latter against legacy SO heuristic-based methods such as SA, GA, or TS. This comparison will be all the more interesting as we argued that using \code{NF} equal to 1 can be thought of as if the agents are trying to find the steepest ascent direction based on their understanding of the curvature of the objective space, while SO-based methods are taking random steps to find it. Can we gain robustness and stability when these directions are learned? Future work should include the comparison to ensure the benefit of introducing a RL-based framework for the PWR optimization problem. But the works conducted in \cite{radaideh2021large,radaideh2021physics} on lattice and bundle designs, respectively, give hope that RL can outperform SO-based heuristic methods on our problem. 

\paragraph{Acknowledgments}
This work is sponsored by Constellation (formally known as Exelon) Corporation under the award (40008739). Feedback from Neil Huizenga  and Antony Damico of Constellation were helpful in setting up the problem constraints. 

\paragraph{Data availability statement}
The sponsor of this study did not give written consent for their data to be shared publicly, so due to the sensitive nature of the research supporting data is not available.

\paragraph{Authors Contributions}
 Methodology, software, formal analysis and investigation, visualization, writing - original draft preparation: Paul Seurin; conceptualization, methodology, funding acquisition, supervision, writing - review and editing: Koroush Shirvan. All authors have read and agreed to the published
version of the manuscript.

\section*{Declarations}
\paragraph{Conflict of Interests}
The authors declare that they have no known competing financial interests or personal relationships that could have appeared to influence the work reported in this paper.
\small
 \setlength{\baselineskip}{12pt}
 \bibliographystyle{manuscripts}
\bibliography{manuscripts}
\pagebreak
\normalsize
\appendix
\input{annexe.tex}

\end{document}

%% file: annexe.tex
\section{Appendix 1: PPO hyper-parameters studies}
\label{sec:ppohyperparameterstarts}
In this appendix we provide the results of the major hyper-parameters alluded in section \ref{sec:ppo_hyperparameters}. The shape of the reward via \code{weights} , the length of an episode via \code{NF} , the \code{n\_steps}, and the entropy coefficient \code{ent\_coef}.
\subsection{\code{weights} study}
\label{sec:weights_study}
\begin{figure}[H]
    \centering
    \includegraphics[scale = 0.45]{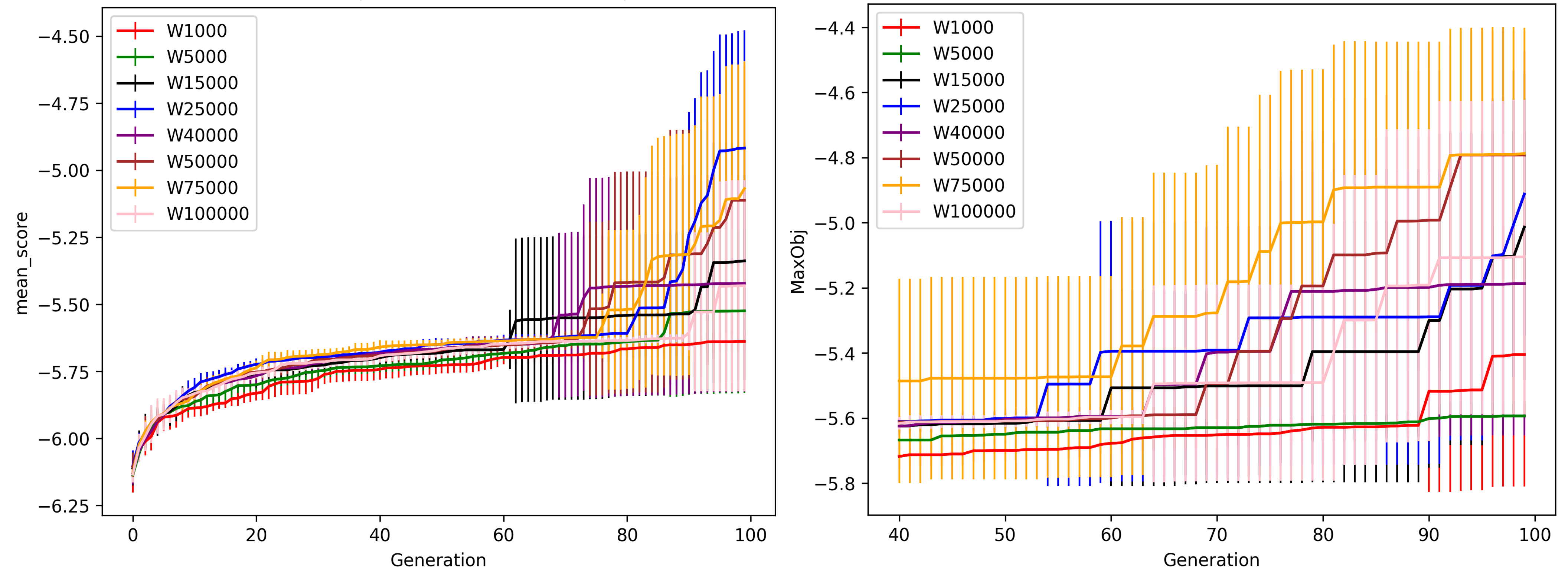}
    \caption{Error bar plots averaged over 10 experiments for the Max (left) and Mean (right) Objective
per episode for different \code{Weights} values. The Mean was obtained by averaging the mean score obtained per episode within a window of X episodes, where X depends on the total number of samples generated (usually around 400 samples for \code{NF} equal to 1 to obtain 100 generations). The length of the bars is two $\sigma$.}
    \label{fig:bothweightstudy}
\end{figure}

Defining the reward for the agent behaving in accordance with the designer's goals is called  \textit{reward hacking}. Defining the reward to optimize the performance of the agents is called \textit{reward shaping}. This study explores the influence of setting weights on design constraints. Weights of 1,000, 5,000, 15,000, 25,000, 40,000, 50,000, 75,000, and 100,000 are considered. Moreover, all constraints will have the assignment of the same weight, since good quality results could not be found when we manually experimented with varying the individual weights. To compare the obtained pattern on the same baseline, the penalization of the obtained pattern is normalized as follows \cite{konak2006multiobjective}:
\begin{equation}
    D(x,\mathcal{C}) = \sqrt{\sum_{i \in \mathcal{C}}(\frac{c_i(x) - c_i}{c_{i_{max}} - c_{i_{min}}})^2}
    \label{eq:normalized_obj}
\end{equation}
where $D(x,\mathcal{C})$ is the distance of the core $x$ to the space of feasible core $\mathcal{C}$ in the objective space, and $c_i, i \in \{1,...,5\}$ are the constraints defined in \ref{sec:objandconstr} ($c_6,c_7$ are always satisfied), which ranges are given in table \ref{tab:rangevlues}. 
\begin{table}[H]
    \centering
    \caption{Upper and lower bounds utilized to calculate the range in equation \ref{eq:normalized_obj}. They were obtained by running one optimization and taking a lower and upper bound with the best and worst cases.}
    \begin{tabular}{c|c|c}
    \hline
    FOMs & Upper Bound & Lower Bound \\
    \hline
    $F_q$ [-] & 2.911 & 1.768 \\
    $F_{\Delta h}$ [-] & 2.206 & 1.429 \\
    $C_b$ [ppm] & 1405.1 & 952.7 \\ 
    Pin Peak $Bu$ [GWd/tHm] & 68.044 & 55.506 \\ 
    Cycle length [days] & 530.8 & 468.9   \\
         \hline
    \end{tabular}
    
    \label{tab:rangevlues}
\end{table}
Figure \ref{fig:bothweightstudy} showcases the evolution of the mean score (left) and max objective (right) per generation. A generation is defined such that the experiments span 100 generations, therefore one contains between 400 to 500 SIMULATE3 evaluations. As expected, the mean score is constantly increasing meaning that the agents are learning. The intricate behavior starting around the 60th generation is happening when a feasible pattern has been found and the agents are only optimizing the LCOE. This occurs at about the same time for all \code{weights}. It is interesting to note that this happens for both mean score and max objective signifying that the agents behave well almost at the same time as they are hitting a feasible pattern and hence received a boost in reward. 

Additionally, the seemingly less smooth behavior compared to other hyper-parameter studies (e.g., see figures \ref{fig:errorbarnfstudy} and \ref{fig:errorbarnstepsstudy}) is due to the normalizing of the objective with equation \ref{eq:normalized_obj}. In all cases, the max objective keeps increasing late during the search, as confirmed by the high IR provided in table \ref{tab:weightstudy_stats} and the standard deviation are in the same range. Therefore, no significant differences can be observed in terms of absolute performance. On top of that, the p-value for the FT with the max score is 0.351 and 0.876 for the mean. Hence these tests failed. However, we decided to stop the experiments as we did not observe significant differences in the performance with the different \code{weights} in contrast to the hyper-parameters studied in the next sections. We defined however more metrics at the beginning of this section and tables \ref{tab:weightstudy_stats} and \ref{tab:weightstudy_stats2} could provide more information and better-informed pathways to decide which \code{weights} to pursue the study with.

In light of table \ref{tab:weightstudy_stats}, the best average behavior for the max objective is obtained with \code{weights} 75,000, which also achieves a very competitive best pattern overall. In RL literature, it is sometimes advised to normalize the reward, otherwise, the magnitude of the gradients will be too different. But here one small change in the core with 75,000 may result in the blowout of the cumulative reward, which could be counter-intuitive. The fact that \code{NF} and \code{n\_steps} equal 2 and 4, respectively, and the use of a clipping term and \code{max\_grad\_norm} mitigate this effect. On top of that, the best pattern was found with \code{weights} 5,000,  while the case 1,000 achieves the best mean score per episode with 25,000 being very close for this metric. But the behavior among the \code{weights} is not linear, as a \code{weights} of 25,000 achieves better results than 40,000, 50,000, and 100,000 in total, while a better mean score than 75,000 and almost as good max objective.  Additionally, case 25,000 achieves the second best performance in terms of mean score, with a mean objective SE at - 5.7360, the highest one. Even though the differences do not look significant, the case 25,000 has the lowest $\sigma$SE at the last generation given in table \ref{tab:weightstudy_stats2}, signifying greater stability. Hence if the input variables were to change without the possibility to run 10 experiments due to time constraints (e.g., in case of emergency core redesign) we could imagine that utilizing 25,000 could help find a feasible pattern with a higher probability.
\begin{table}[H]
    \centering
    \caption{Statistics for each \code{weights} value over 10 experiments. MO stands for Max Objective and Me for Mean objective per episode. The objectives are calculated based on the normalized objective value of equation \ref{eq:normalized_obj}. The best objective values are highlighted in red.}
    \begin{tabular}{c|c|c|c|c|c|c|c|c}
      \hline
    \code{weights} & Avg (MO) & Best (MO) & $\sigma$ (MO) & IR (MO) & Avg (Me) & Best (Me) & $\sigma$ (Me) & IR (Me) \\
     \hline
         1,000   & -5.438 & -4.570 & 0.5247 & 0.946 & \textcolor{red}{-5.500} & \textcolor{red}{-4.570} & 0.4536 & 0.952\\
         5,000   & -5.460 & \textcolor{red}{-4.569} & 0.4362 & 0.992 & -5.680  & -5.557 & 0.086711 & 0.988\\
         15,000  & -5.132 & -4.582 & 0.4994 & 0.787 & -5.725 & -5.602 & 0.08524 & 0.981\\
         25,000  & -4.953 & -4.590 & 0.4716 & 0.973 & -5.573 & -4.683 & 0.2993 & 0.974\\
         40,000  & -5.356 & -4.577 & 0.4834 & 0.968 & -5.693 & -5.614 & 0.04588 & 0.978\\
         50,000  & -5.447 & -4.628 & 0.4037 & 0.993 & -5.685 & -5.588 & 0.05552 & 0.997\\
         75,000  & \textcolor{red}{-4.947} & -4.579 & 0.508 & 0.998 & -5.683 & -5.606 & 0.0555 & 0.9555\\
         100,000  & -5.480 & -4.572 & 0.441 & 0.849 & -5.717 & -5.635 &  0.05234 & 0.966\\
    \end{tabular}
    \label{tab:weightstudy_stats}
\end{table}
\begin{table}[H]
    \centering
    \caption{Mean Objective SE and the $\sigma$SE for the different \code{weights} cases.  The latter is obtained by averaging the reward obtained over the last generation and over 10 experiments. The best performance for both is obtained with \code{weights} equal to 25,000. The best values are highlighted in red.}
    \begin{tabular}{c|c|c}
      \hline
    \code{Weights} & Mean Objective SE & $\sigma$SE \\
     \hline
         1,000   & -5.8573 & 0.09069\\ 
         5,000   & -5.8074 & 0.08265\\
         15,000  & -5.8096 & 0.08514\\
         25,000  & \textcolor{red}{-5.7360} & \textcolor{red}{0.07093}\\
         40,000  & -5.7971 & 0.0841\\
         50,000  &  -5.7956 & 0.0838\\ 
         75,000  &  -5.7590 & 0.09441\\ 
         100,000 & -5.8166 & 0.08657\\ 
    \end{tabular}
    \label{tab:weightstudy_stats2}
\end{table}
The breakdown of the FOMs of the best patterns found in each case is given in table \ref{tab:weightstudy_bestpattern}. The constraints $c_6$ and $c_7$ turned out to be usually satisfied and were omitted from the table, which will be the case in the remainder of the study. It should be noted that for the best pattern, some constraints are close to being tight (e.g., $F_{\Delta h}$, $C_b$, and Pin Peak $Bu$) in contrast to the one found with \code{weights} 25,000 for instance. This may show that to optimize the LCOE further, penalties on some constraints may need to be taken.

\begin{figure}[H]
    \centering
    \includegraphics[scale = 0.6]{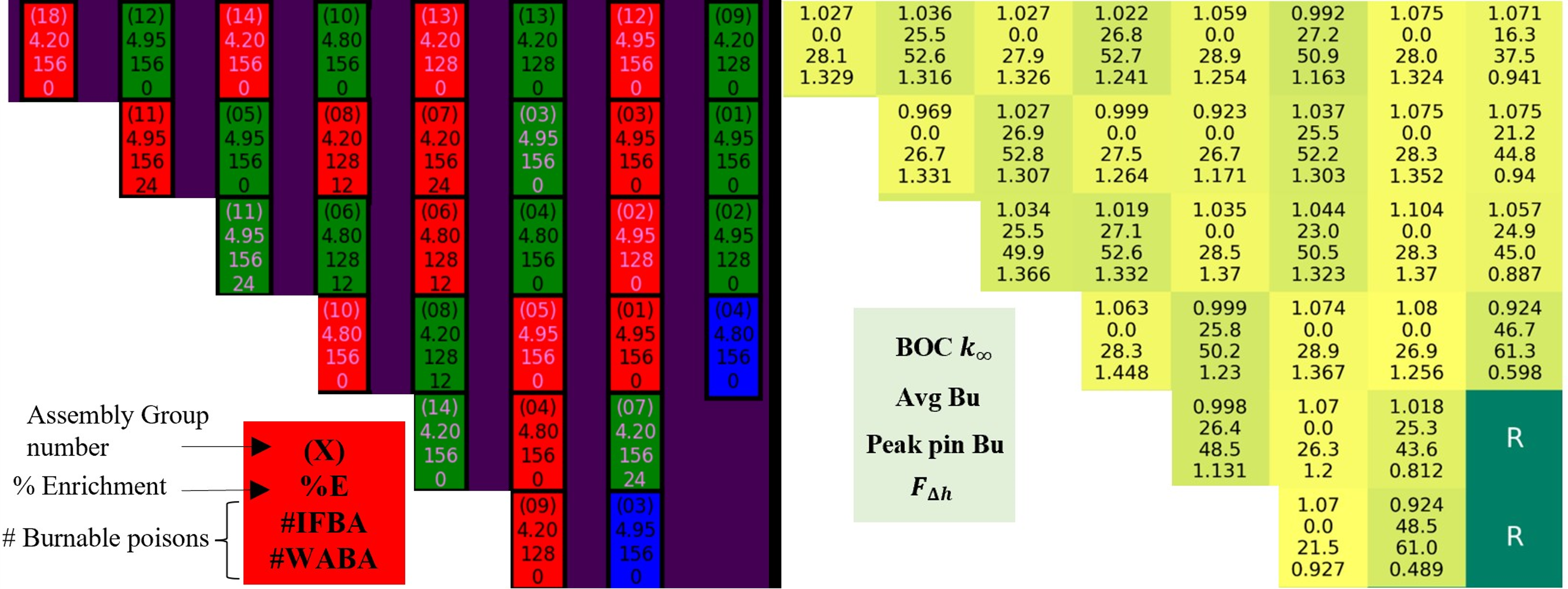}
    \caption{Lower right quadrant of the best pattern found obtained for \code{weights} equal to 5,000. On the left, from top to bottom: Assembly group number, fuel average enrichment, number of IFBA rods, and number of WABA rods. Blue is for twice-burned FAs, green for once-burned, and red for fresh (colors pattern also utilized in ROSA \cite{verhagen1997rosa}). In violet, location where WABA cannot be placed. On the right, from top to bottom: Beginning of cycle $k_{\infty}$ and assembly average burnup, peak pin Burnup at EOC, and $F_{\Delta h}$ when the maximum over the cycle is reached. R means reflector, where the water is.}
    \label{fig:bestpatternpostweights}
\end{figure}

The 2D projection of the best pattern found (i.e., with \code{weights} equal to 5,000) is given in figure \ref{fig:bestpatternpostweights}. This case presents a high IR and a low $\sigma$SE throughout the 10 experiments. It means that the agents are constructing high-quality patterns at each episode, which is a testament of the agents's understanding of the objective space. There, the agents employed several classical designer's tactics even though no prior knowledge of the physics involved was given before the search started: the inner part of the core is mostly arranged like a checkerboard between fresh and once-burned FAs, a higher BP concentration is given to the FAs closer to the center, and FAs with a higher $k_{\infty}$ are located near the periphery, which help to reduce the peaking factor and boron concentration \cite{lin2014themaxmin}. This is worth noting as many papers instead impose that these expert heuristics are satisfied in the decision process as in \cite{wu2016quantum,lin2014themaxmin,garco2008new}. However, usually designer put more WABA in the diagonal and the ring of fire because the power is shifted from the center of the core (where low enrichment and high IFBA are placed) to there. But here the agent decided to also utilize high WABA in-board but not in the ring of fire. This may be explained by the fact that WABA costs money, hence reducing its amount is pivotal, and removing it from the ring of fire may be the path of lowest resistance. Other Expert tactics for twice-burned include choosing once-burned with the highest $k_{\infty}$ to obtain enough carry-over energy for the next cycle and FAs with low average burnup to limit the risk of high peak pin burnup at the periphery. Here the average burnup of twice-burned at EOC is not high but not particularly low neither for the ones on both sides of the diagonal (48.5 GWd/tHm at BOC). These FAs will then peak at 61.0. The peak pin burnup is also close to 62 GWd/tHm (61.3 GWd/tHm for the fourth FA starting from the left or the top). This may be explained because we are not only studying safety parameters. There is no penalty as long as the peak pin burnup is within 62 GWd/tHm, but the LCOE is decreased by extracting more energy out of these FAs.
 \begin{table}[H]
    \centering
    \caption{Best patterns found for each case throughout all experiments. An Objective above - 5.00 signifies that the pattern is feasible and the agents are optimizing the LCOE only.}
    \begin{tabular}{c|c|c|c|c|c|c|c}
      \hline
     \code{Weights} & Objective & LCOE & Cycle length & $F_q$ & $F_{\Delta h}$ & $C_b$ & Pin peak $Bu$\\
     \hline
         5,000    & \textcolor{red}{-4.569} & 5.569 & 519.1 & 1.812 & 1.448 & 1189.3 & 61.262\\
         1,000    & -4.570 & 5.570 & 510.8 & 1.831 & 1.436 & 1195.8 & 61.581\\
         100,000  & -4.572 & 5.572 & 507.4 & 1.801 & 1.439 & 1179.5 & 61.485\\
         40,000   & -4.577 & 5.577 & 505.5 & 1.813 & 1.449 & 1199.1 & 59.759\\
         75,000   & -4.579 & 5.579 & 501.1 & 1.834 & 1.450 & 1198.3 & 60.773\\
         15,000   & -4.582 & 5.582 & 511.3 & 1.810 & 1.419 & 1190.6 & 61.518\\
         25,000   & -4.590 & 5.590 & 511.7 & 1.835 & 1.438 & 1183.0 & 61.058\\
         50,000   & -4.628 & 5.628 & 505.0 & 1.824 & 1.448 & 1138.1 & 59.767\\
   \end{tabular}
    
    \label{tab:weightstudy_bestpattern}
\end{table}

To conclude this section, we can say that the \code{weights} do not participate in the \textit{exploration/exploitation} trade-off, which makes sense as they are affecting the magnitude of the reward only. It remains a very important parameter nonetheless. It behaves as a fudge factor which could result in an improvement on the economic side and the stability of the agents. Comparing the LCOE of the patterns found with \code{weights} 5,000 and 25,000, a gain of about 180,000 and 220,000 \$/year/plant for a 1,000 and 1200 MWe NPP, respectively, could be achieved by tuning the \code{weights} appropriately. However, we demonstrated with a multi-measure analysis that looking at the maximum objective is not sufficient, specifically because it does not necessarily yield to a better mean objective SE, which does not offer high certainty to find a feasible pattern rapidly with a different seed. The latter becomes instrumental if different input variables were given and a new problem must be solved rapidly which is achieved with more certainty with 25,000 as \code{weights}. All the observations made in this section motivated us to pursue with 25,000 as it presented the best trade-off between our multiple performance metrics.

\subsection{Numframes \code{NF} study}
\label{sec:nf_study}
\begin{figure}[H]
    \centering
    \includegraphics[scale = 0.6]{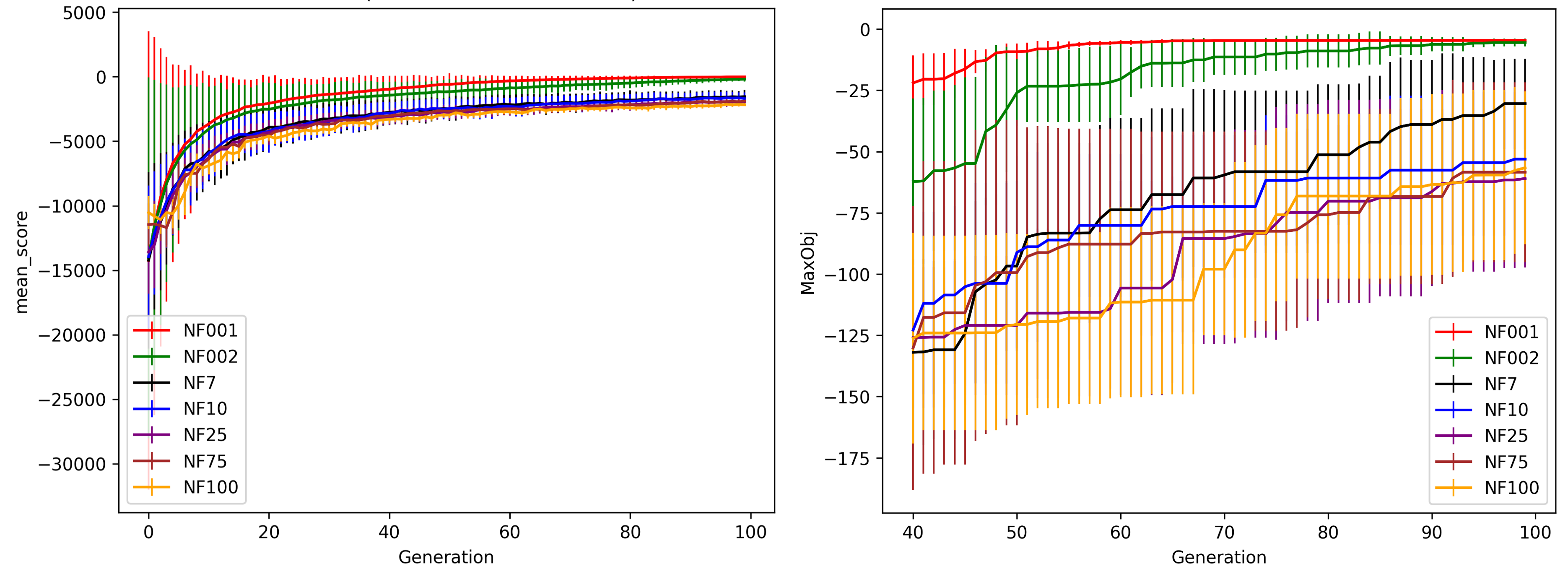}
    \caption{Error bar plots averaged over 10 experiments for the Max (left) and Mean (right) Objective per episode for different \code{NF} values. The Mean was obtained by averaging the mean score obtained per episode within a window of X episodes, where X depends on the total number of samples generated (usually around 400 samples for \code{NF} equal to 1 to obtain 100 generations). The length of the bars is two $\sigma$.}
    \label{fig:errorbarnfstudy}
\end{figure}

A prior study \cite{radaideh2023NEORL} showed that \code{NF} should not vary beyond 100. Here, the values are 1, 2, 7, 10, 25, 75, and 100. Figure \ref{fig:errorbarnfstudy} showcases the evolution of the mean score and max objective over 100 generations. For all \code{NF} there is continuous learning occurring however it is clear in this figure that very low \code{NF} values (i.e., 1 and 2) outperform higher ones, which is confirmed by table \ref{tab:nfstudy_stats} for the Avg (Me). Regarding the right-hand side of figure \ref{fig:errorbarnfstudy}, we could imagine that the performance of the case \code{NF} equal to 1 saturates after the 50th generation. Nevertheless, this is simply the effect of scaling. The IR is high above 87\% meaning that the agents are still improving their learning of the system and the high $\sigma$SE of table \ref{tab:nfstudy_stats2} confirms this hypothesis. More generally for all cases, the IR is high for both the mean and the max objective, meaning that the agent's decisions are indeed improving, except for the max objective of case 100, which may hint that this solution has been found randomly. The very low mean objective SE for case 100 given in table \ref{tab:nfstudy_stats2} confirms this hypothesis. 

10 experiments for each case were ran. The p values for the FT tests are all 0.00. Cases 1,2, and 7 were not statistically significantly different  as provided in tables \ref{tab:pvalue_numframesstudymeanscore} and \ref{tab:pvalue_numframesstudymaxscore}. However, the p-value of case 1 versus case 7 is lower than that of case 2 versus case 7 for both max and mean score, which may hint that case 1 is superior. The differences are sharper than that of the \code{weights}'s study. This should mean that 1 outperforms the other cases, which aligns with our previous observation and we decided to pursue with this value for the following studies.

Overall, it is clear that the case \code{NF} equal to 1 outperforms the others and tables \ref{tab:nfstudy_stats} and \ref{tab:nfstudy_stats2} show that there is a very sharp learning improvement when \code{NF} is decreased at a value lower than 7, while the difference is less obvious above 7.  In addition, we can observe in table \ref{tab:nfstudy_stats} that a lower value of \code{NF} usually yields a lower variance between the results at each generation and always a better average behavior promising better stability.  In the first order, therefore, \code{NF} is a manifestation of the \textit{exploration/exploitation} trade-off. Higher \code{NF} yields higher exploitation and lower mean objective but sometimes a better max objective (e.g., the max objective with \code{NF} equal to 75 exceeds that of 2 to 25), which is of interest to the core designer. For low and high values of \code{NF}, the $\sigma$SE decreases as seen in table \ref{tab:nfstudy_stats2}. This column shows that something else is happening. A high \code{NF} does not necessarily involve slower convergence but in light of the column mean objective SE it does imply a lower mean objective. Therefore high \code{NF} should be avoided even if the optimization was run longer due to the risk of being stuck in a local optimum. 

It could be surprising to observe such a low mean objective SE even for the case \code{NF} equal to 2 compared to 1.  Having \code{NF} equal to 1 with the restarted state being the best ever found can be thought of as a gradient ascent problem: At each episode, based on the understanding of the curvature of the objective space, the agents are trying to find the steepest ascent direction similar to SO-based methods where the steps are taken however randomly.Indeed in SO, the agents are following a rule of thumb (i.e. heuristic) in the decision process. It is a guided random walk. In RL, we have a policy that generates solutions, while a critic learns and evaluates the quality of these solutions. The heuristic is learned. In this sense, it is similar to a Gaussian Process \cite{frazier2018tutorial} in which the acquisition function is a policy we learn, and the policy converges close to the optimum solution found. To the author of this work, there is no obvious mathematical intuition behind greater \code{NF}. We initially thought that asking the agent to solve multiple LPs at the same time would help as one state of a pattern is the restarting state for the action to choose the next pattern. These information could have helped the agent draw better conclusions from one state to the next. Nevertheless, it is clear that we are asking the agents to solve multiple loading designs at the same time, therefore more samples must be drawn to fully comprehend how an episode is constructed, which might be more difficult and explains why it is seemingly easier with the lowest \code{NF}. This study may also explain why value-based methods (e.g., Deep Q-Learning (DQN) \cite{mnih2015human} or Double Q-Learning \cite{van2016deep}) were not performing well.  One explanation might be that they required one or multiple-step look-ahead to optimize the Q-function \cite{hessel2018raindbow}. However, as it was described in section \ref{sec:mdp}, each action is a core created. Since a full core is created for each action, each one fundamentally should not influence the generation of the next and the Q-update might be ineffective.

 \begin{table}[H]
    \centering
     \caption{Statistics for each \code{NF} value over 10 experiments. MO stands for Max Objective and Me for Mean objective per episode. The best result for each column is highlighted in red.}
    \begin{tabular}{c|c|c|c|c|c|c|c|c}
      \hline
     Numframes & Avg (MO) & Best (MO) & $\sigma$ (MO) & IR (MO) & Avg (Me) & Best (Me) & $\sigma$ (Me) & IR (Me) \\
     \hline
          1 & \textcolor{red}{-4.599}  & \textcolor{red}{-4.556} & 0.0257  & 0.879 & \textcolor{red}{-4.600}   &   \textcolor{red}{-4.5560} & 0.0257 & 0.879\\
          2 &-5.454  &  -4.556 & 1.646   & 0.879 & -18.009  &   -4.5600 & 17.6663& 0.807\\
         7  &-30.421 & -14.403 & 18.408  & 0.846 & -624.044 &  -518.000 & 95.611 & 0.797\\
         10 & -51.106 & -9.100 & 27.522 & 0.855 & -790.045 &  -625.610 & 67.623 &  0.998\\
         25 &-44.986 &  -5.710 & 24.495  & 0.567 & -1200.059&  -1036.77 & 91.847 & 0.991\\
        75  &-43.779 &  -4.566 & 35.0578 & 0.863 & -1479.925&  -1316.27 & 90.349 & 0.799\\
        100 &-53.470 & -18.956  & 30.311  & 0.691 &-1648.797 & -1434.82  & 234.060& 0.855\\
    \end{tabular}
    \label{tab:nfstudy_stats}
\end{table}

\begin{table}[H]
    \centering
    \caption{Mean Objective SE and $\sigma$SE for the different \code{weights} cases.  The latter is obtained by averaging the reward obtained over the last generation and over 10 experiments. The case \code{NF} equal to 1 has the best combination of both.}
    \begin{tabular}{c|c|c}
      \hline
    \code{NF} & Mean Objective SE & $\sigma$SE \\
     \hline
         1&  \textcolor{red}{-14.705}    & \textcolor{red}{67.368}\\ 
         2&  -188.447   & 177.915\\
         7&  -1548.257  & 514.939\\
         10&  -1689.118  & 481.794\\
         25&  -1928.256  & 335.519\\
         75&  -1887.0760 & 186.982\\
         100& -2174.317  & 118.620\\
    \end{tabular}
    \label{tab:nfstudy_stats2}
\end{table}%

\begin{table}[H]
    \centering
    \caption{p-value for the NT between the \code{NF} instances for the maximum score. The FT succeeded with p-value equal to 0.00.  \code{NF} equal to 1 is the case that outperforms most of the other experiments. The tests that succeeded are highlighted in red.}
    \begin{tabular}{c|c|c|c|c|c|c|c}
    \hline
        \code{NF} & 1 & 2 & 7 & 10 & 25 & 75 & 100 \\
    \hline
1   & 1.000000 & 0.900000 & \textcolor{red}{0.026662} & \textcolor{red}{0.001000} & \textcolor{red}{0.002000} & \textcolor{red}{0.013555} & \textcolor{red}{0.00100} \\
2   & - & 1.000000 & 0.114484 & \textcolor{red}{0.004459} & \textcolor{red}{0.013555} & \textcolor{red}{0.066772} & \textcolor{red}{0.00654} \\
7   & - & - & 1.000000 & 0.900000 & 0.900000 & 0.900000 & 0.90000 \\
10  & - & - & - & 1.000000 & 0.900000 & 0.900000 & 0.90000 \\
25  & - & - & - & - & 1.000000 & 0.900000 & 0.90000 \\
75  & - & - & - & - & - & 1.000000 & 0.90000 \\
    \end{tabular}
    
    \label{tab:pvalue_numframesstudymaxscore}
\end{table}

\begin{table}[H]
    \centering
    \caption{p-value for the NT between the \code{NF} instances for the mean score per episode. The FT succeeded with the p-value equal to 0.00. If $\alpha = 0.1$, which is sometimes considered \cite{derrac2011apractical}, \code{NF} equal to 1 is the case that outperforms most of the other experiments.}
    \begin{tabular}{c|c|c|c|c|c|c|c}
    \hline
        \code{NF} & 1 & 2 & 7 & 10 & 25 & 75 & 100 \\
    \hline
1  & 1.000000 & 0.900000 & 0.339674 & 0.087581 & \textcolor{red}{0.001315} & \textcolor{red}{0.001000} & \textcolor{red}{0.001000} \\
2  & - & 1.000000 & 0.777817 & 0.404855 & \textcolor{red}{0.019139} & \textcolor{red}{0.001000} & \textcolor{red}{0.001000} \\
7  & - & - & 1.000000 & 0.900000 & 0.503923 & \textcolor{red}{0.022624} & \textcolor{red}{0.005411} \\
10 & - & - & - & 1.000000 & 0.869118 & 0.129190 & \textcolor{red}{0.042827} \\
25 & - & - & - & - & 1.000000 & 0.808251 & 0.564789 \\
75 & - & - & - & - & - & 1.000000 & 0.900000 \\
    \end{tabular}
    \label{tab:pvalue_numframesstudymeanscore}
\end{table}

\begin{figure}[H]
    \centering
    \includegraphics[scale = 0.6]{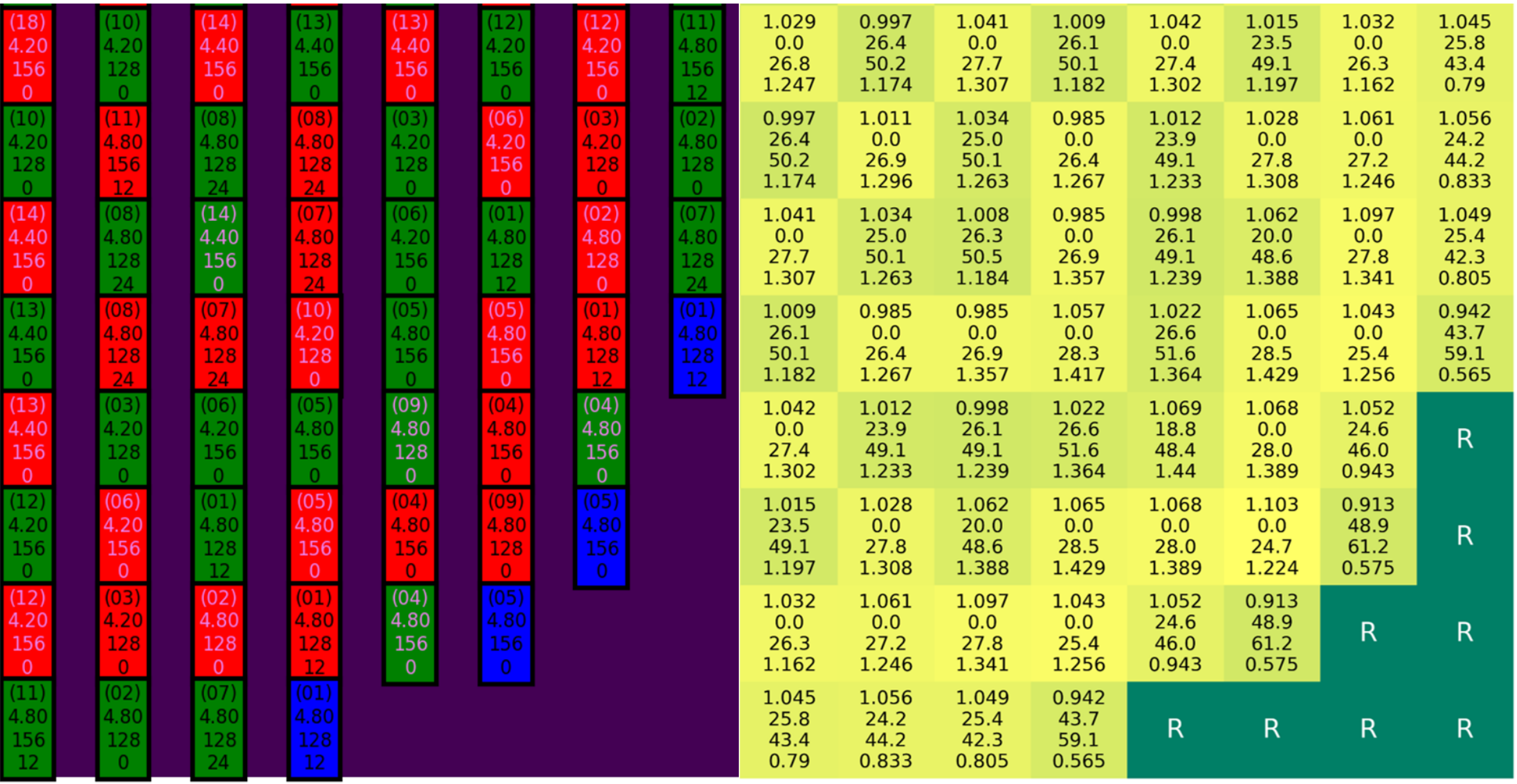}
    \caption{Lower right quadrant of the best pattern found with \code{NF} equal to 1. On the left, from top to bottom: Assembly group number, fuel average enrichment, number of IFBA rods, and number of WABA rods. Blue is for twice-burned FAs, green for once-burned, and red for fresh (colors pattern also utilized in ROSA \cite{verhagen1997rosa}). In violet, location where WABA cannot be placed. On the right, from top to bottom: Beginning of cycle $k_{\infty}$ and assembly average burnup, peak pin Burnup at EOC, and $F_{\Delta h}$ when the maximum over the cycle is reached. R means reflector, where the water is.}
    \label{fig:bestpattern_nfstudy}
\end{figure}

Figure \ref{fig:bestpattern_nfstudy} showcases the 2D projection of the best pattern found in this study (corresponding to the top row of table \ref{tab:nfstudy_bestpattern}). There, in contrast to figure \ref{fig:bestpatternpostweights}, rings of fire are observed from the periphery to the center, which is also a typical designer strategy. Lastly, by comparing the solution of \code{NF} equal to 1 and 2 in table \ref{tab:nfstudy_bestpattern}, the constraints are tighter for this design, which helped again to reduce the LCOE. 

\begin{table}[H]
    \centering
    \caption{Breakdown of the FOMs for the best patterns found for each case throughout all experiments. An objective above - 5.00 signifies that the pattern is feasible and the agents are optimizing the LCOE only.}
    \begin{tabular}{c|c|c|c|c|c|c|c}
      \hline
     Numframes & Objective & LCOE & Cycle length & $F_q$ & $F_{\Delta h}$ & $C_b$ & Pin peak $Bu$\\
     \hline
          1 &  \textcolor{red}{-4.547} & 5.547 & 510.8 & 1.845 & 1.440 & 1178.8 & 61.198 \\
          2 &  -4.556 & 5.556 & 510.6 & 1.841 & 1.438 & 1178.4 & 61.172 \\
        75   & -4.566  & 5.566 & 519.0 & 1.764 & 1.448 & 1195.3 & 61.204\\
         10  & -9.100  & 5.735 & 498.9 & 1.863 & 1.463 & 1191.8 & 61.285\\
         7   & -14.403 & 5.602 & 511.7 & 1.841 & 1.457 & 1221.7 & 62.082\\
         25  & -17.734 & 5.686 & 493.4 & 1.856 & 1.473 & 1208.1 & 59.623\\
         100 & -18.956 & 5.610 & 513.4 & 1.867 & 1.473 & 1169.3 & 62.872\\
    \end{tabular}
    \label{tab:nfstudy_bestpattern}
\end{table}

Because the case \code{NF} equal to 2 presented a higher variance with a good Avg (Me), hence learning potential, we also tried to restart the experiment with the trained policy from this case for another one day of optimization. We also increases \code{NF} to 250 and stopped at 6 experiments because the $\sigma$ were very low. It turned out that 75 was the best case on average but no parametric tests succeeded as the differences were not significant, and the results are not presented for conciseness. The better performance of 75 (max and average max objectives of -4.558 and -4.573 respectively with $\sigma \sim 0.0120$ averaged over 6 experiments) corresponds to a boost in exploration. In consequence, depending on the context of optimization, there is a trade-off between exploration/exploitation. Preparing the next core cycle, which can take weeks, starting with multiple runs with higher \code{NF} might lead to achieving a better solution than lower \code{NF}, but we should be wary of the faster convergence if \code{NF} is increased.

To conclude this section, we can say that \code{NF} plays a role in the \textit{exploration/exploitation} trade-off. A higher value will yield a lower mean objective. Nevertheless, a lower value may lead to faster convergence, hence reaching and being trapped in an unacceptable local optimum (see case 75). This study also highlighted the importance of fine-tuning. Indeed, a very small change in \code{NF} may drastically change the optimum value found. In terms of the LCOE between 1 and 2 for a 1,000 or 1,200 MWe NPP, we get a gain of 79 - 95 k \$/yr.  On top of that, case 7 did not even find a feasible design within one day of optimization, which is fundamental for certain application as we alluded in section \ref{sec:prerequisite}.
\subsection{\code{n\_steps} study}
\label{sec:nsteps_study}
\begin{figure}[H]
    \centering
    \includegraphics[scale = 0.5]{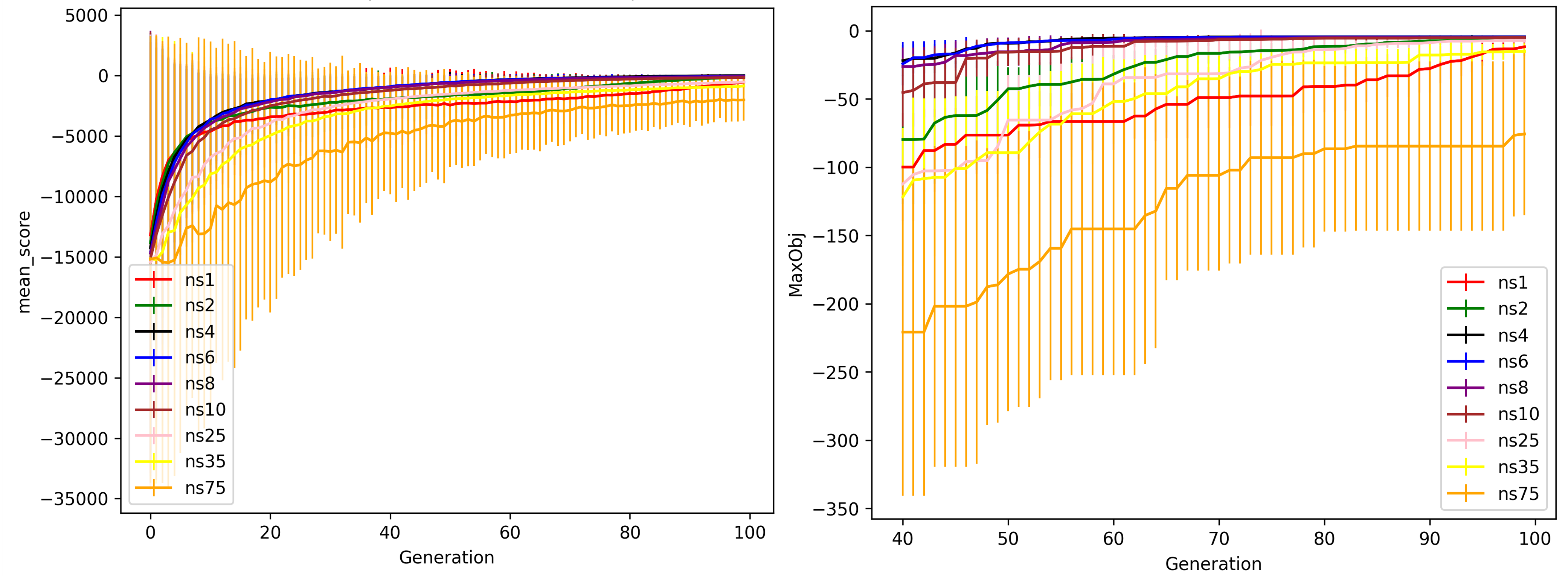}
    \caption{Error bar plots averaged over 10 experiments for the Max (left) and Mean (right) Objective
per episode for different \code{n\_steps} values. The Mean was obtained by averaging the mean score obtained per episode within a window of X episodes, where X depends on the total number of samples generated (usually around 400 samples for \code{NF} equal to 1 to obtain 100 generations). The length of the bars is two $\sigma$.}
    \label{fig:errorbarnstepsstudy}
\end{figure}

The parameter \code{n\_steps} also plays a role in the \textit{exploration/exploitation} trade-off as presented in section \ref{sec:ppo_hyperparameters}. The chosen values for the study are 1, 2, 4, 6, 8, 10, 25, 35, and 75, which were run over 10 experiments. Figure \ref{fig:errorbarnstepsstudy} showcases the evolution of the max and mean objective over 100 generations. It should be observed that there is 
 smooth learning for all values of \code{n\_steps}. 

Similar to the \code{NF} study, large values of \code{n\_steps} yields poor results with an average lower than - 5.00. The best patterns found are not even feasible for cases 35 and 75 as highlighted by table \ref{tab:nsteps_stats}. The mean objective columns of table \ref{tab:nsteps_stats} are omitted as they are similar to that of the max objective with \code{NF} equal to 1. They all have a high IR except for case 75, but its low mean objective SE and $\sigma$SE hint that the solution was found randomly. Moreover, the variance in the results and the $\sigma$SE also increased with \code{n\_steps} for large values. However, this behavior is not linear anymore below 4, where case 1 and 2 yield worse performances than 4, 6, 8, and 10 with a lower mean objective SE and a higher $\sigma$SE as shown in table \ref{tab:nsteps_stats2}. This may be a manifestation of the \textit{exploration/exploitation} trade-off as too few samples are recorded with \code{n\_steps} equal to 1 or 2, which yields sub-optimum behavior. Column $\sigma$SE helps confirm the significance of \code{n\_steps} regarding the trade-off. The higher the value the higher the $\sigma$ except below 4. Moreover, the cases 1 and 2 present a high $\sigma$ (MO), showing that they are also unstable.  

Multiply by \code{ncores}, \code{n\_steps} define the number of samples utilized for the estimation of the gradient of the loss (see Equation \ref{eq:ppoloss}) to update $\theta$, and hence its quality. We hypothesize that the previous behavior is explained because low values (e.g., 1) result in a very poor local approximation of the gradient (i.e., low bias). On the other hand, larger values (e.g., 4, 6, 8...) may decrease the bias of the estimator but increase the variance due to the presence of more terms. In the beginning, the loss is constructed with many samples of poor quality, impeding the speed of learning. Once better samples are obtained, learning happens, and good solutions can be obtained.
  
The mean objective SE and $\sigma$SE increase with \code{n\_steps} above 4, which confirms that this hyper-parameter directly control the level of exploration and exploitation in the search. With an average max objective of -4.599, the best average behavior is observed with 4, while 6 is very close but obtained a better max objective. Looking at the mean objective SE in table \ref{tab:nsteps_stats2} we can see however that case 4 is producing better patterns on average for the last generation with a lower $\sigma$SE, indicating  higher stability and probability to generate a good pattern.

The NTs on the max objective did not succeeded between 2, 4, 6, 8, and 10 as shown in table \ref{tab:pvalue_nstepsstudy}, which is expected considering the few differences observed in table \ref{tab:nsteps_stats}. The case 6 succeeded against every value above 10 while the cases 4, 8,, and 10 succeeded against 2 out of the three values, therefore they might be considered the best cases. However, the considerations summarised above show that case 4 is superior to other metrics. For these reasons, we decided to keep \code{n\_steps} equal to 4 for the remainder of the study. 

 \begin{table}[H]
    \centering
     \caption{Statistics for each \code{n\_steps} values over 10 experiments. MO stands for Max Objective. The Mean objective per episode is not included because it is equal to the Max Objective one for \code{NF} equal to 1. The best result for each column is highlighted in red.}
    \begin{tabular}{c|c|c|c|c}
      \hline
     \code{n\_steps} & Avg (MO) & Best (MO) & $\sigma$ (MO) & IR (MO) \\
     \hline
     1 &  -4.616 & -10.6647 & 6.3040 & 0.947 \\
        2    & -4.902  & -4.551 & 0.548 & 0.925\\
        4    & \textcolor{red}{-4.599}  & -4.556 & 0.027 & 0.879\\
        6    & -4.603  & \textcolor{red}{-4.523} & 0.025 & 0.981\\
        8    & -4.867  & -4.544 & 0.530 & 0.866\\
        10   & -4.944  & -4.540 & 1.026 & 0.925\\
        25   & -7.133  & -4.540 & 1.026 & 0.868\\
        35   & -14.153 & -6.395 & 5.900 & 0.887\\
        75   & -59.128 & -6.542 & 35.286 & 0.554\\
    \end{tabular}
    \label{tab:nsteps_stats}
\end{table}

\begin{table}[H]
    \centering
    \caption{Mean Objective SE for the different \code{n\_steps} cases with the $\sigma$SE. The latter is obtained by averaging the reward obtained over the last generation and over 10 experiments.}
    \begin{tabular}{c|c|c}
      \hline
    \code{n\_steps} &  Mean Objective SE & $\sigma$SE \\
     \hline
        1 & -545.268 & 513.820 \\
         2& -123.293 & 219.7689 \\
         4 & \textcolor{red}{-14.705} &  \textcolor{red}{67.3678}\\
         6& -46.059 & 76.1462\\
         8& -81.982 &  138.7982\\
         10& 162.167 & 208.8235\\
         25&  -529.769 & 548.3574\\
         35& -858.295 & 756.5167\\
         75& -2020.529 & 1711.8901\\
    \end{tabular}
    \label{tab:nsteps_stats2}
\end{table}

\begin{table}[H]
    \centering
    \caption{p-value for the NT between the \code{n\_steps} instances. The FT is successful with p equal to 0.0. Case \code{n\_steps} equal 4, 6 , and 10 are not statistically significantly different.}
    \begin{tabular}{c|c|c|c|c|c|c|c|c|c}
     \hline
      ID & 1 & 2 & 4 & 6 & 8 & 10 & 25 & 35 & 75 \\
     \hline
     1  & 1.000000 & 0.512019 & \textcolor{red}{0.030146} & \textcolor{red}{0.003953} & 0.257126 & 0.079975 & 0.900000 & 0.900000 & 0.809296 \\ 
     2  & - & 1.000000 & 0.900000 & 0.660658 & 0.900000 & 0.900000 & 0.710204 & 0.099098 & \textcolor{red}{0.009917} \\
     4  & - & - & 1.000000 & 0.900000 & 0.900000 & 0.900000 & 0.079975 & \textcolor{red}{0.001473} & \textcolor{red}{0.001000} \\
     6  & - & - & - & 1.000000 & 0.900000 & 0.900000 & \textcolor{red}{0.013263} & \textcolor{red}{0.001000} & \textcolor{red}{0.001000} \\
     8  & - & - & - & - & 1.000000 & 0.900000 & 0.460015 & \textcolor{red}{0.030146} & \textcolor{red}{0.002063} \\
     10 & - & - & - & - & - & 1.000000 & 0.180790 & \textcolor{red}{0.005415} & \textcolor{red}{0.001000} \\
     25 & - & - & - & - & - & - & 1.000000 & 0.900000 & 0.611110 \\
     35 & - & - & - & - & - & - & - & 1.000000 & 0.900000 \\
    \end{tabular}
    \label{tab:pvalue_nstepsstudy}
\end{table}

Additionally, the characteristics of the best patterns found are given in table \ref{tab:nstepsstudy_bestpattern}.  The dissimilarities are not as high as for the \code{NF} study of appendix \ref{sec:nf_study}, but fine-tuning \code{n\_steps} at a very fine level could have significant consequences. With \code{n\_steps} equal to 6 allows an economy of 0.033 \$/MWh compared to the best solution found with 4, which amounts to about 290,000 - 350,000 $\$/yr$ in benefit for a 1,000 or 1,200 MWe NPP, respectively. In this case, the boron's constraint is tight but the $F_q$'s one has more leeway, and agents trying to optimize or a designer taking over the core found could try to take penalties on the latter to improve the LCOE.

\begin{figure}[H]
    \centering
    \includegraphics[scale = 1.4]{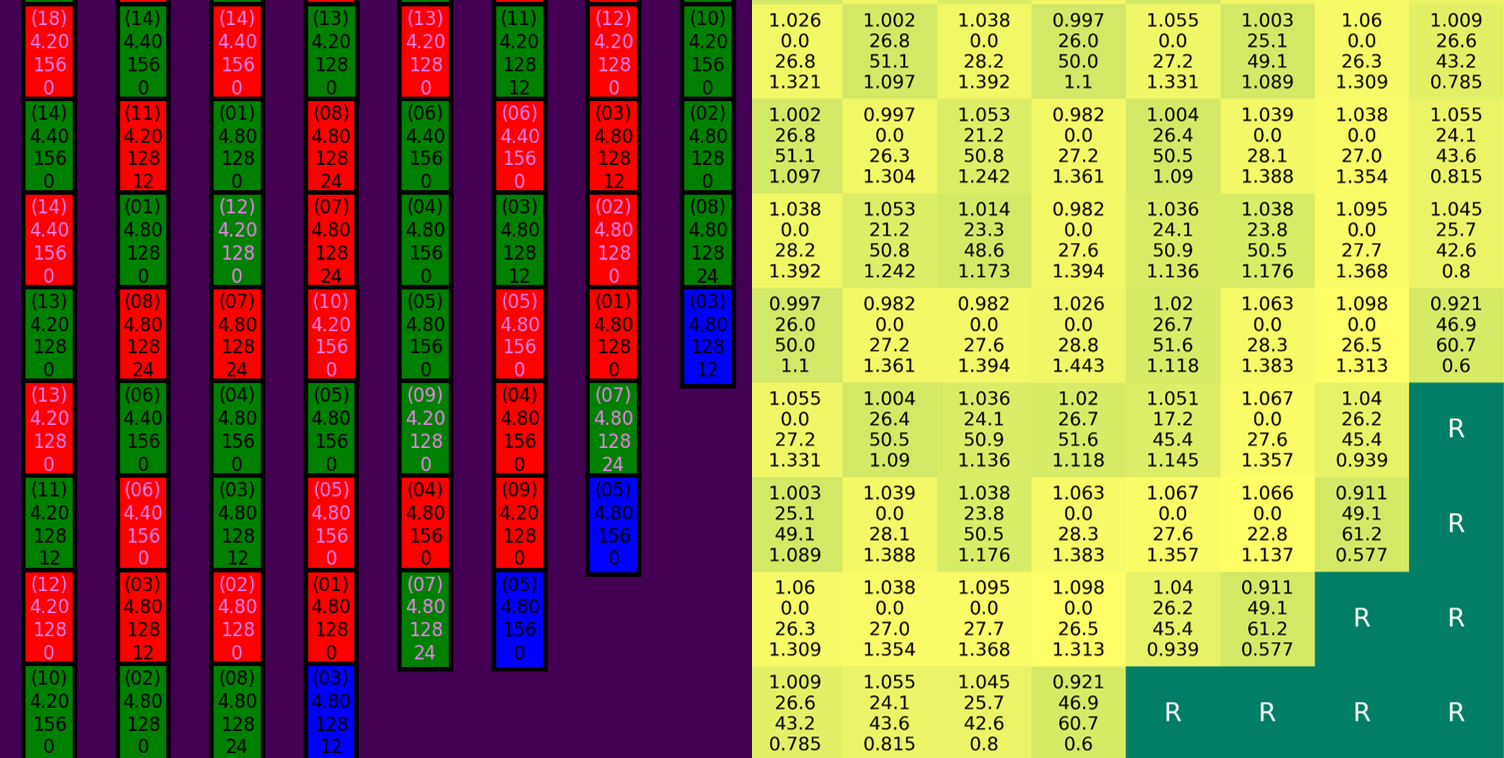}
    \caption{Lower right quadrant for the best \code{n\_steps} case equal to 6. On the left, from top to bottom: Assembly group number, fuel average enrichment, number of IFBA rods, and number of WABA rods. Blue is for twice-burned FAs, green for once-burned, and red for fresh (colors pattern also utilized in ROSA \cite{verhagen1997rosa}). In violet, location where WABA cannot be placed. On the right, from top to bottom: Beginning of cycle $k_{\infty}$ and assembly average burnup, peak pin Burnup at EOC, and $F_{\Delta h}$ when the maximum over the cycle is reached. R means reflector, where the water is.}
    \label{fig:bestpattern_nstespstudy}
\end{figure}

Lastly, the 2D projection of the best pattern found is given in figure \ref{fig:bestpattern_nstespstudy}. The fuel placement is similar to that of figure \ref{fig:bestpattern_nfstudy} but the LCOE is lower. The average enrichment of this core is 4.35\% with an average burnup per assembly at 35.742 GWd/tHm, against 4.4\% and 35.414 GWd/tHm for the previous core. Even though the $F_{\Delta h}$ was lower for the previous core the $F_q$ was higher. Combined with a higher cycle length and lower pin peak Bu, it hints that the power profile must be spread out more evenly hence increasing the average burnup per FAs and the energy extracted at a lower enrichment cost reducing the LCOE.

 \begin{table}[H]
    \centering
     \caption{Best patterns found for each case throughout all experiments. An Objective above - 5.00 signifies that the pattern is feasible and the agents are optimizing the LCOE only.}
    \begin{tabular}{c|c|c|c|c|c|c|c}
      \hline
     \code{n\_steps} & Objective & LCOE & Cycle length & $F_q$ & $F_{\Delta h}$ & $C_b$ & Pin peak $Bu$\\
     \hline
         6    & \textcolor{red}{-4.523} & 5.523 & 513.9 & 1.811 & 1.443 & 1200 & 61.225\\
         10   & -4.540 & 5.540 & 510.3 & 1.813 & 1.444 & 1168.9 & 60.548\\
         8    & -4.544 & 5.540 & 507.8 & 1.822 & 1.449 & 1195.1 & 60.682\\
         2    & -4.551 & 5.628 & 515.7 & 1.822 & 1.446 & 1186.9 & 61.446\\
         4    & -4.556 & 5.556 & 510.6 & 1.841 & 1.438 & 1178.4 & 61.172\\
         25   & -4.620 & 5.623 & 506.3 & 1.825 & 1.432 & 1161.7 & 61.910\\
         1    & -4.616 & 5.616 & 504.7 & 1.832 & 1.449 & 1183.2 & 60.178 \\
         35   & -6.400 & 5.610 & 501.6 & 1.857 & 1.456 & 1186.2 & 60.763\\
         75   & -6.540 & 5.627 & 502.0 & 1.850 & 1.458 & 1146.3 & 62.154\\
   \end{tabular}
    \label{tab:nstepsstudy_bestpattern}
\end{table}

To conclude this section, we saw that \code{n\_steps} is a hyper-parameter that controls the \textit{exploration/exploitation} by playing with the bias/variance trade-offs in the estimation of the gradient of the PPO loss. A very low value may result in instability of the agents and being stuck in local optima. A higher value may result in slow learning and a poorer quality of the best pattern found. Due to the scope of the study (optimization over one day), favoring lower value is the path forward. For a longer run, it might be preferable to utilize larger values up to 10.

\subsection{\code{ent\_coef} study}
\label{sec:entcoef_study}
Lastly, in this section we study the influence of the entropy coefficient \code{ent\_coef}. The entropy term is given by the entropy (a notion originating from information theory) of the distribution satisfied by the policy. This corresponds to the terms $H(\theta)$ and $c_2$ in equation \ref{eq:ppoloss}. The values 0, 0.00001, 0.0001, 0.001, and 0.01 (the default value in stable-baselines) are compared. Greater values than 0.01 were yielding very poor results and were omitted.

\begin{figure}[H]
    \centering
    \includegraphics[scale = 0.6]{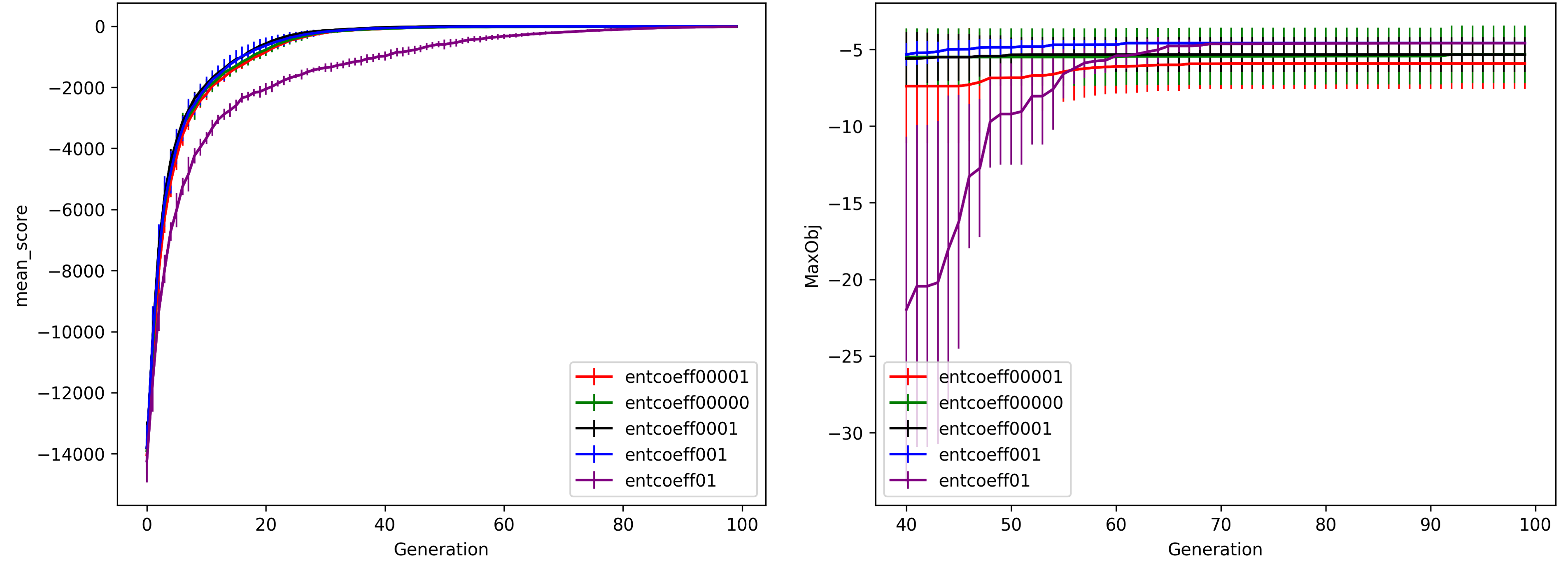}
    \caption{Error bar plots averaged over 10 experiments for Max (left) and Mean (right) Objective per episode for different \code{ent\_coef} values. The Mean was obtained by averaging the mean score obtained per episode within a window of X episodes, where X depends on the total number of samples generated (usually around 400 samples for \code{NF} equal to 1 to obtain 100 generations). The length of the bars is two $\sigma$.}
    \label{fig:entropystudy}
\end{figure}

The analysis of figure \ref{fig:entropystudy} demonstrates the role of the entropy in the \textit{exploration/exploitation} trade-off.  Higher entropy coefficients yield faster convergence and sometimes to lower optima, which is confirmed by the average max objective and mean objective SE columns of tables \ref{tab:entropy_stats} and \ref{tab:entcoef_studystats2}, respectively. The case without entropy has a higher $\sigma$ (MO) between the experiments even though it is supposed to have the highest exploitation. This could be explained by the fact that this entropy coefficient is too low, reaching a different sub-optimal solution anytime depending on the seed that steers initial random actions in different directions. The low mean objective SE and $\sigma$SE confirm this hypothesis. It converged sometimes to great, sometimes to poor local optima demonstrating signs of instability.

With a low mean objective SE but a high $\sigma$SE, value 0.01 is not converging and is still learning but has found the second best pattern overall behind 0.001. The standard deviations ($\sigma$ (MO)) for the cases 0.001 and 0.01 throughout the experiments (table \ref{tab:entropy_stats}) are low compared to the others meaning that they are more stable. Moreover, their $\sigma$SE (table \ref{tab:entcoef_studystats2}) is high, meaning that they are both learning. The sweet spot is therefore located between 0.001 and 0.01. However, the case 0.001 is on average reproducing patterns of high quality (highest Avg (MO) above - 5.00 in table \ref{tab:entropy_stats}) and found an excellent pattern overall with the highest Best (MO). 

 The FT failed with a value of 0.267 and therefore no post-hoc tests were computed. More experiments is needed to reach statistical significance but due to the reasons stated above, we could safely conclude that lowering the entropy coefficient to 0.001 is the way forward.

\begin{figure}[H]
    \centering
    \includegraphics[scale = 0.6]{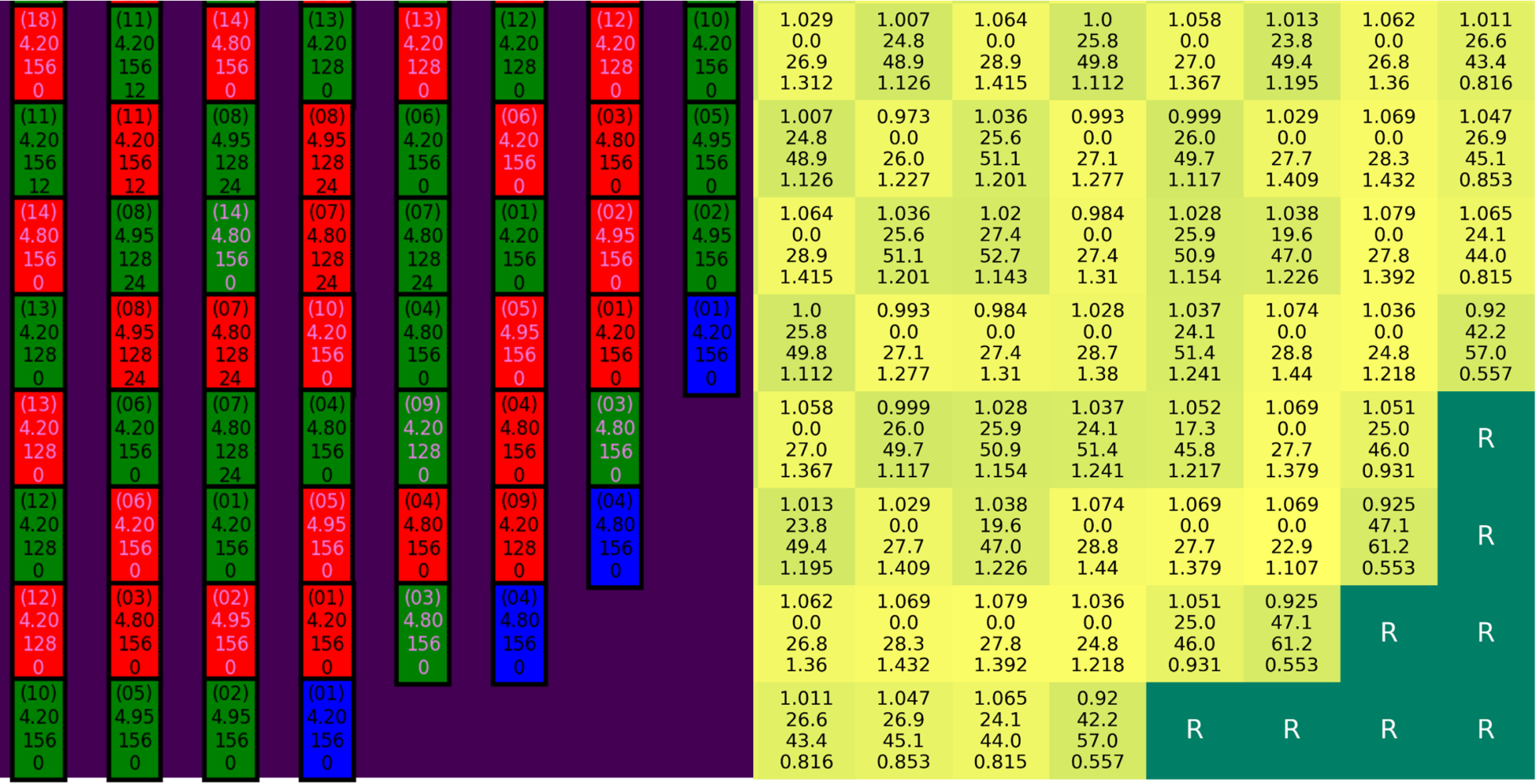}
    \caption{Lower right quadrant of the best pattern found obtained in the \code{ent\_coef} study. On the left, from top to bottom: Assembly group number, fuel average enrichment, number of IFBA rods, and number of WABA rods. Blue is for twice-burned FAs, green for once-burned, and red for fresh (colors pattern also utilized in ROSA \cite{verhagen1997rosa}). In violet, location where WABA cannot be placed. On the right, from top to bottom: Beginning of cycle $k_{\infty}$ and assembly average burnup, peak pin Burnup at EOC, and $F_{\Delta h}$ when the maximum over the cycle is reached. R means reflector, where the water is.}
    \label{fig:bestpattern_entcoefstudy}
\end{figure}

Looking at table \ref{tab:entcoefstudy_bestpattern}, we see that generally again, better LCOE will yield tighter constraints, except for the cycle length which could be higher. The 2D projection on the lower right quadrant of the best pattern overall found (with \code{ent\_coef} equal to 0.001) is given in figure \ref{fig:bestpattern_entcoefstudy}. Similar to the best pattern showcased in figures \ref{fig:bestpattern_nfstudy} and \ref{fig:bestpattern_nstespstudy} we can see typical rings of fire.

 \begin{table}[H]
    \centering
     \caption{Statistics for each \code{ent\_coef} values over 10 experiments. MO stands for Max Objective. The best result for each column is highlighted in red. The Mean Objective SE is obtained by averaging the reward obtained over the last generation and over 10 experiments.}
    \begin{tabular}{c|c|c|c|c}
      \hline
     \code{ent\_coef} & Avg (MO) & Best (MO) & $\sigma$ (MO) & IR (MO)  \\
     \hline
    0.00000 & -5.336 & -4.557 & 1.875 & 0.404 \\
    0.00001 & -5.929 & -4.574 & 1.648 & 0.636 \\
    0.0001  & -5.343 & -4.558 & 1.138 & 0.271 \\
    0.001   & \textcolor{red}{-4.593} & \textcolor{red}{-4.529} & 0.0269  & 0.288 \\
    0.01    & -4.599 & -4.556 & 0.0264 & 0.879 \\
    \end{tabular}
    \label{tab:entropy_stats}
\end{table}
\begin{table}[H]
    \centering
    \caption{Mean Objective SE for the different \code{weights} cases with the $\sigma$SE. The latter is obtained by averaging the reward obtained over the last generation and over 10 experiments. The $\sigma$SE, the standard deviation of the rewards in the last generation is also provided. The case \code{ent\_coef} equal to 0.001 has the highest mean objective SE and a high $\sigma$SE signifying potential for improvement for longer runs.}
    \begin{tabular}{c|c|c}
    \hline
   \code{ent\_coef} &  Mean Objective SE & $\sigma$SE\\
    \hline
       0.00000 &  -16.231& 1.5563\\
       0.00001 & -6.953 & \textcolor{red}{0.2092}\\ 
       0.0001  & -5.691 & 2.5536\\ 
       0.001   & \textcolor{red}{-5.109} & 11.5383\\
       0.01    & -14.705& 67.3678  \\
    \end{tabular}
    \label{tab:entcoef_studystats2}
\end{table}

 \begin{table}[H]
    \centering
    \caption{Best patterns found for each case throughout all experiments. An Objective above - 5.00 signifies that the pattern is feasible and the agents are optimizing the LCOE only. The highest cycle length does not necessarily have the highest LCOE.}
    \begin{tabular}{c|c|c|c|c|c|c|c}
      \hline
     \code{ent\_coef} & Objective & LCOE & Cycle length & $F_q$ & $F_{\Delta h}$ & $C_b$ & Pin peak $Bu$\\
     \hline
         0.001    & \textcolor{red}{-4.529} & 5.529 & 511.9 & 1.850 & 1.440 & 1182.8 & 61.185\\
         0.01    & -4.556 & 5.556 & 510.6 & 1.841 & 1.438 & 1178.4 & 61.172\\
         0.00000    & -4.557 & 5.557 & 510.9 & 1.819 & 1.444 & 1195.0 & 60.287\\
         0.0001    & -4.558 & 5.558 & 516.0 & 1.826 & 1.434 & 1197.2 & 61.035\\
         0.00001   & -4.574 & 5.574 & 510.0 & 1.782 & 1.448 & 1198.9 & 61.256\\
   \end{tabular}
    \label{tab:entcoefstudy_bestpattern}
\end{table}

To conclude this section, we saw that \code{ent\_coef} is a hyper-parameter that controls the \textit{exploration/exploitation} trade-off. Lower values yield faster convergence but are prone to local optima, because the final results may differ depending on the seeding as different directions of the search are early on preferentially locked on. On the other hand, a high value will result in slower convergence. A sweet spot between the speed of convergence, the quality of the pattern found, and the stability of the learning of the agent (i.e., low standard deviation throughout the experiments) was found with a value of 0.001. The latter helps find pattern rapidly but also provide the possibility to find an even better pattern with longer optimization thanks to the reasonably high $\sigma$SE at the last generation.  
\section{Appendix 2: Alternate hyper-parameters studies}
\label{appendix:appendixalternate}
In this appendix we provide the results of the alternate hyper-parameters studies alluded in section \ref{sec:ppo_hyperparameters}. The entropy coefficient \code{ent\_coef} is set at 0.01. The \code{weights}, \code{NF}, and \code{n\_steps} were set at 25000, 1, and 4, respectively.

\subsection{Learning rate \code{LR} study}
\label{appendix:learningra}

\begin{figure}[H]
    \centering
    \includegraphics[scale = 0.5]{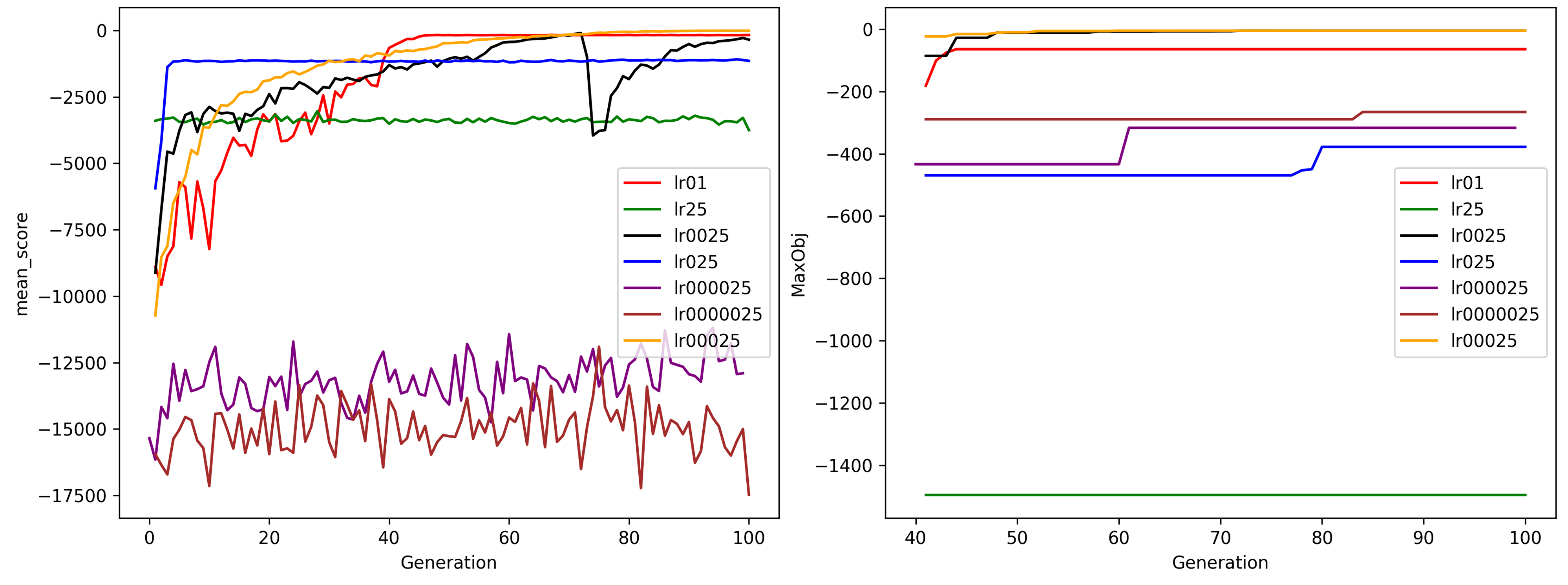}
    \caption{Evolution of the Mean Score per episode (left) and Max Objective (right) overall found by the agents for the \code{LR} study.}
    \label{fig:errorbar_learningratestudy}
\end{figure}

The learning rate is connected to the size of the steps taken in the gradient ascent algorithm that minimizes the loss of equation \ref{eq:ppoloss}. Figure \ref{fig:errorbar_learningratestudy} showcases the evolution of the mean reward per episode of the different values of \code{LR} studied. The behavior of the extreme cases are easily understood. The case 0.25 (green curve on the figure) reach a relatively high mean score early on probably because the learning step is high. However, there is no learning anymore because the steps remains high and the agents cannot manage to improve the pessimistic lower bound. The cases 0.000025 and 0.0000025 (purple and brown on the figure, respectively) are really unstable and are not learning. This may be explained because the learning step is very low. After one generation of samples, barely any learning is happening and the policy remains rather random. Then, new samples are generated randomly, and the distribution of data is probably very different from the previous generation, on which again barely any learning occurs. This then continues which is why we can see such oscillation with very low mean score. The case 0.00025 is then the case with  the smoothest learning. As highlighted in table  \ref{tab:learningrate_study} the closest in performance would be case 0.0025 but its mean objective SE is really low in comparison with the default 0.00025. Therefore we concluded that no study needed to be performed.

\begin{table}[H]
    \centering
    \caption{Figure of Merits for the $\code{LR}$ study. The default case is highlighted in red.}
    \begin{tabular}{c|c|c|c}
    \hline
    Learning Rate $\code{LR}$ values & Best (MO) & Mean Objective SE & $\sigma$SE \\
    \hline
     0.25    &  -1495.6822 &-3287.1906 & 1608.0369 \\
     0.025    &  -377.6574 & -1111.4438 & 381.8044 \\
     0.01   &   -63.9973 & -170.7374 &  57.0793\\
     0.0025     & -4.6830 & -281.1334 & 420.0783\\
     \textcolor{red}{0.00025}   & \textcolor{red}{-4.5820} & \textcolor{red}{-7.3352} & \textcolor{red}{25.7293}\\
     0.000025   &-316.6459 & -12900.4038 &13898.8770\\
     0.0000025  & -265.7414 & -15002.2054 & 18067.5745\\
    \end{tabular}
    \label{tab:learningrate_study}
\end{table}

\subsection{Clipping \code{Clip} study }
\label{appendix:clipstudy}
\begin{figure}[H]
    \centering
    \includegraphics[scale = 0.5]{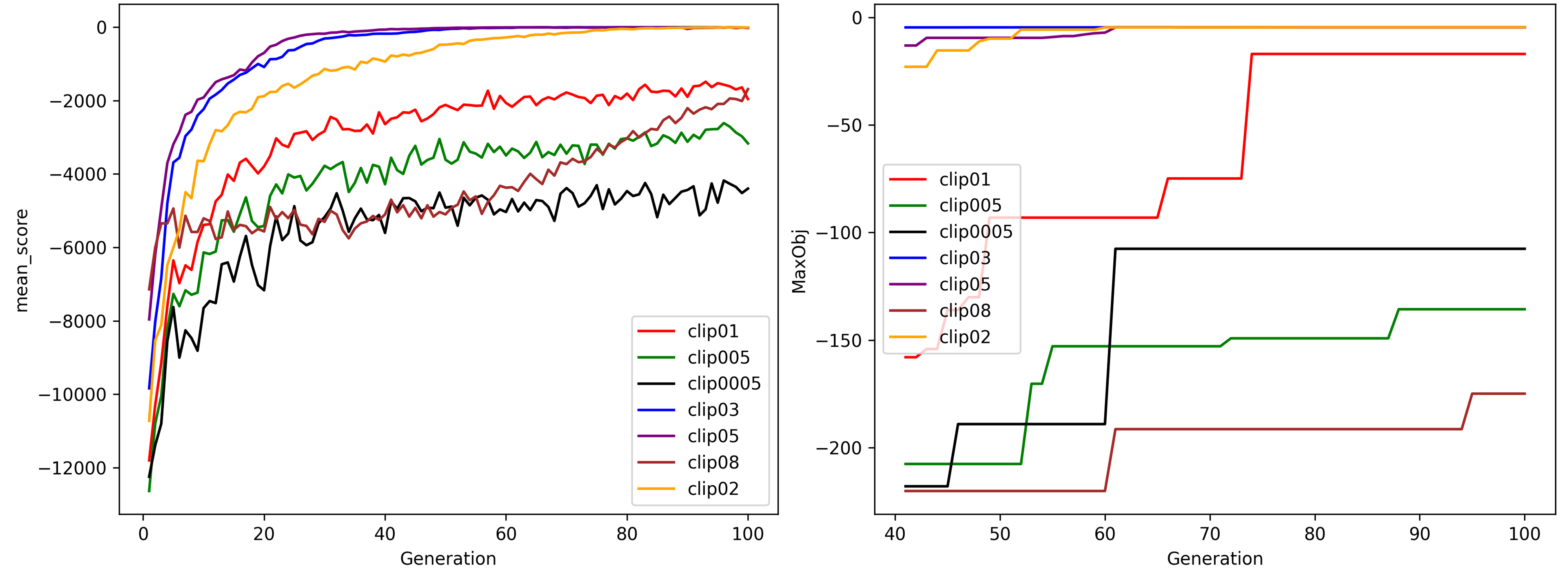}
    \caption{Evolution of the Mean Score per episode (left) and Max Objective (right) overall found by the agents for the $\epsilon$ study.}
    \label{fig:errorbar_clippingstudy}
\end{figure}

The clipping parameter \code{Clip} or $\epsilon$ is the parameter that ensure that the probability ratio $r_t(\theta) = \frac{\pi_{\theta}}{\pi_{\theta_{old}}}$ does not lies outside $[1-\epsilon,1+\epsilon]$ in equation \ref{eq:clipped_obj}. Figure \ref{fig:errorbar_clippingstudy} showcases the evolution of the mean reward per episode of the different values of $\epsilon$ studied. In all cases there seem to be a smooth learning.  Except for case 0.08, the higher the $\epsilon$, the faster the learning early on, although the best mean objective is reached for 0.02. From 0.0005 to 0.02, there seems to be a linear improvement in the final mean score. The lower values enables larger policy updates, which seem to penalize the learning by reaching out local optima, in other word by lacking exploration (recall that the learning is based on the sampled generated by the policy at the previous step). Table \ref{tab:clipping_study} confirms this hypothesis. The case 0.08 seems to prevent learning of the policy by having almost a flat mean score from generation 2 to 45. Then it seems to linearly improve and ends up outperforming all cases from 0.0005 to 0.01. This could be a manifestation of the \textit{exploration/exploitation} trade-off. 0.08 exhibits a high exploration compared to the other cases and after generation 45 it seems that it starts exploiting more. However, its performance remains very poor compared to the case 0.02 to 0.05 and even find the worst pattern overall. The case 0.01 found a relatively good pattern but with regard to its mean objective SE and $\sigma$SE, it is likely to have found it randomly. 

The only cases that compete with the default of 0.02 are 0.03 and 0.05. However, the differences are not significant and 0.02 still exhibits a better behavior. Therefore, we concluded that we did not need to perform a study on this hyper-parameter.

\begin{table}[H]
    \centering
    \caption{Figure of Merits for the $\epsilon$ study. The default case is highlighted in red.}
    \begin{tabular}{c|c|c|c}
    \hline
    Clipping values $\epsilon$ & Best (MO) & Mean Objective SE & $\sigma$SE \\
    \hline
     0.08    &  -174.9414 &  -1688.5742 & 999.7177\\
     0.05    &  -4.6167 & -12.4381 & 30.2795\\
     0.03   &  -4.6125 & -22.5631 & 107.8536\\
     \textcolor{red}{0.02}     & \textcolor{red}{-4.5820} & \textcolor{red}{-9.9267}$^{\star}$ & \textcolor{red}{32.6367}\\
     0.01   & -16.9908 & -1959.3510 & 3409.1235\\
     0.005   & -135.6967 & -3168.8386 & 2934.9369\\
     0.0005  & -107.5879 &  -4398.8450 & 3656.2877\\
    \end{tabular}
    \label{tab:clipping_study}\\
    \begin{tablenotes}
      \item  \footnotesize$^{\star}$The difference between the mean objective SE of table \ref{tab:learningrate_study} and table \ref{tab:clipping_study} is due to the difference between the number of samples generated during the search. The number of generation is normalized to the length of the case with the smallest number of generation (which is sometimes 101 or 100). here, the last mean is then taken at generation number 100 instead of 101 for the \code{LR} study. Interestingly the Mean Objective SE is higher at generation 100, signifying a high variance.
    \end{tablenotes}
\end{table}

\subsection{\code{max\_grad\_norm} study}
\label{appendix:maxgradnormstudy}

\code{max\_grad\_norm} is such that $||\nabla_{\theta}L^{final}|| \le$ \code{max\_grad\_norm}, where $L^{final}$ is the loss of equation \ref{eq:ppoloss}. This is a parameter that control the upper bound of the gradient of the loss in the gradient ascent algorithm. This is a classical technique to avoid catastrophic updates of the weights of a deep learning model, in the context of stochastic gradient descent for instance. Indeed, the estimation of the loss is always local not global, therefore, the steps taken in the training must be controlled. Which is what limiting the gradient does. The clipping $\epsilon$ was initially derived to cover such possibility in PPO \cite{schulman2017proximal} as discussed in appendix \ref{appendix:clipstudy}.  If too many samples separated the different experiments, the mean objective SE and $\sigma$SE could differ too much and we would not be able to fairly compare each one of them. Therefore, we limited the number of samples ran during a day to a similar value of 43,000 to have a more fair comparison.

\begin{figure}[H]
    \centering
    \includegraphics[scale=0.5]{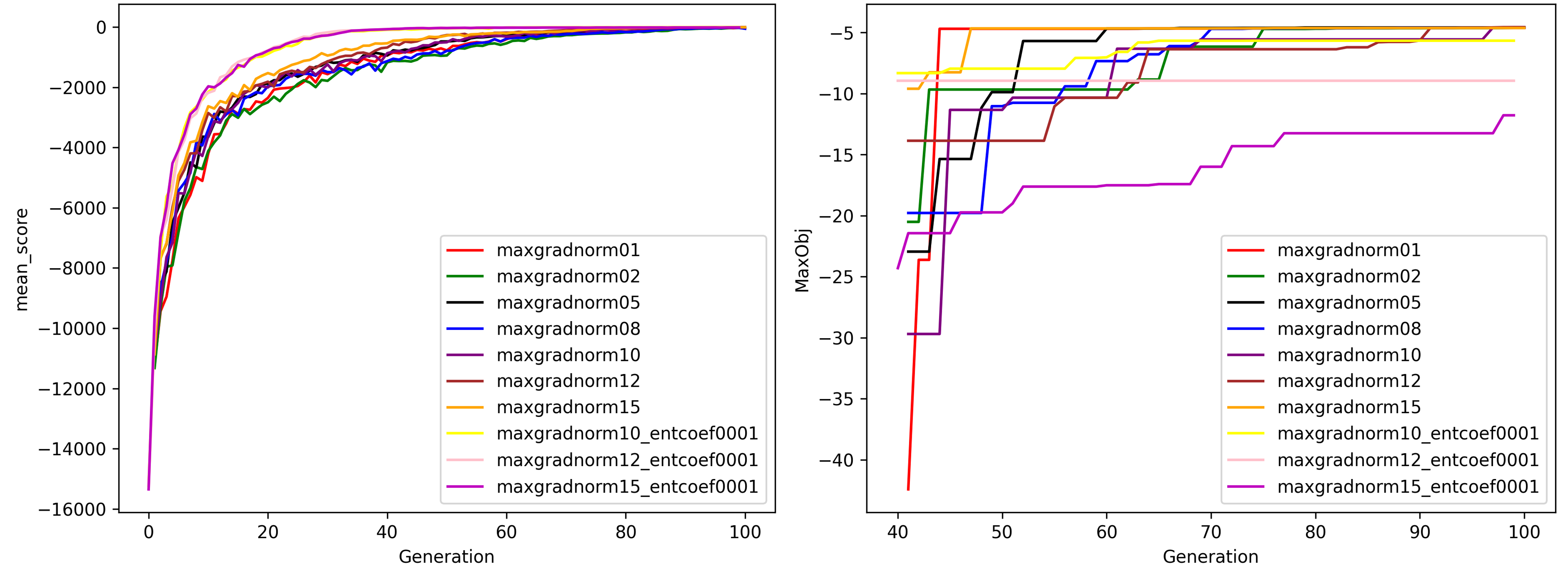}
    \caption{Evolution of the Mean Score per episode (left) and Max Objective (right) overall found by the agents for the \code{max\_grad\_norm} study.}
    \label{fig:maxgradnorm_study}
\end{figure}

Figure \ref{fig:maxgradnorm_study} showcases the evolution of the mean reward per episode of the different values of \code{max\_grad\_norm} studied. In all cases there seem to be a smooth learning and the differences are less drastic as for the $\epsilon$ study.  Most regular cases reached a feasible pattern as displayed in table \ref{tab:maxgradnorm_study}. There are no linear behavior of this parameter and the best cases in terms of SE metrics were 0.50 (the base case), 1.00, and 1.50. However, for 1.00 and 1.50 the $\sigma$SE is rather low and because of our choice of low entropy coefficient (see appendix \ref{sec:entcoef_study}), the final combination if we were to study this hyper-parameter may lead to early convergence to worse local optima. Case 1.20 could also be competitive, as its SE are close to that of case 0.5. Therefore, we decided to also study three cases where the entropy coefficient \code{ent\_coef} was divided by 10 (last three rows of table \ref{tab:maxgradnorm_study}). As expected, the $\sigma$SE is low as well as the Mean objective SE for cases 1.0 and 1.50. For case 1.20, the agent converged to a local optima with a very low $\sigma$SE. These values are to be compared with table \ref{tab:entcoef_studystats2}, where the best case presented a higher mean objective SE and higher $\sigma$SE. Therefore, we decided to not delve into more studies for \code{max\_grad\_norm}.

\begin{table}[H]
    \centering
    \caption{Figure of Merits for the \code{max\_grad\_norm} study. The default case is highlighted in red.}
    \begin{tabular}{c|c|c|c}
    \hline
     \code{max\_grad\_norm} values & Best (MO) & Mean Objective SE & $\sigma$SE \\
    \hline
     0.10   &-4.6085 & -22.6228 & 67.2356 \\
     0.20   &-4.6057& -27.8003 & 56.1412 \\
     \textcolor{red}{0.50}   &\textcolor{red}{-4.5820}& \textcolor{red}{-7.3352} &  \textcolor{red}{25.7293} \\
     0.80   &-4.6080& -25.8551 &  76.1472 \\
     1.00   &-4.5701& -7.6252 &  9.4603 \\
     1.20   &-4.5578& -8.7399 &  36.3495 \\
     1.50   &-4.6241& -6.1963 & 12.7045 \\
     1.0  + \code{ent\_coef} = 0.001 &-5.6643& -5.9101 & 4.9734 \\
     1.2  + \code{ent\_coef} = 0.001 &-8.9469& -9.9184 & 0.3805 \\
     1.5  + \code{ent\_coef} = 0.001 & -11.7771 & -14.9618 & 7.7478 \\
    \end{tabular}
    \label{tab:maxgradnorm_study}
\end{table}

\subsection{\code{noptepochs} study}
\label{appendix:noptepochsstudy}
\begin{figure}[H]
    \centering
    \includegraphics[scale = 0.5]{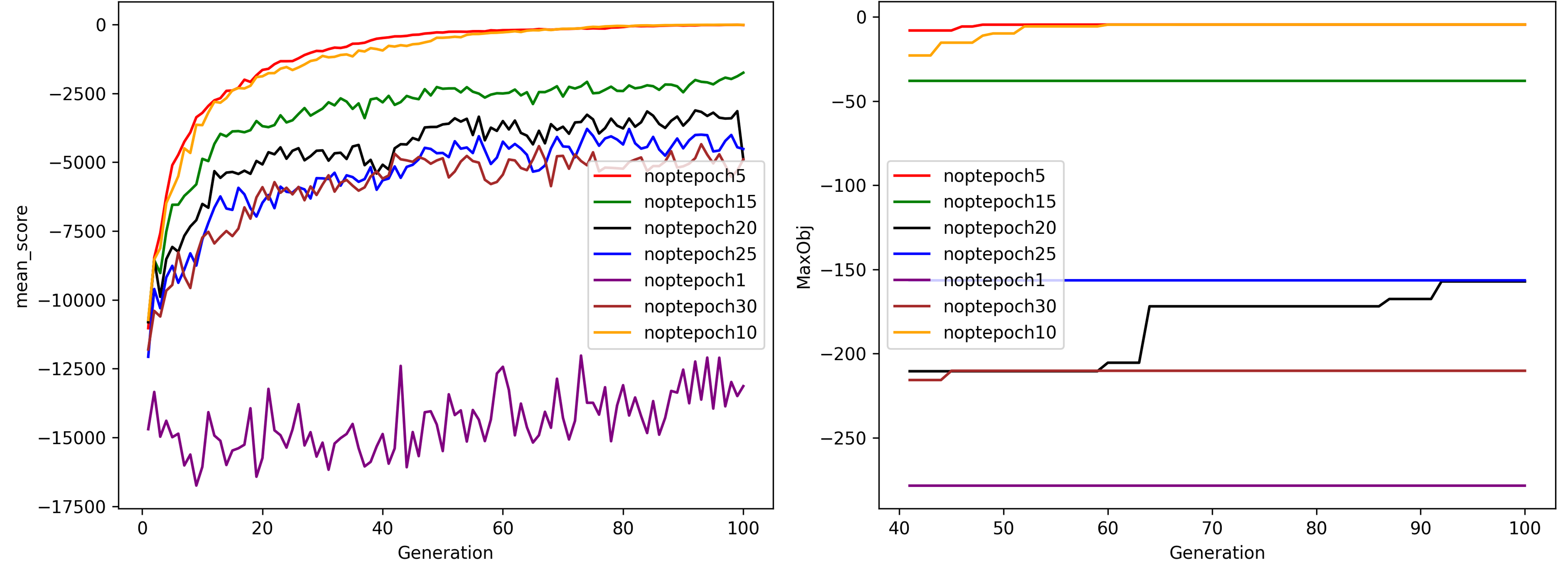}
    \caption{Evolution of the Mean Score per episode (left) and Max Objective (right) overall found by the agents for the \code{noptepochs} study.}
    \label{fig:errorbar_noptepochsstudy}
\end{figure}

The parameter \code{noptepochs} is the parameter that control the number of steps in the gradient ascent algorithm to maximize the surrogate loss of equation \ref{eq:ppoloss}. Figure \ref{fig:errorbar_noptepochsstudy} showcases the evolution of the mean reward per episode of the different values of \code{noptepochs} studied. There is a smooth learning for all cases except for case 1 (purple on the left hand-side of figure \ref{fig:errorbar_noptepochsstudy}). This might be due to the fact that one step only prevents any learning and the weights of the neural architecture representing the policy are not converging to anything. The cases 5 and 10 largely outperform the others. This is confirmed by table \ref{tab:noptepochs_study}. 

On the other hand, early on, lower \code{noptepochs} leads to faster learning. This may be counter-intuitive, as higher \code{noptepochs} means more exploitation of the information given by the samples generated by the policy. However, one could imagine that the original samples generated by the un-trained policy are random or at least representing low quality patterns. Therefore, the cases with higher \code{noptepochs} are exploiting random or low-quality actions resulting in poor learning. From case 10 to 30 the $\sigma$SE increases, indicating that more epochs leads to more exploration. This is the same for the Best (MO) and almost the same for the mean objective SE with the exception of case 20. However, the figure on the left of figure \ref{fig:errorbar_noptepochsstudy} seems to indicate a random local drop in performance at the last generation for case 20, which may be due to the randomness of the policy. Table \ref{tab:lossrelated_fomsnoptepochs} seems to hint about why higher \code{noptepochs} leads to higher $\sigma$SE. The \code{clipfrac} increases with \code{noptepochs}. This might be explained by the fact that the policy is more different with more steps in the gradient ascent, hence increasing the likelihood of changing the policy more. Therefore, the steps taken are less controlled yielding instability in learning. As a consequence, the entropy remains high (see table \ref{tab:lossrelated_fomsnoptepochs}) and the policy remains more random. The sweet spot is then between 5 and 10.

Only the case 5 competed with the default case 10, but the differences were not significant enough and we decided to not delve into any more study.

\begin{table}[H]
    \centering
    \caption{Figure of Merits for the \code{noptepochs} study. The default case is highlighted in red.}
    \begin{tabular}{c|c|c|c}
    \hline
    \code{noptepochs} values & Best (MO) & Mean Objective SE & $\sigma$SE \\
    \hline
     1    &  -278.2698 & -13133.0550 & 18291.9926\\
     5    & -4.6176 & -18.0850 & 36.5557\\
     \textcolor{red}{10}   &  \textcolor{red}{-4.5820} & \textcolor{red}{-9.9267} & \textcolor{red}{32.6367}\\
     15     & -38.1449 & -1747.9683 & 1355.6888\\
     20   & -157.1052 & -4890.5444 & 3636.5023\\
     25   & -156.3375 &-4518.5616 & 4608.1392\\
     30  & -210.1378 & -4894.9650 & 5039.9094\\
    \end{tabular}
    \label{tab:noptepochs_study}
\end{table}

\begin{table}[H]
    \centering
    \caption{Loss related Figure of Merits for \code{noptepochs}. \code{clipfrac} is the fraction for which the probability ratio is outside the $[1-\epsilon,1+\epsilon]$ range. The default case is highlighted in red.}
    \begin{tabular}{c|c|c|c|c|c}
    \hline
    \code{noptepochs} & \code{approx\_kl} & \code{clipfrac} & $\mathbb{H}$ & $L^{CLIP}$ & \code{VF\_loss} \\
    \hline
    1 & 1.2905e-7 & 0.0 & 67.49082 & -6.8025e-5 & 136777920.0\\
    5 & 6.944e-4 & 0.0 & 28.1128 & -8.3063e-3 & 42998.188\\
    \textcolor{red}{10} & \textcolor{red}{2.8199e-2} & \textcolor{red}{0.2875} & \textcolor{red}{44.3150} & \textcolor{red}{-4.0714e-2} & \textcolor{red}{218030.05}\\
    15 & 1.0937e-1 & 0.5615 & 62.2872 & -9.7497e-2 & 2916480.0\\
    20 & 1.0234e-1 & 0.6555 & 65.1677 & -9.7205e-2 & 13611875.0\\
    25 & 1.1506e-1 & 0.7391 & 65.6561 & -1.1411e-1 & 16522170.0\\
    30 & 1.1097e-1 & 0.7560 & 65.8907 & -1.0928e-1 & 25375224.0\\
    \end{tabular}
    \label{tab:lossrelated_fomsnoptepochs}
\end{table}
\begin{figure}[H]
    \centering
    \includegraphics[scale=0.9]{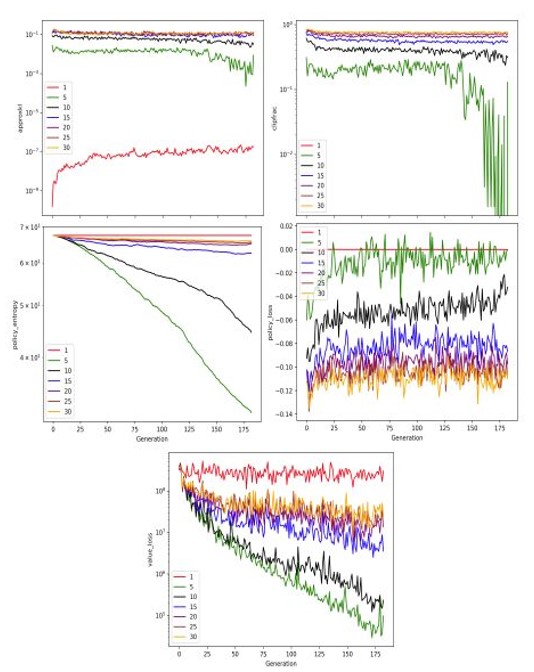}
    \caption{Evolution of the loss-related FOMs for the \code{noptepochs} study. The results are plotted on a log scale for all except for the policy loss $L^{CLIP}$. For cases 5 and 10 the losses evolve as expected while for the other cases the learning is much less smooth.}
    \label{fig:noptepochs_lossFOM}
\end{figure}

\subsection{\code{nminibatches} study}
\label{appendix:nminibatchesstudy}
\begin{figure}[H]
    \centering
    \includegraphics[scale=0.5]{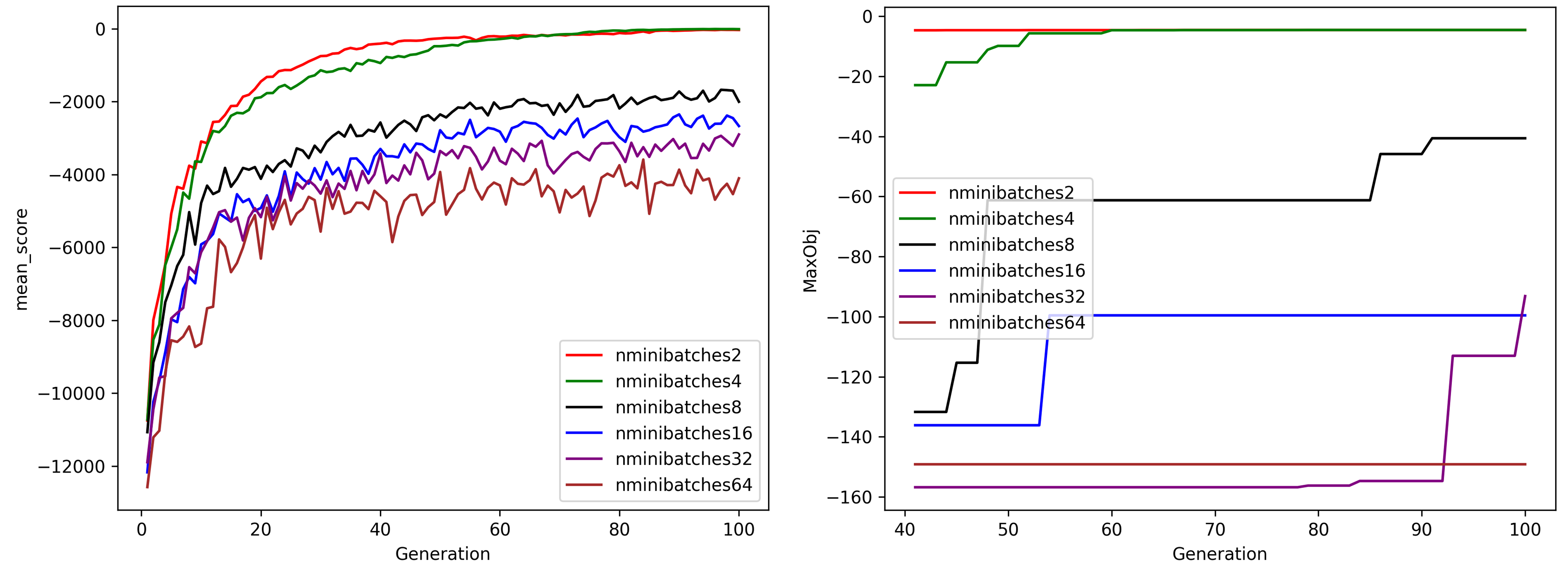}
    \caption{Evolution of the Mean Score per episode (left) and Max Objective (right) overall found by the agents for the \code{nminibatches} study.}
    \label{fig:nminibatches_study}
\end{figure}

\code{nminibatches} (line 16 or algorithm \ref{alg:ppopseudocode}) is the number of mini batches used in training during a step of the algorithm. Essentially, the size of the batches for training is equal to $\frac{\text{\code{n\_steps}} \times \text{\code{ncores}}}{\text{\code{nminibatches}}}$. As showcased in figure \ref{fig:nminibatches_study}, all cases exacerbate a smooth learning. We stopped at 64 because there are no observed policy learning for the case 128 and greater values would not be possible since \code{n\_steps} $\times$ \code{ncores} is equal to the latter.

Overall, a higher value of this parameter lead to slower learning. However, there seem to be a sweet spot for the case equal to 4. There, the size of the batches is equal to the number of cores. Both SE measures are better even though the case 2 found a high quality pattern with only a difference in magnitude about 0.01 in the LCOE. It could be interesting to see what happens when the ratio  $\frac{\text{\code{n\_steps}}}{\text{\code{nminibatches}}}$ is kept constant and we left that for future work. Above 4, the behavior seems to remain linear. Only 16 found a worse best pattern than 32, but regarding the high $\sigma$SE of table \ref{tab:nminibatches_study}, it must have been found randomly.

\begin{table}[H]
    \centering
    \caption{Figure of Merits for the \code{nminibatches} study. The default case is highlighted in red.}
    \begin{tabular}{c|c|c|c}
    \hline
    \code{nminibatches} values & Best (MO) & Mean Objective SE & $\sigma$SE \\
    \hline
     2   &-4.5918   &-31.8550 & 95.3676\\
     \textcolor{red}{4}   &\textcolor{red}{-4.5820}   & \textcolor{red}{-9.9267}  & \textcolor{red}{32.6367}\\
     8   &-40.6294  &-2002.2990  & 2694.3910\\
     16  &-99.5715  &-2669.2012  & 2501.2163\\
     32  &-93.1650  &-2900.84154  & 2621.0595\\
     64  &-149.1445 &-4104.2863 & 4151.5750\\
     128 &-287.3326 &-16225.4743 & 20260.0626\\
    \end{tabular}
    \label{tab:nminibatches_study}
\end{table}

\subsection{$\lambda$ and $\gamma$ studies}
\label{appendix:lambdagammastudy}
In PPO, the advantage function is estimated via Generalized Advantage Estimation (GAE) \cite{schulman2017proximal}. 
The latter represents an exponential average of the advantage function with parameter $\lambda$. This estimate suffers from the bias-variance trade-off and the $\lambda$ parameter helps control it. Greater value of $\lambda$ will decrease the bias of the estimator by allow greater contribution to later states (up to max(\code{NF},\code{n\_steps}) for each agent) at the cost of increasing the variance due to the presence of more terms. Because the length of the episode is 1, neither $\lambda$ or $\gamma$ should influence the algorithm. The only differences observed by running different values must be due to the randomness of the policies resulting in slightly different SE measures. Indeed, the advantage estimation becomes \cite{schulman2018high}:
\begin{equation} A_{\pi_{\theta}}(s_0,a_0;\theta) = r_0 + \gamma \mathbb{E}(V(s_1)) - V(s_0) = r_0 - V(s_0)\end{equation}

$\gamma$ is the discount factor in the cumulative reward, which is a sum over one term. Therefore, no study are presented here.

\subsection{\code{VF\_coef} study}
\label{appendix:vfcoefstudy}
\begin{figure}[H]
    \centering
    \includegraphics[scale=0.5]{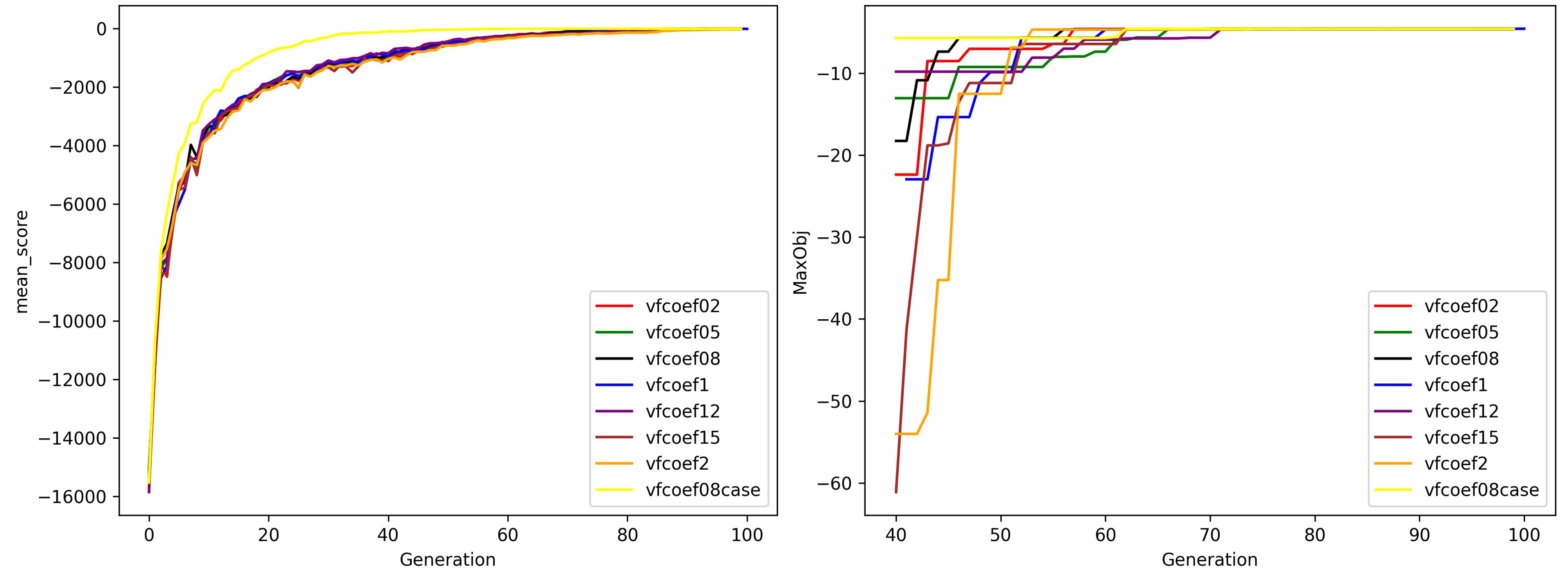}
    \caption{Evolution of the Mean Score per episode (left) and Max Objective (right) overall found by the agents for the \code{VF\_coef} study.}
    \label{fig:vfcoef_study}
\end{figure}

\code{VF\_coef} is the weight assigned to the value function's term in the PPO loss of equation \ref{eq:ppoloss}. The algorithm was very sensitive to this term. It turned out that the number of samples for this particular case in the configuration studied had a big impact on the figure of merits found. We used 43,000 as the maximum number of samples. Figure \ref{fig:vfcoef_study} showcases the evolution of the mean score per episode and max objective for the \code{VF\_coef} study.

In all cases, there is a smooth learning. Case 2 finds a feasible pattern slower than the others, as if it spent more time exploring. It is interesting to see that, except for case 1.5, the lower the \code{VF\_voef}, the higher the mean objective SE and the lower the $\sigma$SE as displayed in table \ref{tab:vfstudy_study}. Reducing the impact of the value function in the loss seems to increase the exploitation. We first hypothesized that reducing the impact of the value function would increase the speed of the convergence of the policy loss. Therefore, we tried to find an explaination looking at the loss FOMsuch as in table \ref{tab:lossrelated_fomsnoptepochs} via table \ref{tab:lossrelated_fomsvfcoef}. Nevertheless, no clear behaviors were drawn. The impact of \code{VF\_coef} cannot be explained by only looking at loss-related performance metrics, especially with the particular case 1.5. Several interesting explanation removing case 1.5 can be however made. Comparing the entropy of case 0.2 to 1.0 versus 1.2, we can say that a lower entropy does not result in a high mean objective or $\sigma$SE. Similarly, for 0.2 to 0.8, a better policy or value function loss results in a higher mean objective SE and lower $\sigma$SE. The case 1.0 and 2.0 both have a better policy loss but a much worse value function loss, resulting in a lower mean objective SE and higher $\sigma$SE. It seems that there is a trade-offs between the losses but that the value loss has a greater impact. This is confirmed by comparing 2.0 and 1.2. 2.0 exacerbates a worse SEs even though the policy losses improve about twice as much.

\begin{table}[H]
    \centering
    \caption{Figure of Merits for the \code{VF\_coef} study. The default case is highlighted in red.}
    \begin{tabular}{c|c|c|c}
    \hline
    \code{VF\_coef} values & Best (MO) & Mean Objective SE & $\sigma$SE \\
    \hline
0.20&-4.6097 & -5.71826 & 10.7914\\
0.5&-4.5756 & -5.6684 & 9.6340 \\
0.8&-4.5967 & -6.2679 & 15.4157\\
\textcolor{red}{1.0}&\textcolor{red}{-4.5820} & \textcolor{red}{-7.3352} &\textcolor{red}{ 25.7293}\\
1.2&-4.6216 & -11.2973 & 44.7394\\
1.5&-4.6335 & -7.3875 & 20.6081\\
2.0&-4.6148 & -22.6870 & 45.4516\\
0.8 + \code{ent\_coef} = 0.001 & 4.59014 & -4.863 & 5.4473 \\
    \end{tabular}
    \label{tab:vfstudy_study}
\end{table}

\begin{table}[H]
    \centering
    \caption{Loss related Figure of Merits for \code{VF\_coef}. The default case is highlighted in red.}
    \begin{tabular}{c|c|c|c}
    \hline
    \code{VF\_coef} & $\mathbb{H}$ & $L^{CLIP}$ & \code{VF\_loss} \\
    \hline
    0.2 & 9.0399 & -0.0372 & 468.7824 \\
    0.5 & 12.3117 & -0.0346 & 872.3915 \\
    0.8 & 3.9508 & -0.0438 & 573.3726 \\
    \textcolor{red}{1.0} & \textcolor{red}{4.0087} & \textcolor{red}{-0.0274} & \textcolor{red}{1248.4656} \\
    1.2 & 5.7303 & -0.0507 & 1588.7722 \\
    2.0 & 15.1559 & -0.0135 & 2906.0786 \\
    0.8  + \code{ent\_coef} = 0.001 & 0.0696 & -0.0046 & 0.64881 \\
    \end{tabular}
    \label{tab:lossrelated_fomsvfcoef}
\end{table}

A nice trade-off could be to utilize a value of 0.8 for \code{VF\_coef} because the mean objective SE is higher than for the case 1.0 and the best pattern found is very close. Nevertheless, the $\sigma$SE is at 15.4157 and we made the choice of reducing the entropy in appendix \ref{sec:entcoef_study}, which already reduced the $\sigma$SE with a better mean objective SE. We would risk to reach a worse local optima if we combined it with a lower \code{VF\_coef} value. To verify that, we also ran a case with \code{VF\_coef} and \code{ent\_coef} equal to 0.8 and 0.001, respectively (last row of table \ref{tab:vfstudy_study}). There, the best pattern found was worse than the base case, and the $\sigma$SE is much lower. Keep in mind that these hyper-parameters may be used to pursue further study and the latter value is too low, meaning that the algorithm was already too close to converge. Therefore, we decided not to pursue deeper studies with this parameter or another combination with \code{ent\_coef}.

\subsection{Depth and breadth of the architecture studies}
\label{appendix:architecturestudy}
\begin{figure}[H]
    \centering
    \includegraphics[scale = 0.5]{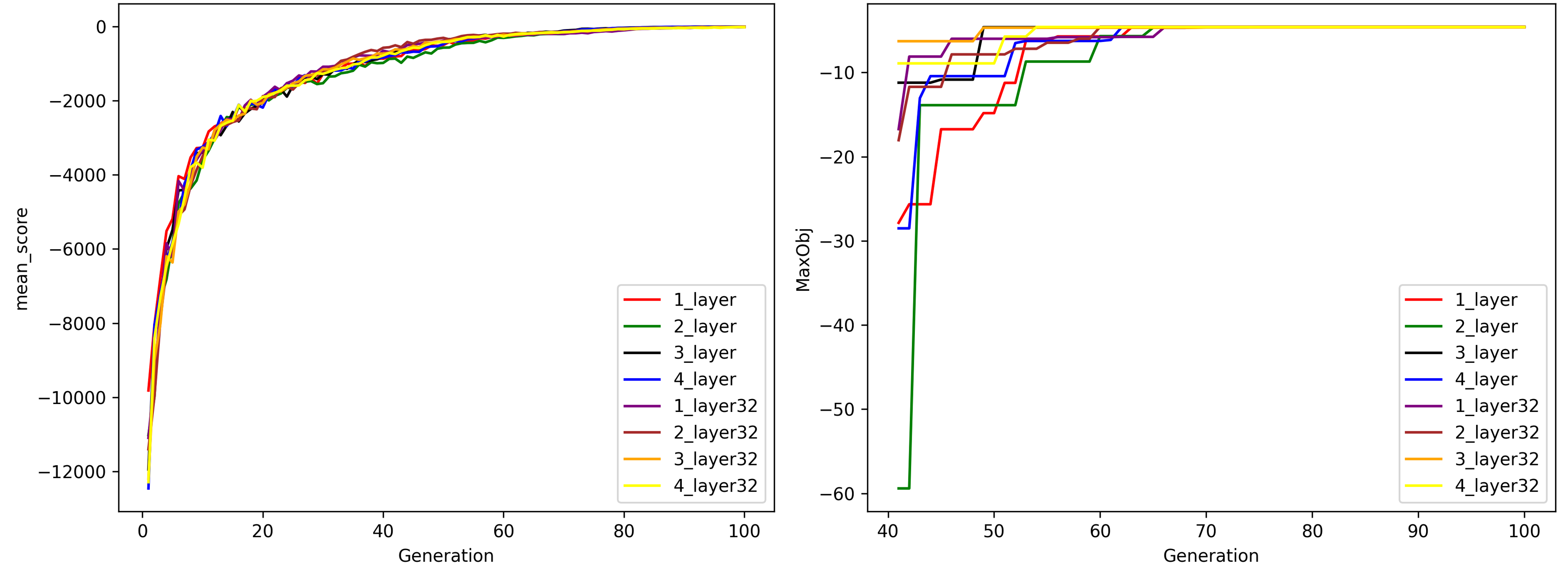}
    \caption{Evolution of the Mean Score per episode (left) and Max Objective (right) overall found by the agents for the depth and breadth of the architecture studies.}
    \label{fig:errorbar_architecturestudy}
\end{figure}

The depth of the neural architecture representing the policy corresponds to the number of hidden layer in the neural network. The breadth is the number of hidden nodes per layer. The number of hidden nodes is equal to 64 or 32.  Figure \ref{fig:errorbar_architecturestudy} showcases the evolution of the mean reward per episode of the different values of the depth and breadth studied. In all cases there is a smooth learning. The effect on the number of hidden layer is small compared to the other hyper-parameters studied in this appendix. The breadth has an impact on the learning as for each case the mean objective SE and $\sigma$SE are lower and higher, respectively, as shown in table \ref{tab:architecture_study}. The best patterns found seem to also be better for smaller number of nodes. 

Another interesting behavior is that the best patterns are found faster when the number of hidden nodes is divided by two as confirmed by figure \ref{fig:errorbar_architecturestudy}. However, the mean objective SE is smaller and the $\sigma$SE higher. In this case what might have happened is that it took more samples for cases with 64 hidden nodes to learn to reproduce better patterns but they managed to produce pattern closer to a better optima because of a better understanding of the search space. The number of hidden nodes therefore plays a role in the \textit{exploration/exploitation} trade-off.

No cases are close to convergence except case 4. There the mean objective SE is lower than - 5.00, meaning that a feasible pattern is not always found in average. What might have happened is that it overfitted towards lower optima due to its larger number of trainable parameters. On top of that, the behavior of the number of hidden layer is opposite when having 64 or 32 nodes. For cases with 64 nodes, the cases 2 and 3 have a higher mean objective SE and $\sigma$SE, which is the opposite for cases with 32 hidden nodes. Moreover, the case 4 + 32 is really far from convergence and has found an excellent best pattern very rapidly. It is as if removing nodes per layer smoothed out its learning and enable to escape local optima.  The default value of 2 exacerbates a higher mean objective SE and $\sigma$SE, promising more potential for improvement and stability if utilized for further optimization. Overall, most of the cases competed with the default case 2 but the differences are not large enough and we concluded that we did not need to delve into more studies.

\begin{table}[H]
    \centering
    \caption{Figure of Merits for the depth and breadth of the architecture studied. Cases 1, 2, 3, and 4 have 64 hidden nodes per layer. The default case is highlighted in red. 43,000 samples were drawn.}
    \begin{tabular}{c|c|c|c}
    \hline
    Depth values $\epsilon$ & Best (MO) & Mean Objective SE & $\sigma$SE \\
    \hline
     1      & -4.5892                            & -6.8169 & 19.2774\\
     \textcolor{red}{2}      & \textcolor{red}{-4.6006$^{\star}$} & \textcolor{red}{-7.9264} & \textcolor{red}{23.4465}\\
     3      & -4.6101                            & -6.2860 & 26.1353\\
     4      & -4.6158                            & -5.0907 & 5.63168\\
     1 + 32 & -4.5999                            & -22.4148 & 59.2805\\
     2 + 32 & -4.5699                            & -13.1631 & 22.2361\\
     3 + 32 & -4.6480                            & -16.9550 & 39.3686\\
     4 + 32 & -4.5973                            & -33.7441 & 194.4261\\
     \end{tabular}
    \label{tab:architecture_study}\\
    \begin{tablenotes}
      \item  \footnotesize$^{\star}$The difference between the average standard deviation in table \ref{tab:architecture_study} and table \ref{tab:entcoef_studystats2} is that here the experiment were ran once with more samples. Looking at experiment per experiment, in some instances, the algorithm ran longer and a generation covered more time-steps, increasing the $\sigma$SE.
    \end{tablenotes}
\end{table}

%% file: Manuscripts.bbl
\begin{thebibliography}{10}
\newcommand{\enquote}[1]{``#1''}
\providecommand{\url}[1]{\texttt{#1}}
\providecommand{\urlprefix}{URL }

\bibitem{nei2020nuclear}
NEI.
\newblock \enquote{NUCLEAR COSTS IN CONTEXT.}
\newblock \emph{Nuclear Energy Institute, NEI} (2020).

\bibitem{kropaczek2011copernicus}
D.~J. Kropaczek.
\newblock \enquote{COPERNICUS: A multi-cycle optimization code for nuclear fuel
  based on parallel simulated annealing with mixing of states.}
\newblock \emph{Progress in Nuclear Energy}, \textbf{volume~53}(6), pp.
  554--561 (2011).

\bibitem{park2009multiobjective}
T.~K. Park, H.~G. Joo, and C.~H. Kim.
\newblock \enquote{Multiobjective Loading Pattern Optimization by Simulated
  Annealing Employing Discontinuous Penalty Function and Screening Technique.}
\newblock \emph{Nuclear Science and Engineering}, \textbf{volume 162}, pp.
  134--147 (2009).

\bibitem{parks1996multiobjective}
G.~T. Parks.
\newblock \enquote{Multiobjective pressurized water reactor reload core design
  by nondominated genetic algorithm search.}
\newblock \emph{Nuclear Science and Engineering}, \textbf{volume 124}(1), pp.
  178--187 (1996).

\bibitem{moura2019optimization}
A.~A. de~Moura~Meneses, L.~M. Araujo, F.~N. Nast, P.~Vasconcelos~da Silva, and
  R.~Schirru.
\newblock \emph{Optimization of Nuclear Reactors Loading Patterns with
  Computational Intelligence Methods. In: Platt G., Yang XS., Silva Neto A.
  (eds) Computational Intelligence, Optimization and Inverse Problems with
  Applications in Engineering.}
\newblock Springer, Cham (2019).

\bibitem{meneses2015acrossentropy}
A.~de~Moura~Meneses and R.~Schirru.
\newblock \enquote{A cross-entropy method applied to the In-core fuel
  management optimization of a Pressurized Water Reactor.}
\newblock \emph{Progress in Nuclear Energy}, \textbf{volume~83}, pp. 326--335
  (2015).

\bibitem{lima2008anuclear}
A.~M.~M. de~Lima, F.~C. Schirru, R. da~Silva, and J.~A. C.~C. Medeiros.
\newblock \enquote{A nuclear reactor core fuel reload optimization using
  artificial ant colony connective networks.}
\newblock \emph{Annals of Nuclear Energy}, \textbf{volume~35}, pp. 1606--1612
  (2008).

\bibitem{wu2016quantum}
S.~C. Wu, T.~H. Chan, M.~S. Hsieh, and C.~Lin.
\newblock \enquote{Quantum evolutionary algorithm and tabu search in
  pressurized water reactor loading pattern design.}
\newblock \emph{Annals of Nuclear Energy}, \textbf{volume~94}, pp. 773--782
  (2016).

\bibitem{lin2014themaxmin}
S.~Lin and Y.~H. Chen.
\newblock \enquote{The max–min ant system and tabu search for pressurized
  water reactor loading pattern design.}
\newblock \emph{Annals of Nuclear Energy}, \textbf{volume~71}, pp. 388--398
  (2014).

\bibitem{erdogan2003apwr}
A.~Erdo\u{g}an and M.~Ge\c{c}kinli.
\newblock \enquote{A PWR reload optimisation (Xcore) using artificial neural
  network and genetic algorithm.}
\newblock \emph{Annals of Nuclear Energy}, \textbf{volume~30}, pp. 35--53
  (2003).

\bibitem{li2022comparative}
Z.~Li, J.~Huang, J.~Wang, and M.~Ding.
\newblock \enquote{Comparative study of meta-heuristic algorithms for reactor
  fuel reloading optimization based on the developed BP-ANN calculation
  method.}
\newblock \emph{Annals of Nuclear Energy}, \textbf{volume 165}, p. 108685
  (2022).

\bibitem{ortiz2004using}
J.~J. Ortiz and I.~Requena.
\newblock \enquote{Using a multi-state recurrent neural networks to optimize
  loading patterns in BWRs.}
\newblock \emph{Annals of Nuclear Energy}, \textbf{volume~31}, pp. 789--803
  (2004).

\bibitem{yamamoto2003application}
A.~Yamamoto.
\newblock \enquote{Application of Neural Network for Loading Pattern Screening
  of In-Core Optimization Calculations.}
\newblock \emph{Nuclear Technology}, \textbf{volume 144}(1), pp. 63--75 (2003).

\bibitem{gocalvez2006sensitivity}
J.~M. Gozalvez, S.~Yilmaz, F.~Alim, K.~Ivanov, and S.~H. Levine.
\newblock \enquote{Sensitivity study on determining an efficient set of fuel
  assembly parameters in training data for designing of neural networks in
  hybrid genetic algorithms.}
\newblock \emph{Annals of Nuclear Energy}, \textbf{volume~33}, pp. 457--465
  (2006).

\bibitem{bello2016neural}
I.~Bello, H.~Pham, Q.~V. Le, M.~Norouzi, and S.~Bengio.
\newblock \enquote{Neural combinatorial optimization with reinforcement
  learning.}
\newblock \emph{arXiv preprint arXiv:161109940} (2016).

\bibitem{khalil2017learning}
E.~Khalil, H.~Dai, Y.~Zhang, B.~Dilkina, and L.~Song.
\newblock \enquote{Learning combinatorial optimization algorithms over graphs.}
\newblock In \emph{NIPS'17: Proceedings of the 31st International Conference on
  Neural Information Processing Systems}, pp. 6348--6358 (2017).

\bibitem{li2021deep}
K.~Li, T.~Zhang, and R.~Wang.
\newblock \enquote{Deep Reinforcement Learning for Multi-Objective
  Optimization.}
\newblock \emph{IEEE Transactions on Cybernetics}, \textbf{volume~51}(6), pp.
  3103--3114 (2021).

\bibitem{nissan1997upgrading}
E.~Nissan, H.~Siegelmann, A.~Galperin, and S.~Kimhi.
\newblock \enquote{Upgrading Automation for Nuclear Fuel In-Core Management:
  from the Symbolic Generation of Configurations, to the Neural Adaptations of
  Heuristics.}
\newblock \emph{Engineering with Computers}, \textbf{volume~13}, pp. 1--19
  (1997).

\bibitem{radaideh2021physics}
M.~I. Radaideh, I.~Wolverton, J.~Joseph, J.~J. Tusar, U.~Otgonbaatar, N.~Roy,
  B.~Forget, and K.~Shirvan.
\newblock \enquote{Physics-informed reinforcement learning optimization of
  nuclear assembly design.}
\newblock \emph{Nuclear Engineering and Design}, \textbf{volume 372}, p. 110966
  (2021).

\bibitem{rempe1989simulate}
K.~R. Rempe, K.~S. Smith, and A.~F. Henry.
\newblock \enquote{SIMULATE-3 pin power reconstruction: methodology and
  benchmarking.}
\newblock \emph{Nuclear Science and Engineering}, \textbf{volume 103}(4), pp.
  334--342 (1989).

\bibitem{seurin2022pwr}
P.~Seurin and K.~Shirvan.
\newblock \enquote{PWR Loading Pattern Optimization with Reinforcement
  Learning.}
\newblock \emph{International Conference on Physics of Reactors (PHYSOR 2022)},
  pp. 1166--1175 (2022).

\bibitem{seurin2023pareto}
P.~Seurin and K.~Shirvan.
\newblock \enquote{Pareto Envelope Augmented with Reinforcement Learning:\\
  Multi-objective reinforcement learning-based approach for \\ Pressurized
  Water Reactor optimization.}
\newblock In \emph{The International Conference on Mathematics and
  Computational Methods Applied to Nuclear Science and Engineering (M\&C
  2023)}. Niagara Falls, Ontario, Canada, August 13-17 (2023).

\bibitem{bertsekas1996neuro}
D.~P. Bertsekas and J.~N. Tsitsiklis.
\newblock \emph{Neuro-Dynamic Programming}, volume~1.
\newblock Athena scientific Belmont, MA (1996).

\bibitem{bengio2018machine}
Y.~Bengio, A.~Lodi, and A.~Prouvost.
\newblock \enquote{Machine Learning for Combinatorial Optimization: a
  Methodological Tour d'Horizon.}
\newblock \emph{European Journal of Operational Research}, \textbf{volume
  290}(2), pp. 405--421 (2018).

\bibitem{hessel2018raindbow}
M.~Hessel, J.~Modayil, H.~van Hasselt, T.~Schaul, W.~Ostrovski, G.and~Dabney,
  D.~Horgan, B.~Piot, M.~Azar, and D.~Silver.
\newblock \enquote{Rainbow: Combining Improvements in Deep Reinforcement
  Learning.}
\newblock \emph{AAAI'18/IAAI'18/EAAI'18: Proceedings of the Thirty-Second AAAI
  Conference on Artificial Intelligence and Thirtieth Innovative Applications
  of Artificial Intelligence Conference and Eighth AAAI Symposium on
  Educational Advances in Artificial Intelligence}, \textbf{volume 393}, pp.
  3215--3222 (2018).

\bibitem{schulman2017proximal}
J.~Schulman, F.~Wolski, P.~Dhariwal, A.~Radford, and O.~Klimov.
\newblock \enquote{Proximal policy optimization algorithms.}
\newblock \emph{arXiv preprint arXiv:170706347} (2017).

\bibitem{schulman2017trust}
J.~Schulman, S.~Levine, P.~Moritz, M.~Jordan, and P.~Abbeel.
\newblock \enquote{Trust Region Policy Optimization.}
\newblock \emph{arXiv preprint arXiv: 150205477v5} (2017).

\bibitem{botvinick2019reinforcement}
M.~Botvinick, S.~Ritter, J.~X. Wang, Z.~Kurth-Nelson, C.~Blundell, and
  D.~Hassabis.
\newblock \enquote{Reinforcement Learning: Fast and Slow.}
\newblock \emph{Trends in Cognitive Sciences}, \textbf{volume~23}(5), pp.
  408--422 (2019).

\bibitem{williams1992simple}
R.~J. Williams.
\newblock \enquote{Simple statistical gradient-following algorithms for
  connectionest reinforcement learning.}
\newblock \emph{Machine Learning}, \textbf{volume~8}(3), pp. 229--256 (1992).

\bibitem{wu2017scalable}
Y.~Wu, E.~Mansimov, S.~Liao, R.~Grosse, and J.~Ba.
\newblock \enquote{Scalable trust-region method for deep reinforcement learning
  using the Kronecker-factored approximation.}
\newblock \emph{NIPS'17: Proceedings of the 31st International Conference on
  Neural Information Processing Systems}, pp. 5285--5294 (2017).

\bibitem{kakade2002approximately}
S.~Kakade and J.~Langford.
\newblock \enquote{Approximately Optimal Approximate Reinforcement Learning.}
\newblock \emph{ICML}, \textbf{volume~2}, pp. 267--274 (2002).

\bibitem{kakade2001anatural}
S.~Kakade.
\newblock \enquote{A natural policy gradient.}
\newblock \emph{NIPS'01: Proceedings of the 14th International Conference on
  Neural Information Processing Systems: Natural and Synthetic}, p. 1531–1538
  (2001).

\bibitem{stable-baselines}
A.~Hill, A.~Raffin, M.~Ernestus, A.~Gleave, A.~Kanervisto, R.~Traore,
  P.~Dhariwal, C.~Hesse, O.~Klimov, A.~Nichol, M.~Plappert, A.~Radford,
  J.~Schulman, S.~Sidor, and Y.~Wu.
\newblock \enquote{Stable Baselines.} (2018).
\newblock \urlprefix\url{https://github.com/hill-a/stable-baselines}.

\bibitem{mazyavkina2021reinforcement}
N.~Mazyavkina, S.~Sviridov, S.~Ivanov, and E.~Burnaev.
\newblock \enquote{Reinforcement Learning for Combinatorial Optimization: A
  survey.}
\newblock \emph{Computers and Operations Research}, \textbf{volume 134}, p.
  105400 (2021).

\bibitem{bertsimas2008introduction}
D.~Bertsimas and J.~N. Tsitsiklis.
\newblock \emph{Introduction to Linear Optimization}.
\newblock Dynamic Ideas, Athena Scientific (2008).

\bibitem{dai2016discriminative}
H.~Dai, B.~Dai, and L.~Song.
\newblock \enquote{Discriminative embeddings of latent variable models for
  structured data.}
\newblock \emph{ICML'16: Proceedings of the 33rd International Conference on
  International Conference on Machine Learning}, \textbf{volume~48}, pp.
  2701--2711 (2016).

\bibitem{vinyals2015pointer}
O.~Vinyals, M.~Fortunato, and N.~Jaitly.
\newblock \enquote{Pointer networks.}
\newblock \emph{NIPS'15: Proceedings of the 28th International Conference on
  Neural Information Processing Systems}, \textbf{volume~2}, pp. 2692--2700
  (2015).

\bibitem{nazari2018deep}
M.~Nazari, A.~Oroojlooy, L.~V. Snyder, and M.~Takac.
\newblock \enquote{Deep Reinforcement Learning for Solving the Vehicle Routing
  Problem.}
\newblock \emph{NIPS'18: Proceedings of the 32nd International Conference on
  Neural Information Processing Systems}, pp. 9861--9871 (2018).

\bibitem{mnih2016asynchronous}
V.~Mnih, A.~P. Badia, M.~Mirza, A.~Graves, T.~P. Lillicrap, T.~Harley,
  D.~Silver, and K.~Kavukcuoglu.
\newblock \enquote{Asynchronous methods for deep reinforcement learning.}
\newblock \emph{ICML'16: Proceedings of the 33rd International Conference on
  International Conference on Machine Learning}, \textbf{volume~48}, pp.
  1928--1937 (2016).

\bibitem{xing2020solve}
R.~Xing, S.~Tu, and L.~Xu.
\newblock \enquote{Solve Traveling Salesman Problem by Monte Carlo Tree Search
  and Deep Neural Network.}
\newblock \emph{arxiv preprint arXiv:200506879v1} (2020).

\bibitem{emami2018learning}
P.~Emami and S.~Ranka.
\newblock \enquote{Learning Permutations with Sinkhorn Policy Gradient.}
\newblock \emph{arXiv preprint arXiv:180507010v1} (2018).

\bibitem{kool2018attention}
W.~Kool, H.~Van~Hoof, and M.~Welling.
\newblock \enquote{Attention Solves Your TSP, Approximately.}
\newblock \emph{arXiv preprint arXiv:180308475v2} (2018).

\bibitem{solozabal2020constrained}
R.~Solozabal, J.~Ceberio, and M.~Takáč.
\newblock \enquote{Constrained Combinatorial Optimization with Reinforcement
  Learning.}
\newblock \emph{arXiv} (2020).
\newblock \urlprefix\url{https://arxiv.org/abs/2006.11984}.

\bibitem{delarue2020reinforcement}
A.~Delarue, R.~Anderson, and C.~Tjandraatmadja.
\newblock \enquote{Reinforcement Learning with Combinatorial Actions: An
  Application to Vehicle Routing.}
\newblock \emph{NIPS'20: Proceedings of the 34th International Conference on
  Neural Information Processing Systems}, \textbf{volume~52}, pp. 609--620
  (2020).

\bibitem{radaideh2021large}
M.~I. Radaideh, B.~Forget, and K.~Shirvan.
\newblock \enquote{Large-scale design optimisation of boiling water reactor
  bundles with neuroevolution.}
\newblock \emph{Annals of Nuclear Energy}, \textbf{volume 160}, p. 108355
  (2021).

\bibitem{kerkar2007exploitation}
N.~Kerkar and P.~Paulin.
\newblock \emph{Exploitation des coeurs REP}.
\newblock EDP SCIENCES, 17, avenue du Hoggar, Parc d'activités de Courtaboeuf,
  BP 112, 91944 Les Ulis Cedex A, France. (2007).

\bibitem{delcampo2004development}
C.~M. del Campo, J.~L. François, L.~Avendano, and M.~Gonzalez.
\newblock \enquote{Development of a BWR loading pattern design system based on
  modified genetic algorithms and knowledge.}
\newblock \emph{Annals of Nuclear Energy}, \textbf{volume~31}, pp. 1901--1911
  (2004).

\bibitem{castillo2004bwr}
A.~Castillo, G.~Alonso, L.~B. Morales, C.~M. del Campo, J.~L. François, and
  E.~del Valle.
\newblock \enquote{BWR fuel reloads design using a Tabu search technique.}
\newblock \emph{Annals of Nuclear Energy}, \textbf{volume~31}, pp. 151--161
  (2004).

\bibitem{radaideh2021rule}
M.~I. Radaideh and K.~Shirvan.
\newblock \enquote{Rule-based reinforcement learning methodology to inform
  evolutionary algorithms for constrained optimization of engineering
  applications.}
\newblock \emph{Knowledge-Based Systems}, \textbf{volume 217}, p. 106836
  (2021).

\bibitem{nijimbere2021tabu}
D.~Nijimbere, S.~Zhao, X.~Gu, and M.~O. Esangbedo.
\newblock \enquote{TABU SEARCH GUIDED BY REINFORCEMENT LEARNING FOR THE
  MAX-MEAN DISPERSION PROBLEM.}
\newblock \emph{JOURNAL OF INDUSTRIAL AND MANAGEMENT OPTIMIZATION},
  \textbf{volume~17}, pp. 3223--3246 (2021).

\bibitem{saccheri2004atight}
J.~G.~B. Saccheri, N.~E. Todreas, and M.~J. Driscoll.
\newblock \enquote{A tight lattice, Epithermal Core Design for the Integral
  PWR.}
\newblock In \emph{Proceedings of ICAPP '04}, p. 4359. Pittsburgh, PA, USA
  (2004).

\bibitem{nrcXXXX0523}
\enquote{0523 - 0504P - Westinghouse Advanced Technology - 03.4 - Analysis of
  Technical Specifications Unit 4.}
\newblock \urlprefix\url{nrc.gov/docs/ML1121/ML11216A087.pdf}.

\bibitem{liu2021policy}
Y.~Liu, A.~Halev, and X.~Liu.
\newblock \enquote{Policy Learning with Constraints in Model-free Reinforcement
  Learning: A Survey.}
\newblock \emph{International Joint Conferences on Artificial Intelligence
  Organization}, pp. 4508--4515 (2021).
\newblock \urlprefix\url{https://doi.org/10.24963/ijcai.2021/614}.
\newblock Survey Track.

\bibitem{li2022areview}
Z.~Li, J.~Wang, and M.~Ding.
\newblock \enquote{A review on optimization methods for nuclear reactor fuel
  reloading analysis.}
\newblock \emph{Nuclear Engineering and Design}, \textbf{volume 397}, p. 111950
  (2022).

\bibitem{kropaczek1991incore}
D.~J. Kropaczek and P.~J. Turinsky.
\newblock \enquote{In-core nuclear fuel management optimization for pressurized
  water reactors utilizing simulated annealing.}
\newblock \emph{Nuclear Technology}, \textbf{volume~95}(1), pp. 9--32 (1991).

\bibitem{francois2012comparison}
J.~L. François, J.~J. Ortiz-Sevrin, C.~Martin-del Campo, A.~Castillo, and
  J.~Esquivel-Estrada.
\newblock \enquote{Comparison of metaheuristic optimization techniques for BWR
  fuel reloads pattern design.}
\newblock \emph{Annals of Nuclear Energy}, \textbf{volume~51}, pp. 189--195
  (2012).

\bibitem{ivanov2021assessment}
B.~D. Ivanov and D.~J. Kropaczek.
\newblock \enquote{ASSESSMENT OF PARALLEL SIMULATED ANNEALING PERFORMANCE WITH
  THE NEXUS/ANC9 CORE DESIGN CODE SYSTEM.}
\newblock \emph{EPJ Web of Conferences}, \textbf{volume 247}, p. 02019 (2021).
\newblock \urlprefix\url{https://doi.org/10.1051/epjconf/202124702019}.

\bibitem{meneses2009particle}
A.~A. de~Moura~Meneses, M.~D. Machado, and R.~Schirru.
\newblock \enquote{Particle Swarm Optimization applied to the nuclear reload
  problem of a Pressurized Water Reactor.}
\newblock \emph{Progress in Nuclear Energy}, \textbf{volume~51}, pp. 319--326
  (2009).

\bibitem{derrac2011apractical}
J.~Derrac, S.~Garcia, D.~Molina, and F.~Herrera.
\newblock \enquote{A practical tutorial on the use of nonparametric statistical
  tests as a methodology for comparing evolutionary and swarm intelligence
  algorithms.}
\newblock \emph{Swarm and Evolutionary Computation}, \textbf{volume~1}, pp.
  3--18 (2011).

\bibitem{schlunz2016acomparative}
E.~Schlünz, P.~Bokov, and J.~van Vuuren.
\newblock \enquote{A comparative study on multiobjective metaheuristics for
  solving constrained in-core fuel management optimisation problems.}
\newblock \emph{Computers \& Operations Research}, \textbf{volume~75}, pp.
  174--190 (2016).

\bibitem{casella2002statistical}
G.~Casella and R.~L. Berger.
\newblock \emph{Statistical Inference Second Edition}.
\newblock Pacific Grove (2002).

\bibitem{2020SciPy-NMeth}
P.~Virtanen, R.~Gommers, T.~E. Oliphant, M.~Haberland, T.~Reddy, D.~Cournapeau,
  E.~Burovski, P.~Peterson, W.~Weckesser, J.~Bright, S.~J. {van der Walt},
  M.~Brett, J.~Wilson, K.~J. Millman, N.~Mayorov, A.~R.~J. Nelson, E.~Jones,
  R.~Kern, E.~Larson, C.~J. Carey, {\.I}.~Polat, Y.~Feng, E.~W. Moore,
  J.~{VanderPlas}, D.~Laxalde, J.~Perktold, R.~Cimrman, I.~Henriksen, E.~A.
  Quintero, C.~R. Harris, A.~M. Archibald, A.~H. Ribeiro, F.~Pedregosa, P.~{van
  Mulbregt}, and {SciPy 1.0 Contributors}.
\newblock \enquote{{{SciPy} 1.0: Fundamental Algorithms for Scientific
  Computing in Python}.}
\newblock \emph{Nature Methods}, \textbf{volume~17}, pp. 261--272 (2020).

\bibitem{Terpilowski2019}
M.~Terpilowski.
\newblock \enquote{scikit-posthocs: Pairwise multiple comparison tests in
  Python.}
\newblock \emph{The Journal of Open Source Software}, \textbf{volume~4}(36), p.
  1169 (2019).

\bibitem{yilmaz2022lot}
B.~G. Yilmaz and {\"O}.~F. Yilmaz.
\newblock \enquote{Lot streaming in hybrid flowshop scheduling problem by
  considering equal and consistent sublots under machine capability and limited
  waiting time constraint.}
\newblock \emph{Computers \& Industrial Engineering}, \textbf{volume 173}, p.
  108745 (2022).

\bibitem{yilmaz2022tactical}
{\"O}.~F. Yilmaz and B.~Yazici.
\newblock \enquote{Tactical level strategies for multi-objective disassembly
  line balancing problem with multi-manned stations: an optimization model and
  solution approaches.}
\newblock \emph{Ann Oper Res}, \textbf{volume 319}, p. 1793–1843 (2022).
\newblock \urlprefix\url{https://doi.org/10.1007/s10479-020-03902-3}.

\bibitem{awad2017problem}
N.~H. Awad, M.~Z. Ali, P.~N. Suganthan, J.~J. Liang, and B.~Y. Qu.
\newblock \enquote{Problem Definitions and Evaluation Criteria for the CEC 2017
  Special Session and Competition on Single Objective Real-Parameter Numerical
  Optimization.}
\newblock Technical report, Nanyang Technological University, Singapore (2017).

\bibitem{radaideh2023NEORL}
M.~I. Radaideh, P.~Seurin, K.~Du, D.~Seyler, X.~Gu, H.~Wang, and K.~Shirvan.
\newblock \enquote{NEORL: NeuroEvolution Optimization with Reinforcement
  Learning—Applications to carbon-free energy systems.}
\newblock \emph{Nuclear Engineering and Design}, p. 112423 (2023).

\bibitem{mnih2015human}
V.~Mnih, K.~Kavukcuoglu, D.~Silver, A.~A.~. Rusu, J.~Veness, M.~G. Bellemare,
  A.~Graves, M.~Riedmiller, A.~K. Fidjeland, and G.~e.~a. Ostrovski.
\newblock \enquote{Human-level control through deep reinforcement learning.}
\newblock \emph{nature}, \textbf{volume 518}(7540), pp. 529--533 (2015).

\bibitem{silver2017mastering}
D.~Silver, J.~Schrittwieser, K.~Simonyan, I.~Antonoglou, A.~Huang, A.~Guez,
  T.~Hubert, L.~Baker, M.~Lai, A.~Bolton, Y.~Chen, T.~Lillicrap, F.~Hui,
  L.~Sifre, G.~van~den Driessche, T.~Graepel, and D.~Hassabis.
\newblock \enquote{Mastering the game of Go without human knowledge.}
\newblock \emph{nature}, \textbf{volume 550} (2017).

\bibitem{konak2006multiobjective}
A.~Konak, D.~W. Coit, and A.~E. Smith.
\newblock \enquote{Multi-Objective optimizaton using genetic algorithms: A
  tutorial.}
\newblock \emph{Reliability Engineering \& System Safety}, \textbf{volume~91},
  pp. 992--1007 (2006).

\bibitem{garco2008new}
F.~Alim, I.~Kostadin, and S.~Levine.
\newblock \enquote{New genetic algorithms (GA) to optimize PWR reactors: Part
  I: Loading pattern and burnable poison placement optimization techniques for
  PWRs.}
\newblock \emph{Annals of Nuclear Energy}, \textbf{volume~35}(1), pp. 93--112
  (2008).

\bibitem{verhagen1997rosa}
F.~Verhagen, M.~Van~der Schaar, W.~De~Kruijf, T.~Van~de Wetering, and R.~Jones.
\newblock \enquote{ROSA, a utility tool for loading pattern optimization.}
\newblock \emph{Proc of the ANS Topical Meeting--Advances in Nuclear Fuel
  Management II}, \textbf{volume~1}, pp. 8--31 (1997).

\bibitem{frazier2018tutorial}
P.~Frazier.
\newblock \enquote{A Tutorial on Bayesian Optimization.}
\newblock \emph{ArXiv}, \textbf{volume abs/1807.02811} (2018).

\bibitem{van2016deep}
H.~Van~Hasselt, A.~Guez, and D.~Silver.
\newblock \enquote{Deep reinforcement learning with double q-learning.}
\newblock In \emph{AAAI'16: Proceedings of the Thirtieth AAAI Conference on
  Artificial Intelligence}, pp. 2094--2100 (2016).

\bibitem{schulman2018high}
J.~Schulman, P.~Moritz, S.~Levine, M.~Jordan, and P.~Abbeel.
\newblock \enquote{High-Dimensional Continuous Control Using Generalized
  Advantage Estimation.}
\newblock \emph{arXiv preprint arXiv:150602438v6} (2018).

\end{thebibliography}
